\documentclass{article}

\usepackage{microtype}
\usepackage{graphicx}
\usepackage{subcaption}
\usepackage{booktabs} 

\usepackage{hyperref}

 \usepackage[main, final]{neurips_2026}

\usepackage{amsmath}
\usepackage{amssymb}
\usepackage{mathtools}
\usepackage{amsthm}

\usepackage[capitalize,noabbrev]{cleveref}

\usepackage[utf8]{inputenc} 
\usepackage[T1]{fontenc}    
\usepackage{url}            
\usepackage{booktabs}       
\usepackage{amsfonts}       
\usepackage{nicefrac}       
\usepackage{xcolor}         
\usepackage{tabularx}
\usepackage{makecell}
\usepackage{adjustbox}
\usepackage{enumitem}

\usepackage{xcolor}
\usepackage{tikz}
\usepackage{tcolorbox}
\usepackage[dvipsnames]{xcolor}

\providecommand{\countdown}{\textsc{Countdown}}
\providecommand{\passk}{Pass@32}
\providecommand{\strictacc}{All-correct@32}
\providecommand{\avgacc}{Average Accuracy}

\title{How Does RL Post-training Induce Skill Composition? \\ A Case Study on \countdown{}}

\author{
Simon Park\begin{NoHyper}\thanks{Equal contribution}\end{NoHyper}~, \; Simran Kaur\footnotemark[1]~, \; Sanjeev Arora\\
Princeton Language and Intelligence (PLI) \\
Princeton University\\
\texttt{\{juhyunp,skaur,arora\}@princeton.edu}
}

\begin{document}
\maketitle

\addtocounter{footnote}{-1}
\begin{abstract}
While reinforcement learning (RL) successfully enhances reasoning in large language models, its role in fostering compositional generalization (the ability to synthesize novel skills from known components) is often conflated with mere length generalization. To this end, we study what RL post-training teaches models about skill composition and how the composition structure affects the skill transfer. We focus on the \countdown{} task (given $n$ numbers and a target, form an expression that evaluates to the target) and analyze model solutions using expression trees, where each subtree corresponds to a reusable subtask and thus can be viewed as a ``skill.'' 
Tracking tree shapes and their success rates over training, we find:
(i) out-of-distribution (OOD) generalization to larger $n$ and to unseen tree shapes, indicating compositional reuse of subtasks; (ii) the order of skill acquisition depends upon the composition structure---models master shallow balanced trees (workload is balanced between subtasks) before deep unbalanced ones, with persistent fragility on right-heavy structures for a fixed composition depth.
Our diagnostic reveals what is learned, in what order, and where generalization fails, clarifying how RL-only post-training induces OOD generalization beyond what standard metrics such as pass@k reveal. 
To study the implications of our findings for real-world tasks, we present a preliminary study on formal theorem proving.~\footnote{Source code can be found at the \href{https://github.com/princeton-pli/RL-skill-comp}{PLI GitHub codebase}.}

\end{abstract}
\section{Introduction}
A key objective in training large language models (LLMs) is to develop robust reasoning abilities that generalize to novel problem instances. Reinforcement learning (RL) is a widely adopted technique for this purpose \citep{ouyang2022training,grattafiori2024llama,openai2024openaio1card,deepseekai2025deepseekr1incentivizingreasoningcapability}, yet the precise mechanism by which it improves reasoning remains poorly understood. The prevailing hypotheses suggest that RL either sharpens the model's probability distribution over known reasoning paths \citep{huang2024sharpening,yue2025does} or encourages in-context exploration via chaining in longer solution sequences \citep{deepseekai2025deepseekr1incentivizingreasoningcapability,setlur2025e3learningexploreenables}. However, these notions overlook a crucial aspect of intelligence: the ability to compose known skills in novel ways.

Current evaluations \citep{deepseekai2025deepseekr1incentivizingreasoningcapability,setlur2025e3learningexploreenables,sun2025omegallmsreasonoutside,zhang2025interplaypretrainingmidtrainingrl,Yuan2025FromFGtoFGx,chu2025sft} often conflate two fundamentally different modes of generalization. The first, \emph{length generalization},
is the ability to solve problems of greater sequential depth
by iterating a known algorithmic structure.
The second, \emph{compositional generalization}, 
is the ability to solve problems of a fixed depth by synthesizing a novel algorithmic structure from familiar components.
While complex reasoning often involves both, we argue that a rigorous understanding requires operationalizing and testing these two modes of generalization independently. Without a clear way to distinguish them, it is difficult to ascertain the true algorithmic contribution of fine-tuning methods like RL.

To formalize this distinction and enable empirical investigation, a controlled setting is required. We utilize the \countdown{}\footnote{\countdown{} is a task where given \(n\) integers and a target, the model needs to output an expression using each integer once and operators \(\{+,-,\times,\div\}\) that attains the target.
For example, given $[4, 3, 2, 2]$ and a target $16$, one solution would be $2 \times 2 \times 3 + 4 = 16$.} task, which serves as a clean testbed where the structure of a solution is isomorphic to a binary expression tree. This allows for an unambiguous, canonical representation of the reasoning pattern used to solve a problem. 

This pattern framework allows us to move beyond a monolithic view of the solution  to a nuanced skill-based view.
Solving a complex pattern like $A \times (B-C/D)$ requires a composition of distinct abilities: \emph{decomposition skills} to identify how to explore and correctly partition the solution (e.g., \texttt{factor\_identification}: identifying the factors of the target to explore potential candidates) and the recursive \emph{sub-problem skills} to solve the resulting sub-patterns (e.g., generating an expression for $B-C/D$ given the remaining three numbers and a modified target).\footnote{The atomic skills would be mastering the four operators: $a+b$, $a-b$, $a\times b$, and $a \div b$.} The tree structure of the pattern determines which atomic skills are combined, in what order, and how they are grouped.

By tracking the development of complex skills for each pattern during training, we probe two fundamental questions:
1) To what extent does RL equip models for genuine structural composition? and 2) How does the underlying composition structure of a problem affect how quickly a model learns to solve it? We summarize our main contributions:

\begin{itemize}[leftmargin=*]
  \item \textbf{Pattern framework.} We introduce a framework to disentangle length and compositional generalization (\Cref{sec:method}). We propose a method to remove distributional bias in \countdown{} datasets (\Cref{sec:balanced-dataset}). This may inspire future studies on other domains with a subtask structure.

  \item \textbf{Models exhibit two forms of OOD generalization.} Using our framework, we separately show that models generalize to \textit{larger puzzle sizes} (\Cref{sec:results:length-generalization-alt}), where skills for smaller patterns need to be reused more times, and to \textit{unseen patterns} (\Cref{sec:results:generalization-heldout-tree-alt}), where skills for other patterns are combined to synthesize a held-out pattern and its higher-order extensions.
  
  \item \textbf{Pattern structure determines difficulty.} Upon fixing an input length $n$, we observe that the difficulty of a pattern depends on its structure: shallow balanced structures (with equal amount of ``workload'' across subtasks) are easier than deep unbalanced structures (\Cref{sec:results:structure-alt}). Within the same depth, right-heavy trees, which require committing to a plan ahead of the complex subroutine appearing later, are particularly hard (\Cref{sec:results:right-heavy-alt}). A controlled supervised finetuning (SFT) comparison exhibits a different order of skill acquisition, suggesting that the hierarchy is inherent to RL, not an artifact of the task (\Cref{sec:sft}).

  \item \textbf{The effect of composition structure extends to formal theorem proving.} We show that an RL-trained model for formal theorem proving favors proofs with balanced subtasks, while avoiding proofs with right-heavy structure (\Cref{sec:lean}). This suggests that RL post-training selectively amplifies a model's behavior based on the difficulty induced by the composition structure.
  
\end{itemize}

Overall, our findings show that RL post-training imparts length and compositional generalization, with difficulty governed primarily by composition structure. 

This result motivates targeted curriculum design as a promising direction for future work. We hypothesize that during RL post-training, once early generalization gains appear, introducing a minimal set of structurally hard, near-frontier examples could offer a more sample-efficient method for pushing the model's reasoning capabilities.
\section{Methodology: Formalizing Length and Compositional Generalization}
\label{sec:method}
Standard metrics for reasoning, such as \textit{pass@k}, \textit{response length}, or \textit{number of attempts} offer an incomplete picture of a model's capabilities. They fail to distinguish between a model's ability to iterate a known procedure and the more formidable challenge of synthesizing a novel one. To address this, we introduce a formal framework that disentangles these phenomena by analyzing the underlying computational structure of arithmetic solutions, which we apply to the \countdown{} task.

\clearpage

\subsection{Canonical Patterns of Reasoning}
We define an \textit{atomic skill} as the application of a single binary operator from $\{+,-,\times,\div\}$. Then the ability to solve a \countdown{} puzzle with an arithmetic expression requires a \textit{skill composition}. By replacing numbers with placeholder symbols, an arithmetic expression can be abstracted (e.g., $9/3+1 \rightarrow \texttt{A/B+C}$) and parsed into a binary tree, which provides insights into how the atomic skills are combined.

To ensure that syntactically different but algorithmically identical expressions (e.g., $(A+B)+C$ and $C+(A+B)$) are treated as a single computational procedure, we apply a principled normalization process that maps them to a unique canonical \textit{pattern} (\Cref{fig:countdown}). This process, detailed in \Cref{appendix:pattern}, resolves ambiguities arising from properties like commutativity and associativity, allowing us to analyze the underlying computational graph and distinguish a simple iterative structure like $A+B+C+D$ from a complex, nested one like $A/(B-C/D)$.

We classify the structure of a computational pattern by its shape signature $[x]\circ[y]$, denoting the root operator $\circ$ and the number of leaf operands $x$ and $y$ in its left and right sub-trees. This partitioning into left-heavy ($x>y$), balanced ($x \approx y$), and right-heavy ($x<y$) structures correlates with synthesis difficulty (see Figure \ref{fig:tree-structure}). For instance, 
to generate a right-heavy pattern like $A/(B-C/D)$ (signature $[1]\div[3]$), a model cannot greedily select the root operator; it must first look ahead to the complex sub-expression $(B-C/D)$ and recognize its utility within the broader computation.

\begin{figure*}[!t]
    \centering
    \includegraphics[width=0.99\linewidth]{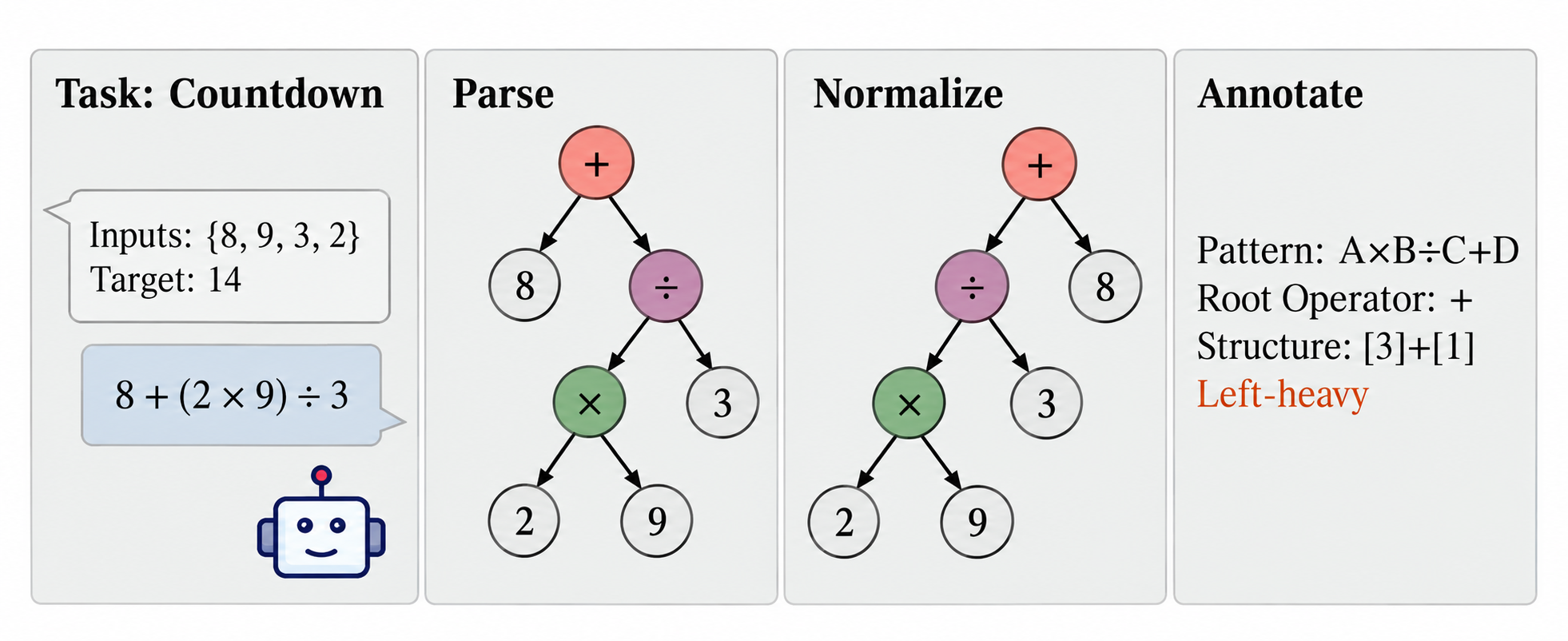}
    \captionof{figure}{\textbf{From generated expression to canonical \emph{pattern}.} The model-generated expression $8 + (2 \times 9) \div 3$ is parsed into an expression tree, normalized with respect to algebraic identities, and mapped to the unique canonical pattern $A \times B \div C + D$. Patterns are then grouped by their computational shape. Here, the root operator is ``$+$'', with a left sub-expression over three leaves $(A,B,C)$ and a right sub-expression over one leaf $(D)$. The signature $[3]+[1]$ serves as a compact representation of the pattern's structure at the root-level.}
    \label{fig:countdown}
\end{figure*}

\subsection{Disentangling Modes of Generalization}
A primary contribution of our framework is its ability to formally disentangle two distinct and often-conflated axes of skill composition. We define \textbf{length generalization} as the ability to generalize to larger problem sizes than those seen during training, which typically requires using more atomic skills. We define \textbf{compositional generalization} as the ability to reconstruct novel combinations of known atomic skills at problem sizes that remain within the training range.

In \countdown{}, we quantify length generalization by the model’s success rate at puzzles larger than those it was trained on, and compositional generalization by its success rate on held-out patterns (holding the puzzle size fixed).

Our detailed study reveals interesting phenomena that were missed by earlier works (e.g., \citet{Yuan2025FromFGtoFGx}), which consider composition simply as a sequential application (``chaining'') of atomic skills. For example, these works do not differentiate between the difficulty of atomic skills ($+$ and $\div$ are considered equal in difficulty, which violates intuition). 

\clearpage

\subsection{Controlling for Distributional Bias in Dataset Generation}
\label{sec:balanced-dataset}
A rigorous analysis of structural difficulty requires that the distribution of problem types be controlled. The standard approach for creating \countdown{} datasets \citep{tinyzero,gandhi2024streamsearchsoslearning,stojanovski2025reasoninggymreasoningenvironments} proceeds as follows: 1) choose a random binary tree; 2) choose random operators, numbers; 3) then filter for valid expressions that evaluate to an integer in $[1, 99]$. This introduces severe distributional biases, creating a confounding variable between a pattern's frequency and its intrinsic complexity. There are two problems with this design:

\begin{itemize}[leftmargin=*]
    \item \textbf{Structural bias.} $+$ and $\times$ are commutative and associative, and many expressions involving the two operators may have the same underlying pattern and have a higher probability of being generated. For instance, the same $A+B+C+D$ puzzle can be generated in $4! \times C_3 = 120$ ways (e.g., $((1+2)+3)+4$ and $1+(4+(3+2))$), whereas an asymmetric pattern like $A/(B-C/D)$ only has one expression.

    \item \textbf{Selection bias.} The second, more severe bias stems from the common practice of filtering for solutions that evaluate to an integer within a fixed range, such as $[1,99]$. When random inputs are drawn from the set range, $+$ or $-$ yields a result in this range approximately 50\% of the time, whereas $\times$ or $\div$ do so only about 5\% of the time. Consequently, each $\times$ or $\div$ is approximately one-tenth as likely to appear in an expression.
\end{itemize}

These two biases together produce a skewed dataset where aggregate performance metrics are dominated by a model's performance on simple, additive patterns. To eliminate this simplicity bias, we designed a different generation protocol. We generate examples by first selecting a target canonical pattern and then sampling numbers that satisfy it. This enforces a near-uniform distribution over all patterns, a methodological prerequisite for isolating and analyzing the true effect of compositional structure on reasoning difficulty. See \Cref{appendix:dataset} for more details.

\begin{figure*}[t]
  \centering
  \includegraphics[width=\textwidth]{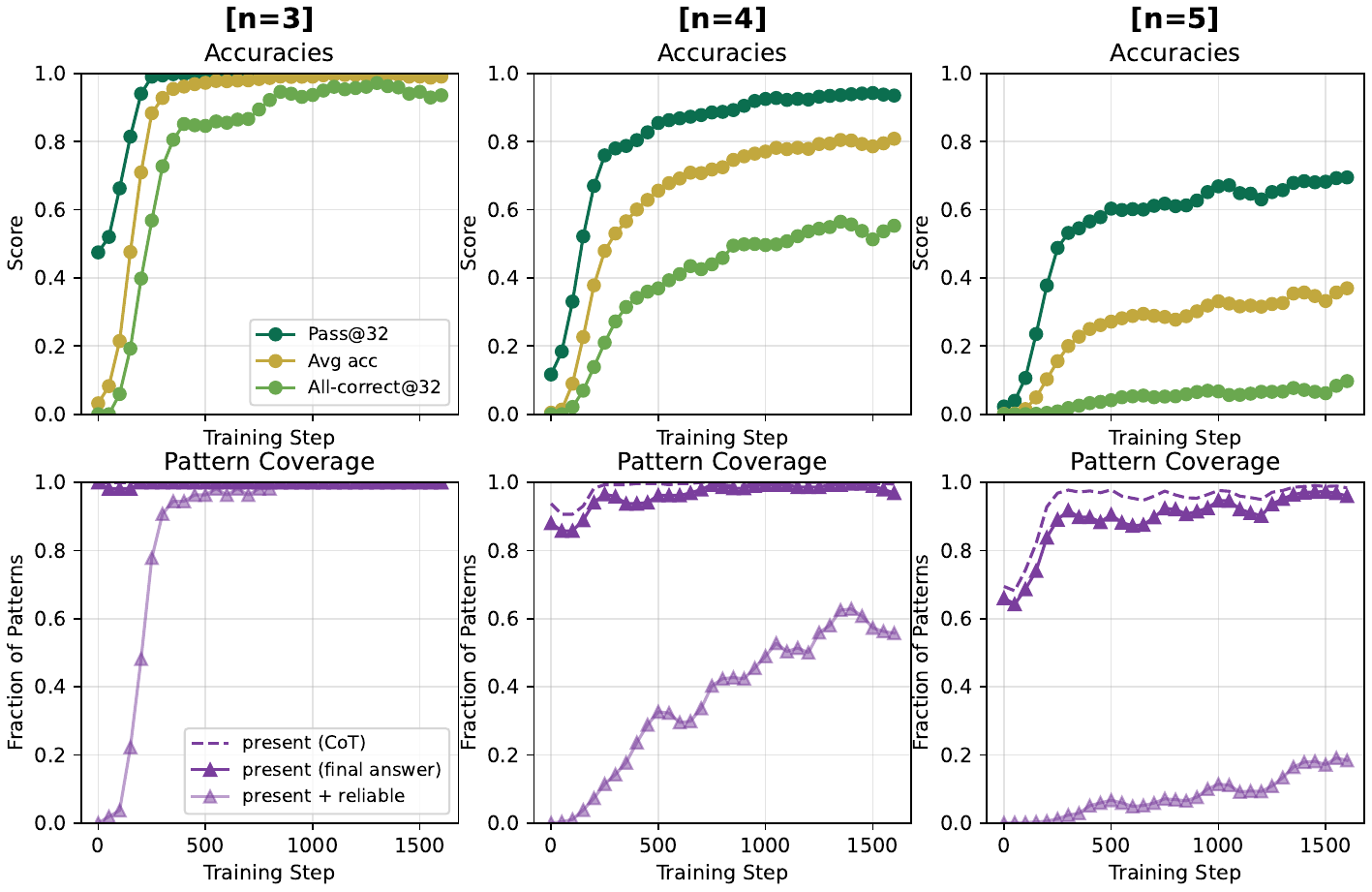}
\captionof{figure}{
    \textbf{Successful length generalization, but a gap between first producing a pattern and producing it reliably.}
    We train \texttt{Qwen2.5-1.5B} on problem sizes $n \in \{3, 4\}$ and evaluate on held-out data for $n \in \{3,4,5\}$. \textbf{(Top)} The model demonstrates strong \textit{length generalization}: accuracy on the held-out $n=5$ problems improves over the base model (leftmost data point). \textbf{(Bottom)} The model quickly learns to produce most $n=5$ patterns at least once, but the ability to reliably produce the correct pattern (see \Cref{sec:eval} for the definition) lags significantly behind. 
}
  \label{fig:learning-dynamics}
\end{figure*}

\section{Experimental Setup}
\label{sec:setup}

\subsection{Training}
We finetune \texttt{Qwen-2.5} models (1.5B/3B/7B) \citep{qwen2025qwen25technicalreport} with GRPO \citep{shao2024deepseekmathpushinglimitsmathematical} on problems of size $n \in \{3, 4\}$. Each rollout receives a reward of 0.1 for correctly formatting the final expression and an additional reward of 0.9 if the expression is correct.
We train for one epoch across three seeds, 
discarding any unstable runs (see \Cref{sec:reward_hacking}). Full 
details and results on other model families (\texttt{Llama-3.1, Llama-3.2, Qwen3}) are available in \Cref{sec:generating-balanced-dataset,appendix:training,appendix:llama,appendix:qwen3}.

\subsection{Evaluation}
\label{sec:eval}
We evaluate each checkpoint by sampling $k=32$ outputs per held-out question (temperature $=0.6$), with the maximum token length increasing for larger problem sizes (1024 for $n \in \{3,4\}$, 2048 for $n=5$, and 4096 for $n=6$). We then compute the following metrics on the extracted final answers:

\begin{itemize}[leftmargin=*]
    \item \textbf{Pass@32 / All-correct@32:} The fraction of questions where at least one (Pass@32), or all (All-correct@32), of the $k$ sampled answers is correct.
   
    \item \textbf{Average Accuracy:} The total number of correct answers across all questions and attempts, divided by the total number of answers generated (i.e., number of questions $\times$ $k$).
   
    \item \textbf{Pattern Coverage:} The fraction of all possible canonical patterns that the model produces at least once across the evaluation set (such patterns will be labeled \texttt{present}). 
   
    \item \textbf{Per-Pattern Precision:} The number of correct solutions using a pattern, divided by the total number of times the model produced that pattern as a final answer across the entire evaluation set.
    
    \item \textbf{High-Precision Coverage:} The fraction of all possible canonical patterns for which the model's \textit{per-pattern precision} is at least $80\%$ (such patterns will be labeled \texttt{reliable}).\footnote{The 80\% threshold checks if the model can execute a pattern with reasonable reliability, but allows some mistakes. The main findings are robust to reasonable variations of the threshold (70-90\%). See \Cref{appendix:plots}.}
\end{itemize}
\section{Main Results}

We show that models achieve length (\Cref{sec:results:length-generalization-alt}) and compositional generalization (\Cref{sec:results:generalization-heldout-tree-alt}) independently. We then demonstrate that this success is fundamentally governed by the composition structure of the task (\Cref{sec:results:structure-alt}). We hypothesize that ``lookahead bottleneck'' is the primary barrier to skill composition (\Cref{sec:results:right-heavy-alt}). We find that the ``lookahead bottleneck'' does not appear in a controlled SFT setup and conclude that it is not an artifact of the task (\Cref{sec:sft}). Finally, we present extensions to formal theorem proving, a real-world application of our findings (\Cref{sec:lean}).

\subsection{Models Demonstrate Length Generalization}
\label{sec:results:length-generalization-alt}
RL effectively teaches models to apply learned reasoning procedures to problems of greater lengths than seen during training. Models trained only on $n \in \{3,4\}$ problems show substantial accuracy improvements on held-out $n=5$ puzzles (\Cref{fig:learning-dynamics}) over the base model, with final \passk{} reaching nearly 70\% for the 1.5B model (\Cref{tab:final-accuracy}). This generalization extends to $n=6$, where models achieve \passk{} around 40\%. These results confirm that models can achieve \textit{length generalization}.


\begin{table}[!h]
  \centering
  \caption{\textbf{Performance across model size.}
  \passk{} averaged across all successful seeds \protect\footnotemark (\textbf{final} checkpoints for 1.5B/3B and \textbf{best} checkpoints for 7B) \protect\footnotemark on $n{=}3,4$ (trained) and $n{=}5,6$ (held out). Models learn $n=3, 4$ patterns almost perfectly and generalize to larger puzzle sizes $n=5, 6$. See \Cref{appendix:alt-tables} for a more complete version of the table.
  }
  \label{tab:final-accuracy}
  \begin{tabular}{c *{4}{c}} \\
    \toprule
    Model & \multicolumn{1}{c}{$n{=}3$} & \multicolumn{1}{c}{$n{=}4$} & \multicolumn{1}{c}{$n{=}5$}  & \multicolumn{1}{c}{$n{=}6$}\\
    \midrule
    1.5B / 3B / 7B & 
    1.00 / 1.00 / 1.00 & 
    0.94 / 0.95 / 0.98 & 
    0.68 / 0.68 / 0.75 & 
    0.41 / 0.45 / 0.45 \\
    \bottomrule
  \end{tabular}
\end{table}

\addtocounter{footnote}{-1}
\footnotetext{2 seeds for 1.5B/3B and 3 for 7B. Due to compute budget, we evaluate only 1 seed per model for $n=6$.}

\addtocounter{footnote}{1}
\footnotetext{We select the checkpoint right before the training reward declines. See \Cref{appendix:alt-tables} for more details. }

\clearpage

\subsection{Models Demonstrate Compositional Generalization}
\label{sec:results:generalization-heldout-tree-alt}
We next test whether RL can enable a model to synthesize a novel reasoning pattern it has never been explicitly trained on, \textit{while holding the length constant}. We remove the $n=3$ pattern $A/B+C$ and its entire family of $n=4$ derivatives from the training dataset (e.g., $A/B+C+D$, $A/(B+C)+D$).~\footnote{See \Cref{table:heldout-tree3-list} in \Cref{sec:pattern_heldout} for the full list.} 

To isolate the effect of the RL training, we first test for the pretraining exposure. When evaluating the base model, the models are unsurprisingly able to generate all of these patterns, but very rarely to the correct puzzles where they should be used (even with $512\times$ the training budget). This observation suggests that the base model does not have the skill (or even if does, a very weak skill) to determine when to apply these patterns. See \Cref{sec:pattern_heldout_details} for more details.

\begin{figure*}[t]
  \centering
  \includegraphics[width=0.90\linewidth]{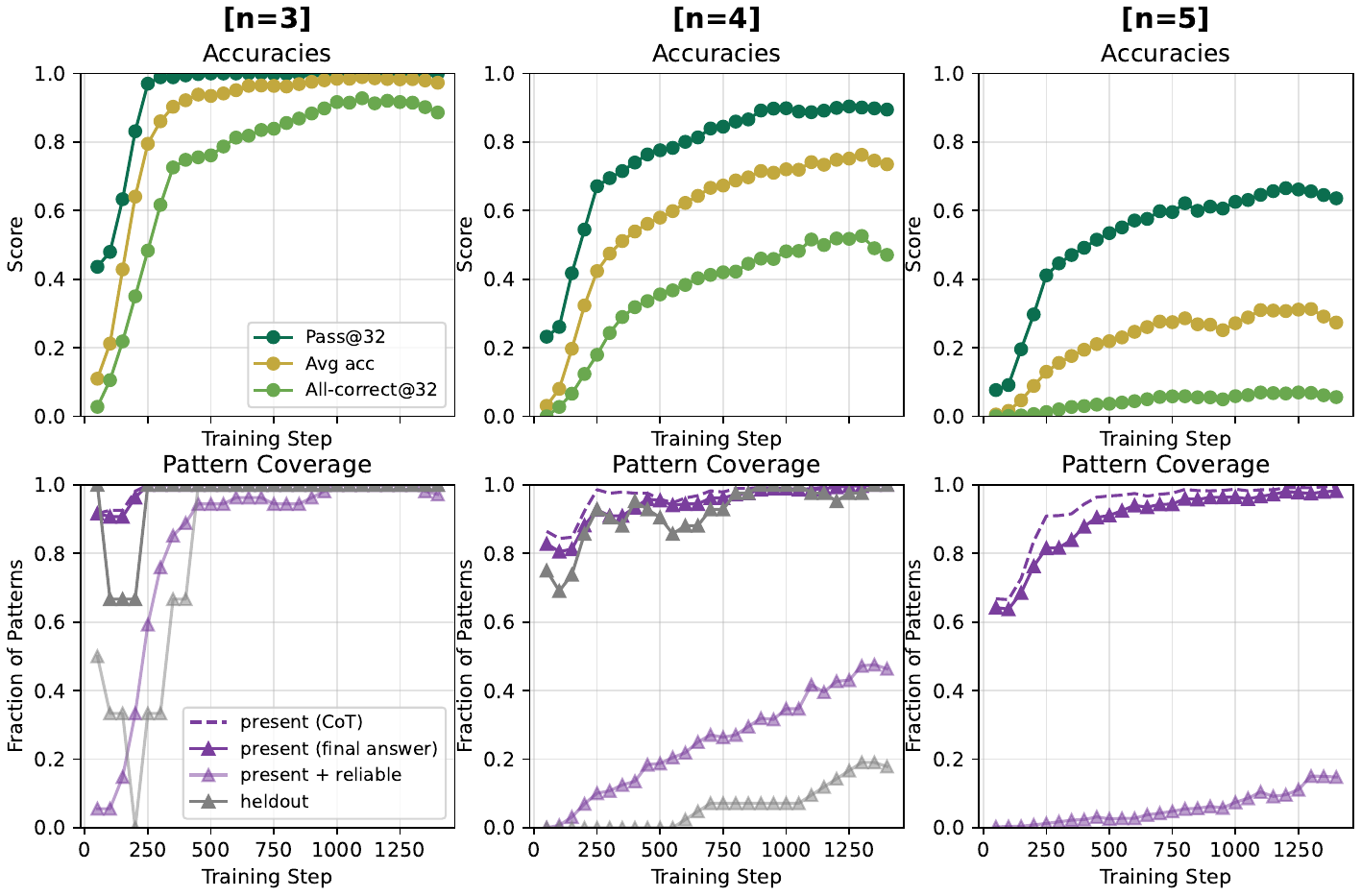}
  \caption{\textbf{Generalizes to entire families of held-out compositional patterns.} To evaluate compositional generalization, we removed an entire family of related patterns from the training set: the base $n=3$ structure $A/B+C$ and all of its $n=4$ extensions (e.g., $A/B+C+D$, $A/(B+C)+D$). The model successfully recovers these unseen patterns. Coverage first emerges on the held-out $n=3$ subpattern before generalizing to its more complex $n=4$ dependents, matching the typical learning hierarchy where presence precedes precision. This result demonstrates that the model can reuse learned substructures to synthesize novel operator-tree shapes it has never been explicitly trained on.}

  \label{fig:subtrees-heldout-accs}
\end{figure*}

\Cref{fig:subtrees-heldout-accs,fig:subtrees-heldout-accs-seed2,fig:subtrees-heldout-accs-seed3} show that \texttt{Qwen2.5-1.5B} trained on this modified dataset successfully learns to generate the held-out patterns despite zero training exposure. The learning dynamics also reveal: coverage first emerges on the simpler $n=3$ subpattern before the model learns to compose it into its more complex $n=4$ dependent forms. This demonstrates that the model is not merely matching seen templates but is actively reusing learned substructures to assemble novel patterns, fulfilling our definition of \textit{compositional generalization}.

\subsection{Composition Structure Determines Difficulty}
\label{sec:results:structure-alt}
\begin{figure}[h]
  \centering
  \includegraphics[width=0.75\linewidth]{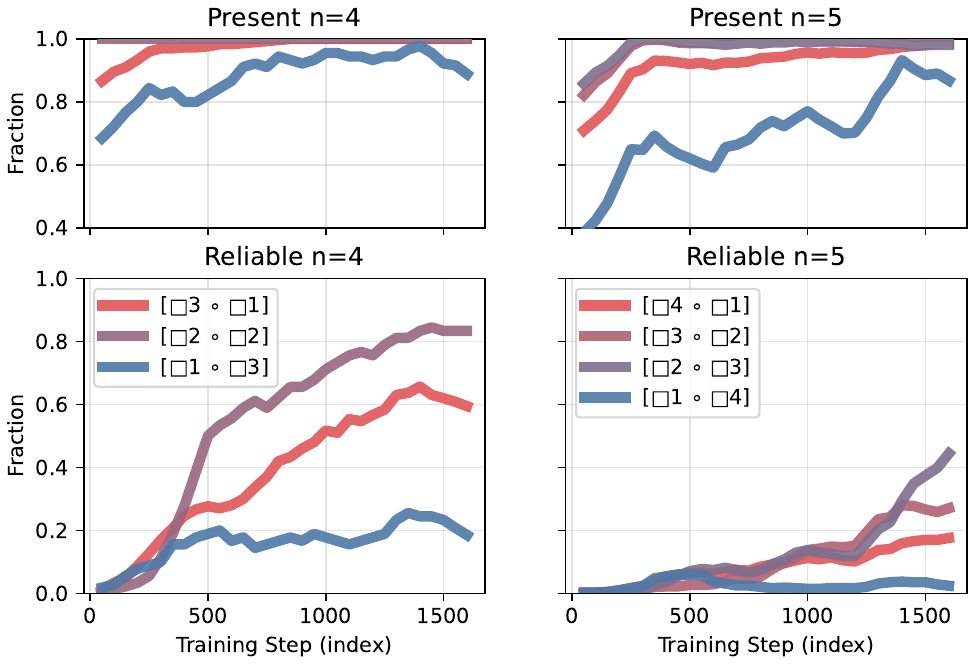}
    \captionof{figure}{\textbf{Composition structure (not input length) determines difficulty.} Within $n=4$, balanced patterns ($[2]\circ[2]$; purple) are discovered (top row) and mastered more reliably (bottom row) than the unbalanced ones (red/blue). Balanced patterns have shallow trees, with solution depth of 2 equal to that of $n=3$ puzzles. Even when controlling for depth, right-heavy patterns ($[1]\circ[3]$; blue) are harder than left-heavy patterns ($[3]\circ[1]$; red). This evidence points to a ``lookahead bottleneck,'' the challenge of committing to a solution ahead of a complex subroutine. The entire structural hierarchy persists for $n=5$. Results shown for a representative run of \texttt{Qwen2.5-1.5B}. See \Cref{appendix:plots} for equivalent plots for all other runs.}
  \label{fig:subtrees}
\end{figure}

However, the improvement in aggregate accuracy masks the true determinant of difficulty. The ability to generalize in length is fundamentally constrained by the composition structure of the pattern. This is evident in the out-of-distribution $n=5$ evaluation in \Cref{fig:subtrees}, which reveals the same hierarchy of difficulty seen on the training distribution (with $n=4$): balanced patterns are learned most readily, while right-heavy patterns remain the most challenging. This demonstrates that composition structure, not problem length, is the primary predictor of reasoning difficulty, even when generalizing to unseen lengths.

\begin{table*}[!h]
  \centering
  \captionof{table}{\textbf{Per-pattern precision averaged by pattern structure on $n{=}4$.}
  Averaged across all successful seeds (\textbf{final} checkpoints for 1.5B/3B and \textbf{best} checkpoints for 7B). Averaged by (a) root operator; (b) pattern with root operator $-$; (c) pattern with root operator $\div$. Models struggle more on $-, \div$ as the root operator. Shapes in the form of $[1] - [3]$ or $[1] \div [3]$ are particularly challenging. Within each column, the highest number is \textbf{boldfaced} and the second highest number is \underline{underlined}.}
  \label{tab:root-and-subtree}
  \setlength{\tabcolsep}{6pt}

  \begin{subtable}[t]{0.32\linewidth}
    \centering
    \caption{by root operator} 
    \begin{tabular}{@{}l c c c@{}}
      \toprule
      \textbf{Operator} & \textbf{1.5B} & \textbf{3B} & \textbf{7B} \\
      \midrule
      $+$      & \underline{0.86} & \underline{0.85} & \textbf{0.91} \\
      $-$      & 0.83 & 0.80 & 0.86 \\
      $\times$ & \textbf{0.90} & \textbf{0.86} & \textbf{0.91} \\
      $\div$   & 0.75 & 0.78 & 0.83 \\
      \bottomrule
    \end{tabular}
  \end{subtable}\hfill
  \begin{subtable}[t]{0.32\linewidth}
    \centering
    \caption{by subtree shape ($-$)}
    \begin{tabular}{@{}l c c c@{}}
      \toprule
      \textbf{Shape} & \textbf{1.5B} & \textbf{3B} & \textbf{7B} \\
      \midrule
      $\fbox{\scriptsize 3}\!-\!\fbox{\scriptsize 1}$ & \underline{0.81} & \underline{0.78} & 0.84 \\
      $\fbox{\scriptsize 2}\!-\!\fbox{\scriptsize 2}$ & \textbf{0.92} & \textbf{0.89} & \textbf{0.90} \\
      $\fbox{\scriptsize 1}\!-\!\fbox{\scriptsize 3}$ & 0.73 & 0.76 & \underline{0.86} \\
      \bottomrule
    \end{tabular}
  \end{subtable}\hfill
  \begin{subtable}[t]{0.32\linewidth}
    \centering
    \caption{by subtree shape ($\div$)}
    \begin{tabular}{@{}l c c c@{}}
      \toprule
      \textbf{Shape} & \textbf{1.5B} & \textbf{3B} & \textbf{7B} \\
      \midrule
      $\fbox{\scriptsize 3}\!\div\!\fbox{\scriptsize 1}$ & \underline{0.77} & \underline{0.77} & \underline{0.83} \\
      $\fbox{\scriptsize 2}\!\div\!\fbox{\scriptsize 2}$ & \textbf{0.88} & \textbf{0.85} & \textbf{0.91} \\
      $\fbox{\scriptsize 1}\!\div\!\fbox{\scriptsize 3}$ & 0.61 & 0.76 & 0.80 \\
      \bottomrule
    \end{tabular}
  \end{subtable}
\end{table*}

\subsection{The ``Lookahead Bottleneck'' Is a Fundamental Barrier to Skill Composition}
\label{sec:results:right-heavy-alt}
To isolate the effect of composition structure while holding problem length constant, we analyze performance on $n=4$ puzzles. Our findings reveal a clear difficulty hierarchy that suggests that ``lookahead bottleneck'' is a fundamental challenge for autoregressive models. In \Cref{sec:prefix}, we show an ablation on a prefix notation of \countdown{}, which further supports our hypothesis about the ``lookahead bottleneck.''

\Cref{fig:subtrees} ($n=4$ column) shows that balanced patterns ($[2]\circ[2]$), which decompose a problem into two equal subtasks, are mastered far more reliably than unbalanced ones. Crucially, right-heavy patterns ($[1]\circ[3]$) are substantially harder to learn than left-heavy ($[3]\circ[1]$) ones. To generate a solution for a right-heavy pattern like $A / (B - C/D)$, the model must commit to the root operator ($\div$) before generating the complex, three-term subroutine it applies to. This need to plan ahead for a complex subsequent operation is the primary failure mode.

\Cref{tab:root-and-subtree} quantifies this bottleneck across all model sizes: at the same depth of 3, $[1] \circ [3]$ generally underperforms $[3] \circ [1]$, especially with the root operator $\div$ and with smaller models. 

\clearpage

\subsection{Preliminary Evidence Beyond \countdown{}: Formal Theorem Proving}
\label{sec:lean}

We next ask whether our observations on \countdown{} also appear in a more realistic reasoning domain. We consider formal theorem proving in Lean \citep{lean4}, where many proof strategies naturally decompose into subtasks. In particular, we focus on tactics (e.g., \texttt{constructor}) that require the model to generate exactly two subtasks. A proof attempt can be categorized as balanced, left-heavy, or right-heavy based on the relative amount of workload in the two subtasks. To isolate the effect of RL post-training, we evaluate the behavior of the \texttt{Goedel-Prover-V2-32B} \citep{goedelv2}, whose SFT dataset is fully available. \footnote{Since the SFT checkpoint is not available, we use the distribution of the SFT dataset as the proxy for the distribution of proofs generated by the SFT checkpoint (before RL training).} For more details, see \Cref{app:lean}. 

In \Cref{tab:lean}, we provide three proof distributions: 1) SFT dataset; 2) output of the final checkpoint (after RL training) on the same SFT dataset (in-distribution); and 3) output of the final checkpoint on an out-of-distribution dataset. Compared to pre-RL training, the final checkpoint prefers balanced (51\% $\rightarrow$ 56\%) and left-heavy (14\% $\rightarrow$ 16\%) proofs and avoids right-heavy proofs (35\% $\rightarrow$ 29\%). This trend is consistent with the ``lookahead bottleneck'' identified in \countdown{}: if balanced proofs were easier to produce and right-heavy patterns were harder, RL post-training would have amplified the correct behavior (balanced), while inhibiting the incorrect behavior (right-heavy). 

\definecolor{deltaPos}{HTML}{0072B2}
\definecolor{deltaNeg}{HTML}{D55E00}

\begin{table}[ht]
    \centering
    \caption{\textbf{Distribution of proofs by composition structure (\%).} After RL post-training, the final checkpoint prefers balanced proofs and avoids right-heavy proofs. This trend is consistent with the ``lookahead bottleneck'' identified in \countdown{}: structures whose complex subroutine appears later are harder to produce reliably and are therefore less likely to be reinforced. The last row shows that the same qualitative shift persists on an out-of-distribution dataset which contains an approximately equal number of reference proofs across balanced / left-heavy / right-heavy categories.\\}
    \label{tab:lean}

    \begin{tabular}{@{}l r@{\;}l r@{\;}l r@{\;}l@{}}
        \toprule
        & \multicolumn{2}{c}{balanced}
        & \multicolumn{2}{c}{left-heavy}
        & \multicolumn{2}{c}{right-heavy} \\
        \midrule
        Present in SFT dataset
            & 51 & 
            & 14 & 
            & 35 & \\
        Generated by RL model (ID)
            & 56 & \textcolor{deltaPos}{+5}
            & 16 & \textcolor{deltaPos}{+2}
            & 29 & \textcolor{deltaNeg}{-6} \\
        Generated by RL model (OOD)
            & 68 & \textcolor{deltaPos}{+17}
            & 19 & \textcolor{deltaPos}{+5}
            & 13 & \textcolor{deltaNeg}{-22} \\
        \bottomrule
    \end{tabular}
\end{table}
\section{Ablations}
In \Cref{appendix:ablations}, we run ablations on different parts of the training pipeline. Small models often find a formatting-only shortcut, but increasing the group size initially mitigates the instability (\Cref{sec:reward_hacking}). Normalizing by the standard deviation in the group advantage leads to longer responses and a lower final performance (\Cref{sec:stddev}); increasing the number of rollouts may improve stability but leads to a lower final performance (\Cref{sec:rollout}); models require harder and a diverse set of patterns to escape the formatting-only shortcut (\Cref{sec:pattern_heldout}); models trained without a random subset of patterns can still generalize to unseen patterns and larger pattern sizes (\Cref{sec:pattern_heldout}); training with PPO may improve \strictacc{} but hurts \passk{} (\Cref{sec:ppo}).
\section{Related Work}
\label{sec:related-works}

\paragraph{Skill composition in LLMs.} 
\citet{arora2023theory} establish a theoretical foundation for the emergence of skill composition in LLMs, and \citet{yu2024skillmix} benchmark the ability of the models to compose skills. Regarding how to elicit skill composition, previous works focus mainly on pre-training and supervised fine-tuning: \citet{chen2023skillit} propose optimizing data ordering, whereas other works investigate in-context prompting \citep{chen-etal-2024-skills} and supervised fine-tuning \citep{zhao2024can}. This work extends the analysis of skill composition to the RL post-training stage.

\paragraph{Length and compositional generalization.}
Previous works \citep{anil2022exploring,zhou2024transformers,lee2025selfimprovingtransformersovercomeeasytohard} study the ability of LLMs to generalize to a sequence length longer than seen during supervised training. Similar length generalization capability has been studied for RL post-training \citep{tong2026generalization}, albeit sometimes conflated with compositional generalization. \begin{NoHyper}\citet{deepseekai2025deepseekr1incentivizingreasoningcapability,setlur2025e3learningexploreenables}\end{NoHyper} claim that RL helps models generalize to more difficult problems by encouraging more attempts, considering length as a proxy for difficulty. \citet{sun2025omegallmsreasonoutside,zhang2025interplaypretrainingmidtrainingrl,Yuan2025FromFGtoFGx,chu2025sft} study the ``compositional ability'' of RL, but their axis of generalization is limited to the number of skills or compositional depth, which coincides with length. While \citet{cheng2025atomiccompositereinforcementlearning} study true compositional generalization, they do not study length generalization. More broadly, \citet{lake2018generalization} study length and compositional generalization separately, but they do it for seq2seq RNN models. Our work studies length and compositional generalization separately, particularly during RL post-training of transformer-based LLMs.

\paragraph{Tree structure of reasoning.}
In previous works \citep{yehudai2026compositional,jiang2025makes}, the reasoning structure behind a compositional task has been represented as a tree. \citet{opedal2025mathgap} further studies how the tree structure (e.g., width and depth) affects the generalization behavior of LLMs on the corresponding task. However, the analysis of \citet{opedal2025mathgap} is restricted to close-to-binary trees, whereas our work also analyzes how unbalanced trees affect the learning, especially in RL post-training.


\paragraph{Sharpening vs discovery and pass@k.}
Previous works \citep{yue2025does,dang2025weight} often use the metric of pass@k to label the effect of post-training as either \emph{sharpening} \citep{huang2024sharpening}—concentrating probability on pre-existing good paths—or \emph{discovery}—acquiring new knowledge. While useful, trajectory-level metrics like pass@k are too coarse to provide detailed insight into model behavior. Our work focuses on the learning dynamics, which are compatible with both possibilities. Instead of concluding whether RL is introducing a new type of reasoning or resurfacing it, we focus more on skill composition: \emph{how} and \emph{in which order} it happens.

\paragraph{\countdown{} and the game of 24.}
Prior works \citep{yao2023tree,gandhi2024streamsearchsoslearning,herr2025lfs,ni2025learnforget} consider \countdown{}, and its variant Game of 24, as a testbed for model reasoning due to its clean structure. They mainly focus on optimizing inference-time search or training on heuristics-based search traces, 
but these approaches can make it difficult to identify which skills are acquired versus which are a part of the search algorithm / heuristics. By contrast, our work focuses on \emph{natural RL} on \countdown{}---no handcrafted search heuristics or specialized decoding---so the only signal is the natural task reward. This design lets us attribute behavioral changes directly to RL and to analyze them with the pattern framework.

\paragraph{Design choices in RL training.}
RL training is sensitive to algorithmic choices. The loss function in GRPO \citep{shao2024deepseekmathpushinglimitsmathematical} involves multiple terms, including group-level advantage normalization, KL regularization, and gradient clipping. Follow-up works \citep{liu2025understanding,yu2025dapo,shrivastava2025sample} analyze the effect of these choices; for example, arguing that removing standard-deviation normalization can improve stability and avoid biases toward easy prompts. Our work complements these works by providing additional empirical observations on the effect of the design choices in RL.
\section{Discussion and Future Work}

Our work studies the acquisition of skill composition during RL. On natural data, compositional complexity is confounded with length, making it impossible to isolate true failure modes. A synthetic testbed is therefore methodologically essential to control for these confounders and identify drivers of reasoning failure. Thus, the \countdown{} task introduces a formal lens through which to analyze the acquisition of compositional skills. By representing solutions as computational trees, we can disentangle the effects of a problem's length from its underlying compositional structure. This decomposition reveals a fundamental insight: the primary barrier to learning complex reasoning are not length-based but structural. Models do not fail simply because a task is long; they fail when its compositional form creates specific bottlenecks.

Our analysis reveals that problems decomposable into balanced sub-problems are learned most readily. In contrast, those requiring deep-sequential commitments (particularly ``right-heavy'' structures that demand significant lookahead before executing a complex subroutine) constitute a fundamental bottleneck for autoregressive models; crucially, these right-heavy structures are often harder than left-heavy ones, despite having the same compositional depth. This demonstrates that a task's compositional structure, rather than merely its size, is what dictates difficulty. This finding should be understood not as an artifact of our task, but as highlighting a plausible explanation of where difficulty in extending beyond the reasoning boundary arises. Our results suggest that the inherent challenge is tied to the structure of skill composition itself, not to incidental features of the data.

Our preliminary observation on formal theorem proving hints that this structural view of difficulty likely extends to other domains--such as program synthesis or multi-step tool use--where solutions can be represented as computation or proof trees. In such settings, structural properties (balance, depth, and lookahead/sequential dependencies) may predict difficulty. We argue that the overall sample complexity of learning skill composition is not monolithic; instead, it is dominated by the challenge of mastering these structurally-hard instances.

This structural view of difficulty likely extends to other domains--such as program synthesis, multi-step tool use, and theorem proving--where solutions can be represented as computation or proof trees. In such settings, structural properties (balance, depth, and lookahead/sequential dependencies) may predict difficulty. We argue that the overall sample complexity of learning skill composition is not monolithic; instead, it is dominated by the challenge of mastering these structurally-hard instances. 

This insight points to curriculum design as a promising direction for future work. We hypothesize that, once RL post-training yields signs of length and some compositional generalization, dynamically augmenting the training distribution with a minimal set of structurally hard instances may further advance the model’s capability frontier. Verifying this hypothesis and developing such curricula remain a crucial direction for future work.

\section{Limitations}
\label{sec:limitations}
The scope of the paper is primarily on \countdown{}. While we extend our analysis to real-world settings (formal mathematical reasoning), we only do a post-hoc evaluation on an existing RL checkpoint; there may be confounding factors that we were not able to remove. Also, to remove any bias from the dataset distribution, we train on a uniform random mix of a balanced dataset; we do not consider any data curriculum, which may further accelerate the rate of learning or reveal other learning dynamics of RL.
\section{Acknowledgements}
Supported by DARPA (AIQ program, DARPA CMO AWD8860). We thank Dr. Pat Shafto of AIQ Program for useful discussions. SA also acknowledges support from NSF, ONR, and the Schmidt Foundation. Additionally, SP is partially supported by the Gordon Wu Fellowship and the Kwanjeong Educational Foundation Scholarship. We thank Abhishek Panigrahi, Yun Cheng, Xingyu Zhu, Haoyu Zhao, Zixuan Wang, Noam Razin, Sanghyuk Chun, and Howard Yen for helpful discussions and feedback.

\clearpage
\bibliography{reference}
\bibliographystyle{plainnat}


\clearpage
\appendix
\onecolumn
\section{Experimental Setup (More Details)}

\subsection{Expressions to Patterns}
\label{appendix:pattern}
Here are the heuristics (in plain English) to map an expression to its canonical form.

\begin{itemize}
    \item Remove any unnecessary parentheses. For example, map $(A + B) + C$ to $A + B + C$.
    \item If possible, use $+$ and $\times$ instead of $-$ and $/$. For example, map $A - (B - C)$ to $A - B + C$.
    \item For commutative operations ($+$ and $\times$), the operands are sorted in decreasing order of the ``weight'' of the operand (i.e., the number of symbols involved in the intermediate term). For example, $(A + B) \times C$ is preferred over $A \times (B + C)$.
    \item If in the same intermediate level of computation, $+$ and $\times$ should appear before $-$ and $/$. For example, $A + B - C$ is preferred over $A - B + C$. For example, $(A + B) \times (A - B)$ is preferred over $(A - B) \times (A + B)$.
\end{itemize}

The full list of mappings is given in \Cref{table:patterns_list_1,table:patterns_list_2,table:patterns_list_3,table:patterns_list_4}. In \Cref{appendix:alt-patterns}, we show that the choice of the canonicalization scheme does not affect the main findings of the paper.

\subsubsection{A Tree View of Expressions and Patterns}

Each \textbf{expression} can be viewed as a binary tree, with each node representing an operation and each leaf node representing a symbol. To define congruence between equivalent expressions, we convert them to a signed $n$-ary tree:
\begin{itemize}
    \item Each node now represents a chain of addition/subtraction operations ($\oplus$) or a chain of multiplication/division operations ($\otimes$). Each node can now connect to any number of operands that are being computed in that chain. For example, $A + B + C$ is represented with one node with 3 children.
    \item Each edge is assigned a positive sign if it connects to an operand that is being added/multiplied in the chain or a negative sign if connects to an operand that is being subtracted/divided in the chain. 
    \item To convert from a binary tree to a $n$-ary tree:
    \begin{itemize}
        \item For nodes representing addition/multiplication, mark both edges as positive. For nodes representing subtraction/division, mark the left edge as positive but the right as negative.
        \item Recursively merge any pair of parent/child if they belong in the same chain. The sign of the edges of the child node will be flipped if the edge between the parent and the child was negative.
    \end{itemize}
\end{itemize}

If the binary trees have the same representation in the signed $n$-ary tree, they are considered equivalent. The \textbf{pattern} (canonical form) can be retrieved as follows:
\begin{itemize}
    \item At any level, sort the children by 
    \begin{enumerate}
        \item sign of the edge between the shared parent and the child (positive comes first);
        \item the weight of the subtree rooted at the child (heavy comes first);
        \item the number of positive edges to the grandchildren (more positive edges comes first).
    \end{enumerate}
    \item Do an in-order traversal of the tree. Print the first node as $A$, the second node as $B$, and so forth. Include parentheses only if a child subtree has weight $>$ 1 and is rooted in a $\oplus$ node.
\end{itemize}

\begin{figure}[!h]
    \centering
    \includegraphics[width=0.75\linewidth]{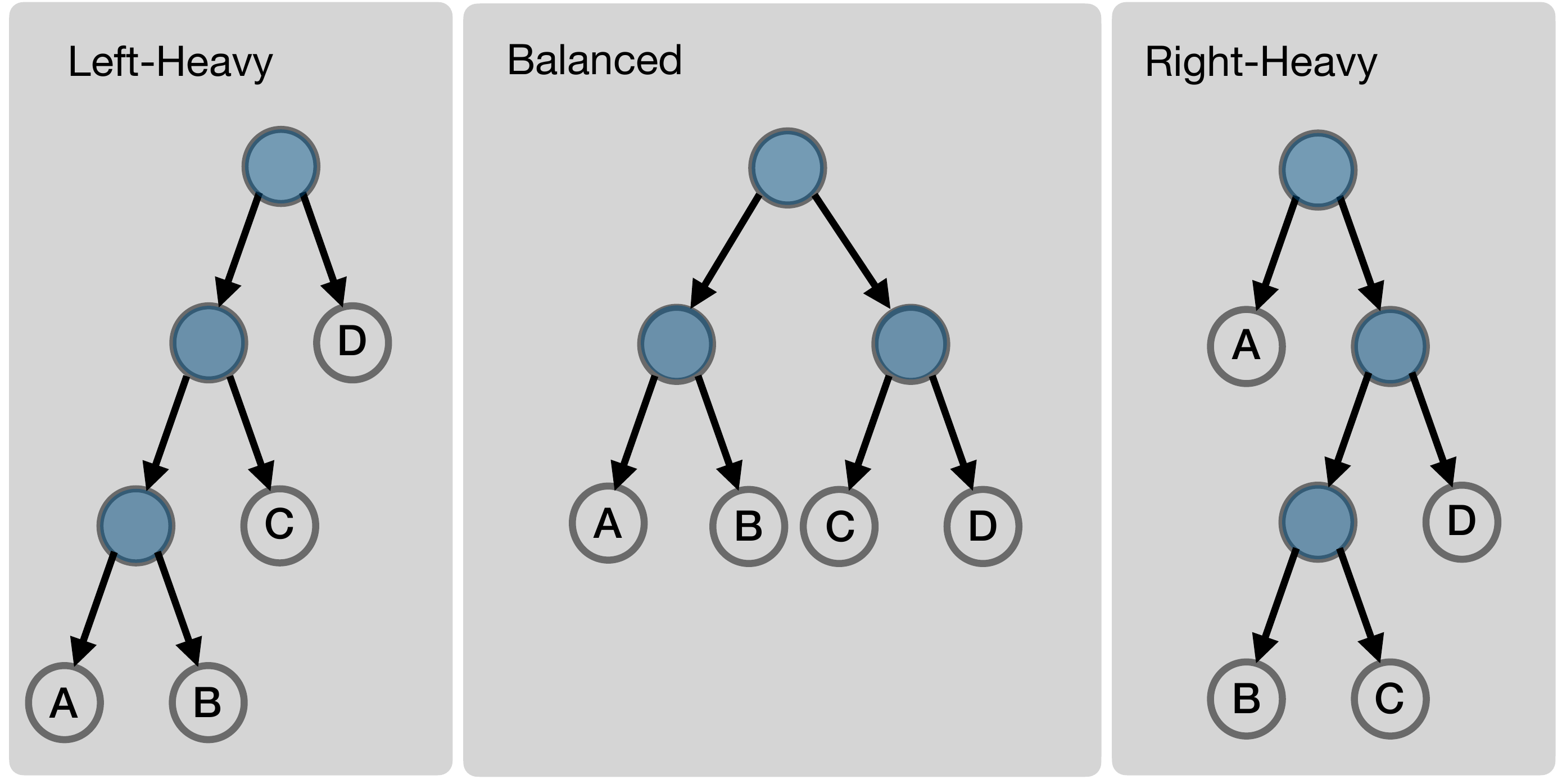}
    \caption{\textbf{Grouping canonical patterns by tree structure ($n=4$).} Internal (blue) nodes are operators and leaves $(A,B,C,D)$ are operands. We categorize patterns by the relative size of the root's subtrees: left-heavy, balanced, or right-heavy.}
    \label{fig:tree-structure}
\end{figure}

\subsubsection{Number of Expressions and Patterns}
Given $n$, the number of distinct \textbf{expressions} is $C_{n - 1} \times 4^{n - 1}$ where $C_n$ is the Catalan number. There are $C_{n - 1}$ ways to assign the order of $n - 1$ operations (equivalently, choose the structure of the binary tree). For example, when $n = 4$, there are $5$ ways: $((AB)C)D$, $(A(BC))D$, $(AB)(CD)$, $A((BC)D)$, and $A(B(CD))$. There are additionally $4^{n-1}$ different ways to choose the operations.

The number of distinct \textbf{patterns} $T_n$ is equal to the number of trees with $n$ leaf nodes with the following constraints: 1) any child of a $\oplus$ node must be a leaf node or a $\otimes$ node (and vice versa); 2) no node shall have all negative edges. We let $A_n, M_n$ respectively denote the number of such trees with a $\oplus$ root node and a $\otimes$ root node. We let $A_1 = 1$ (for the single leaf $A$) and $M_1 = 0$. For $n \geq 2$, $T_n, A_n, M_n$ can be computed with the following recursive definition.
\begin{equation*}
    \begin{cases}
        &A_n = n + \underset{p \in \underset{{2 \leq k < n}}{\bigcup} P(n, k)}{\sum} \left[ (g_1 + 1) \left( \underset{j=1}{\overset{m_p}{\prod}} \binom{g_{p, j} + 2M_{d_{p, j}} - 1}{g_{p, j}} \right) - \left( \underset{j=1}{\overset{m_p}{\prod}} \binom{g_{p, j} + M_{d_{p, j}} - 1}{g_{p, j}} \right) \right] \\
        &M_n = n + \underset{p \in \underset{{2 \leq k < n}}{\bigcup} P(n, k)}{\sum} \left[ (g_1 + 1) \left( \underset{j=1}{\overset{m_p}{\prod}} \binom{g_{p, j} + 2A_{d_{p, j}} - 1}{g_{p, j}} \right) - \left( \underset{j=1}{\overset{m_p}{\prod}} \binom{g_{p, j} + A_{d_{p, j}} - 1}{g_{p, j}} \right) \right] \\
        &T_n = A_n + M_n
    \end{cases}
\end{equation*}
where 
\begin{itemize}
    \item $P(n, k)$ is the unordered set of partitions of $n$ into a sum of $k$ positive integers
    \item Each partition $p \in \underset{{2 \leq k < n}}{\bigcup} P(n, k)$ is written as 
    \begin{equation*}
        p = \overbrace{\left( d_{p, 1} + d_{p, 1} + \cdots + d_{p, 1} \right)}^{g_{p, 1}\text{ times}} + \cdots + \overbrace{\left( d_{p, m_p} + \cdots + d_{p, m_p} \right)}^{g_{p, m_p}\text{ times}} + \overbrace{\left( 1 + \cdots + 1 \right)}^{g_1\text{ times}}
    \end{equation*} where
    \begin{itemize}
        \item $m_p$ is the number of distinct part sizes in $p$ (other than 1)
        \item $d_{p, j} > 1$ are the distinct part sizes in $p$
        \item $g_{p, j}$ are their corresponding multiplicities, and $g_1$ is the multiplicity for $1$
    \end{itemize}
\end{itemize}
The first $n$ corresponds to assigning $\{ 0, 1, \cdots, n-1 \}$ negative edges to $n$ leaf nodes directly connected to the root node (which we assume to be $\oplus$ without loss of generality). Then for any partition $p \in P(n, k)$ where $2 \leq k < n$, any $d_{p, j} > 1$ corresponds to a $\otimes$ child node with weight $d_{p, j}$, whereas $d_{p, j} = 1$ corresponds to a leaf node directly connected. If $d_{p, i} \not = d_{p, j}$, then the corresponding $\otimes$ child nodes are distinguishable, so we can independently count the number of choices for each child node.

For each $d_{p, j} > 1$, there could be multiple $\otimes$ child nodes that have the same weight and can have one of $M_{d_{p_j}}$ shapes. If they have the same shape, they are indistinguishable. For each $i = 1, 2, \cdots, M_{d_{p, j}}$, let $x_i \geq 0$ denote the number of the child nodes that have the corresponding shape such that
\begin{equation*}
    x_1 + x_2 + \cdots + x_{M_{d_{p, j}}} = g_{p, j}
\end{equation*}
For each $x_i$, there are $x_i + 1$ distinguishable ways to assign a sign to each child node ($\{ 0, 1, \cdots, x_i \}$ negative edges). Therefore, we apply the stars and bars formula to get:
\begin{align*}
    &\underset{\substack{x_1 + \cdots + x_{M_{d_{p, j}}} = g_{p, j} \\ x_i \geq 0 \; \forall i}}{\sum} \quad (x_1 + 1)(x_2 + 1) \cdots (x_{M_{d_{p, j}}} + 1) \\
    &= \underset{\substack{y_1 + \cdots + y_{M_{d_{p, j}}} = g_{p, j} + M_{d_{p, j}} \\ y_i \geq 1 \; \forall i}}{\sum} \quad y_1 y_2 \cdots y_{M_{d_{p, j}}} \\
    &= \binom{(g_{p, j} + M_{d_{p, j}}) + M_{d_{p, j}} - 1}{2M_{d_{p, j}} - 1} \\
    &= \binom{g_{p, j} + 2M_{d_{p, j}} - 1}{g_{p, j}}
\end{align*}
When $d = 1$, the leaf children are all indistinguishable and there are $g_1 + 1$ ways to assign a sign to each leaf. We finally need to subtract the case where we assigned a negative edge to all possible $g_{p, j}$ child nodes of all possible weight of $d_{p, j}$. For each choice of $x_i$, there is exactly 1 way to assign a negative sign to all child nodes. Again, by the stars and bars formula, we get
\begin{equation*}
    \underset{\substack{x_1 + \cdots + x_{M_{d_{p, j}}} = g_{p, j} \\ x_i \geq 0 \; \forall i}}{\sum} \quad 1 = \binom{g_{p, j} + M_{d_{p, j}} - 1}{g_{p, j}}
\end{equation*}

\subsection{Dataset}
\label{appendix:dataset}

\subsubsection{Why is the Existing Dataset Lacking?}
\begin{figure}[!h]
    \centering
    \includegraphics[width=0.95\linewidth]{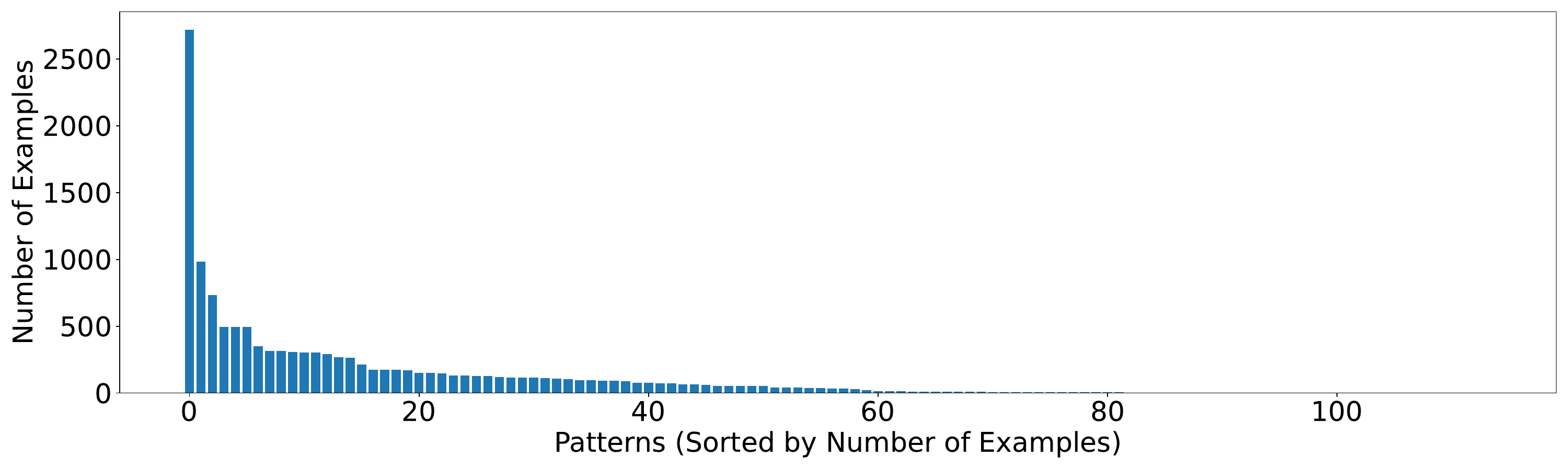}
    \caption{\textbf{Number of examples per pattern in \citet{tinyzero}.} 18 out of 114 patterns do not appear in a randomly selected 10000 examples from the dataset. Few patterns (mostly from $n=3$) appear disproportionately frequently. }
    \label{fig:analyze_unbalanced}
\end{figure}

\citet{tinyzero} do not release the exact generation code for their dataset. So instead, we sample 10000 examples from the training split of the dataset and count the number of examples that can be solved with a given pattern (if one example can be solved with multiple patterns, increment the count of all such patterns). 18 of the 114 patterns never appear in the dataset (i.e., none of the examples can be solved with the patterns), and some patterns occur much more often than others (\Cref{fig:analyze_unbalanced}). In the test split (the last 1000 examples), there are 36 out of 114 patterns that never appear. 

One reason behind the imbalance is that almost half of the dataset consists of examples for $n=3$, even though there are only $18$ patterns for $n=3$ compared to $96$ for $n=4$. Another explanation is that the requirement that the target number is an integer within $[1, 99]$ biases towards addition and subtraction---multiplication would frequently cause the expression to be larger than the bound; division would frequently make the final value non-integer.

Two other codebases that provide the generation code for Countdown \citep{gandhi2024streamsearchsoslearning,stojanovski2025reasoninggymreasoningenvironments} wrongly assume that intermediate expressions cannot evaluate to non-integer values. This incorrectly removes puzzles like $8 \div (3 - 8 \div 3) = 24$ which are very difficult even for humans \citep{4nums} and should be used to test against models.

\subsubsection{What Happens When Training on the Existing Dataset?}
\begin{figure}[h]
  \centering
  \includegraphics[width=0.48\linewidth]{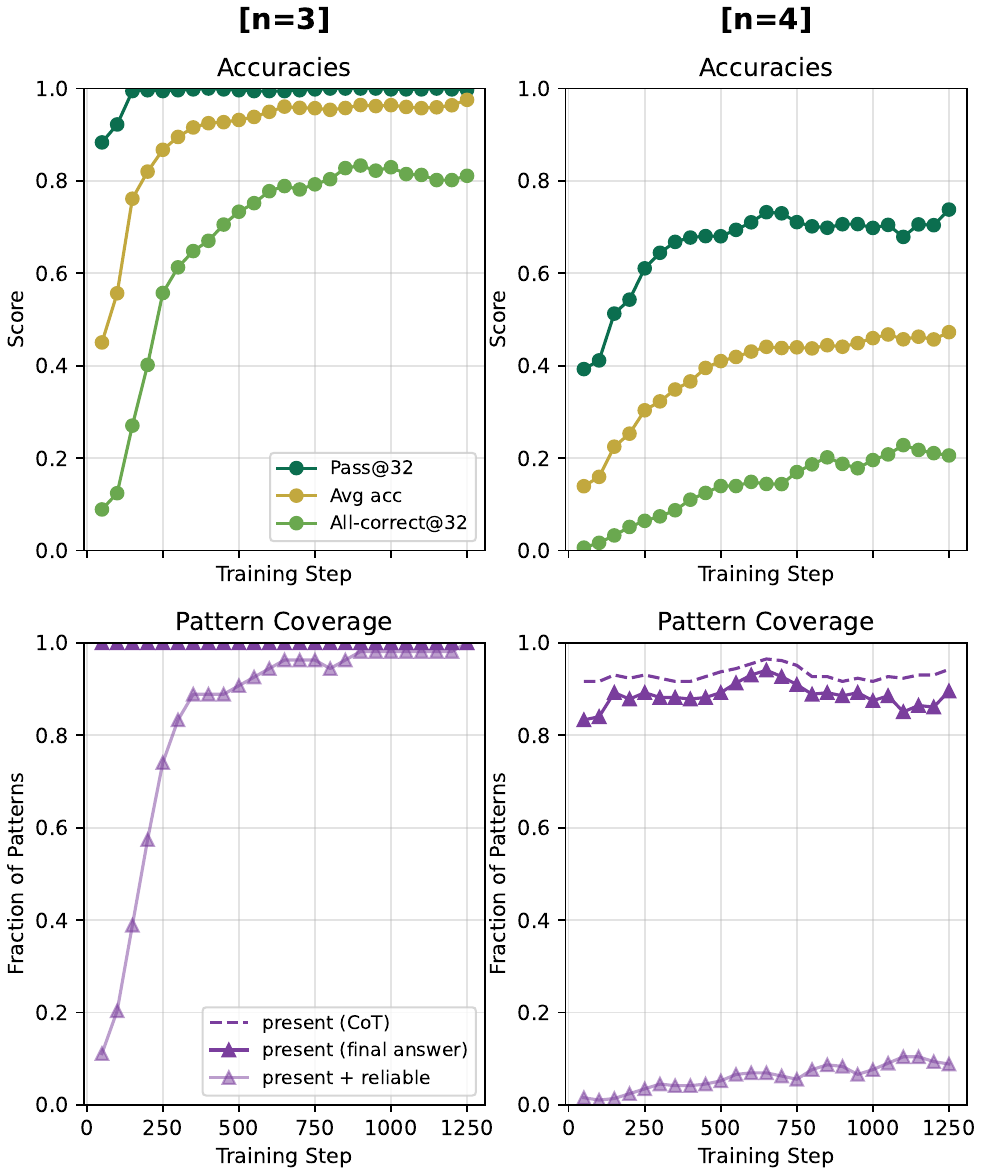}
  \hfill
  \includegraphics[width=0.48\linewidth]{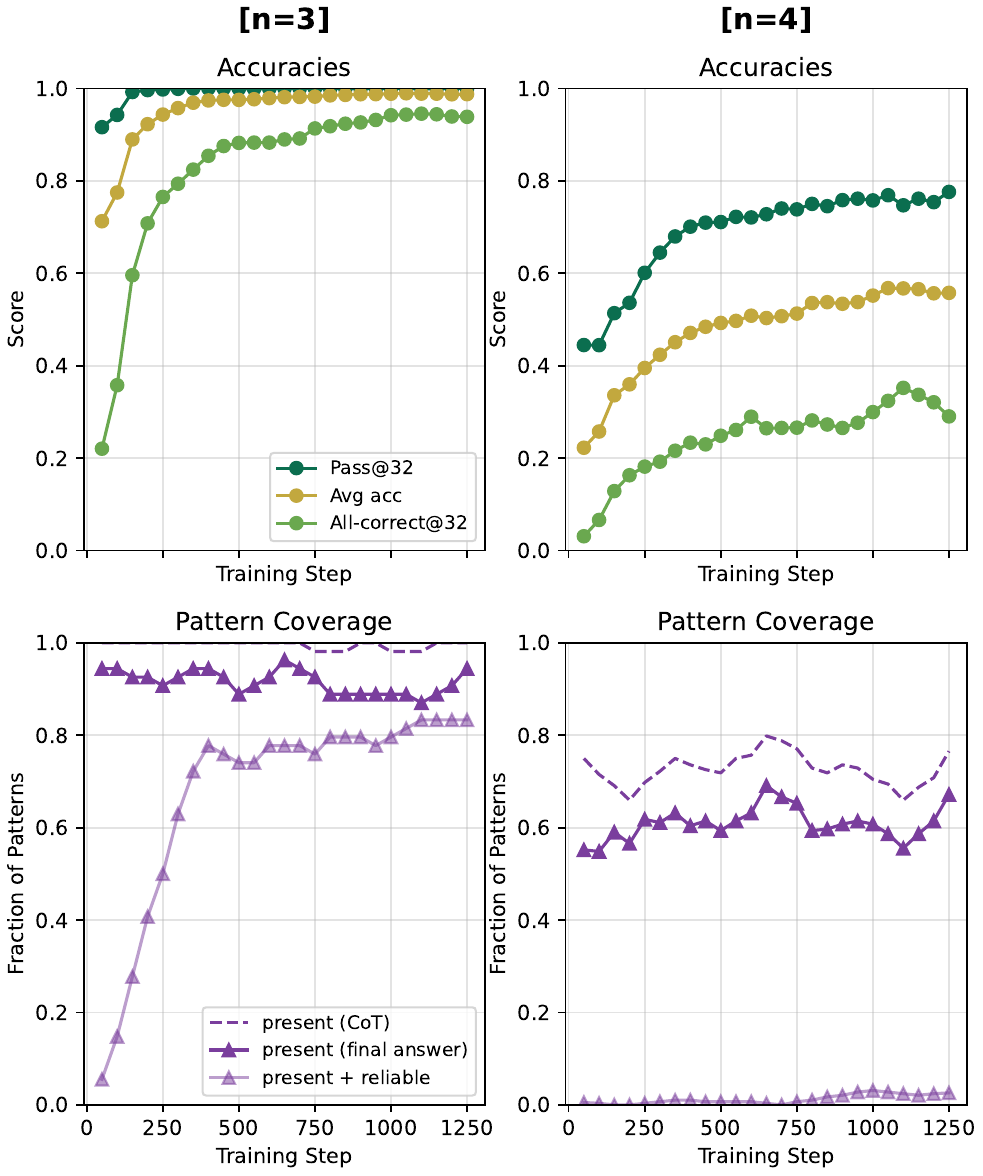}
  \caption{\textbf{Learning dynamics at a glance (Qwen2.5-3B)} trained on dataset from \citet{pan2025learning}. \textbf{(Left)} evaluated on our dataset; \textbf{(Right)} evaluated on the held-out data from \citet{pan2025learning}. Even though model accuracy smoothly increases (top), models do not reliably learn most $n=4$ patterns and fails to generate at least 10\% of $n=4$ patterns (bottom). The held-out data from \citet{pan2025learning} do not include 36 possible patterns and fail to test the models on the corresponding skills.}
  \label{fig:subtrees-qwen2.5-3b-tinyzero}
\end{figure}

We train \texttt{Qwen2.5-3B} on the dataset by \citet{pan2025learning} with the same set of hyperparameters and evaluate on both our and their held-out data. Since we do not explicitly check if our held-out data is in their training data, the results on our held-out data may be slightly higher than expected. In \Cref{fig:subtrees-qwen2.5-3b-tinyzero}, we recreate \Cref{fig:learning-dynamics}. 

Even though the model performance seems to smoothly improve, the final performance is much lower than training on our balanced dataset, which stems from the model failing to reliably learn most $n=4$ patterns. This suggests that the diversity of the data and the presence of challenging patterns helps the model identify reusable components. Additionally, many patterns are not present when evaluating on the held-out data from \citet{pan2025learning}, precisely because none of the puzzles test those patterns. This points to another deficiency in the existing datasets for \countdown{} that test simple additive skills, rather than a well-rounded portfolio of arithmetic reasoning.

\subsubsection{Creating a Balanced Dataset}
\label{sec:generating-balanced-dataset}
To create our balanced dataset, we first choose a pattern and then plug in $n$ numbers that are sampled independently and uniformly at random from $[1, 99]$. If the output of the expression is an integer in $[1, 99]$, we include it in our dataset. \textit{For each pattern}, we repeat this process until we have 4000 distinct examples or have sampled 10 million combinations. There are only $15$ patterns ($4$ from $n=3$ and $11$ from $n=4$) where we sample less than 4000 examples. The exact numbers are given in \Cref{table:balanced_num_examples}. Note that only 2 among the undersampled $15$ patterns are right-heavy, and removing the 15 undersampled patterns entirely from the analysis does not change the result of \Cref{sec:results:structure-alt}. In \Cref{fig:subtrees-qwen2.5-1.5b-remove-undersampled}, we plot \Cref{fig:subtrees} again but after removing the undersampled patterns from the analysis.

For each pattern, we select the first 10 examples as heldout examples. In total, we have 418619 examples for training and 1140 examples for testing. We modify the codebase from \citet{tinyzero}. 

For evaluation on $n = 5$, we similarly sample 10 examples for each of 558 patterns and for $n = 6$, 1 example for each of 3494 patterns.

\begin{table}[h!]
\caption{\textbf{Number of training examples per pattern.} Any pattern not listed has 4000 examples.}
\label{table:balanced_num_examples}
\centering
\begin{tabular}{|l|l|l|l|}
\toprule
Pattern & Number of Examples & Pattern & Number of Examples \\ \hline
$(A+B) \times C$ & 3333 & $A \times B \times C$ & 315  \\ \hline
$A/(B+C)$ & 3905 & $A/B/C$ & 791  \\ \hline
$(A \times B+C) \times D$ & 2314 & $(A+B) \times C \times D$ & 927  \\ \hline
$A \times B \times C \times D$ & 188 & $A/(B \times C+D)$ & 2704  \\ \hline
$A/(B+C)/D$ & 1058 & $A/B/C/D$ & 264  \\ \hline
$A/B-C \times D$ & 2546 & $(A+B) \times (C+D)$ & 647  \\ \hline
$A/(B+C)-D$ & 932 & $A/B/C-D$ & 1033  \\ \hline
$(A/B-C)/D$ & 2802 & - & -  \\ \bottomrule
\end{tabular}
\end{table}

\subsection{Training}
\label{appendix:training}
We train with batch size and mini batch size 256, prompt and response length 1024, learning rate $10^{-6}$, and KL coefficient $10^{-3}$, where the KL divergence is computed with the $k_3$ estimator (low variance) \citep{schulman2020kl}. Following \citet{liu2025understanding}, we do not normalize by the standard deviation in the GRPO advantage computation. We use $n=4$ rollouts except during a warmup phase for 1.5B and 3B models (\Cref{sec:reward_hacking}). For the PPO experiments in \Cref{sec:ppo}, we use a critic learning rate of $10^{-5}$ and PPO minibatch size of 64. For all other hyperparameters, we use the default values from the \texttt{verl} package version 0.5.0.dev0 \citep{sheng2024hybridflow}. See \Cref{prompt-rl} for the prompt template used. Training the 1.5B model for 1 epoch takes roughly 8 hours on 4 H100 GPUs. \clearpage
\section{Main Results (Continued)}

\subsection{Additional Tables}
\label{appendix:alt-tables}
1.5B/3B models smoothly improve performance throughout training (1635 steps), whereas the 7B model reaches peak performance before 1 epoch. In the main paper, we reported the performance of the checkpoints right before the training reward declines (1350/950/650 steps respectively for each seed). 
For completeness, we also provide the performance of the final checkpoints of the 7B model in \Cref{tab:final-accuracy-final,tab:root-and-subtree-final}. The 7B model can eventually improve on right-heavy patterns ($[1] - [3]$ or $[1] \div [3]$) but at the cost of sacrificing performance on balanced and left-heavy patterns. Additionally, we report the standard deviation when averaging across random seeds. 

In \Cref{tab:final-accuracy-final}, we also report the performance of the base models before GRPO training. Note that the \passk{} for these models is very close to $1 - (1 - p)^{32}$ where $p = $ \avgacc{}, suggesting that the models are randomly guessing the output. At the same time, within each output, the model tends to randomly guess multiple expressions, until it runs out of the generation budget. Since there is only a small subset of expressions that are valid for $n = 3, 4$ (up to $18 \times 3! = 108$ for $n=3$ and $96 \times 4! = 2304$), the random baseline for the base models is quite high (assuming 10 guesses per output, $p \approx 0.1$ for $n=3$ and $p \approx 0.05$ for $n=4$). The observed performance of the base models is not far from the random baseline.

\begin{table}[h]
  \centering
  \caption{\textbf{Performance across model size (Full version).}
  Averaged across all successful seeds (\textbf{final} (F) checkpoints for 1.5B/3B and \textbf{best} (B) and \textbf{final} (F) checkpoints for 7B). \strictacc{} (All), \avgacc{} (Avg), and \passk{} (Pass) on $n{=}3,4$ (trained) and $n{=}5,6$ (held out). Models learn $n=3, 4$ patterns almost perfectly and generalize to larger puzzle sizes $n=5, 6$, but the gap between \passk{} and \avgacc{} increases with $n$, signaling that the generalization is not perfect.}
  \label{tab:final-accuracy-final}
  \begin{tabular}{l *{2}{lll}}
    \toprule
    Model & \multicolumn{1}{c}{All} & \multicolumn{1}{c}{Avg} & \multicolumn{1}{c}{Pass} & \multicolumn{1}{c}{All} & \multicolumn{1}{c}{Avg} & \multicolumn{1}{c}{Pass} \\
    \midrule
    & \multicolumn{3}{c}{$n{=}3$} & \multicolumn{3}{c}{$n{=}4$} \\
    \cmidrule(lr){2-4}\cmidrule(lr){5-7}
    1.5B & 0.00 & 0.02 & 0.42 & 0.00 & 0.00 & 0.00 \\
    1.5B (F) & 0.97 $\pm$ 0.01 & 1.00 $\pm$ 0.00 & 1.00 $\pm$ 0.00 & 0.60 $\pm$ 0.09 & 0.82 $\pm$ 0.02 & 0.93 $\pm$ 0.00 \\
    \midrule
    3B & 0.00 & 0.05 & 0.66 & 0.00 & 0.01 & 0.20 \\
    3B (F) & 0.93 $\pm$ 0.01 & 0.99 $\pm$ 0.01 & 1.00 $\pm$ 0.00 & 0.56 $\pm$ 0.06 & 0.81 $\pm$ 0.03 & 0.94 $\pm$ 0.01 \\
    \midrule
    7B & 0.00 & 0.10 & 0.88 & 0.00 & 0.02 & 0.36 \\
    7B (B) & 0.99 $\pm$ 0.01 & 1.00 $\pm$ 0.00 & 1.00 $\pm$ 0.00 & 0.61 $\pm$ 0.07 & 0.86 $\pm$ 0.03 & 0.98 $\pm$ 0.01 \\
    7B (F)   & 0.95 $\pm$ 0.02 & 0.99 $\pm$ 0.00 & 1.00 $\pm$ 0.00 & 0.56 $\pm$ 0.03 & 0.80 $\pm$ 0.06 & 0.92 $\pm$ 0.06 \\

    \midrule    
    & \multicolumn{3}{c}{$n{=}5$} & \multicolumn{3}{c}{$n{=}6$} \\
    \cmidrule(lr){2-4}\cmidrule(lr){5-7}
    1.5B & 0.00 & 0.00 & 0.02 & 0.00 & 0.00 & 0.01 \\
    1.5B (F) & 0.13 $\pm$ 0.05 & 0.36 $\pm$ 0.03 & 0.68 $\pm$ 0.00 & 0.00 $\pm$ 0.00 & 0.10 $\pm$ 0.00 & 0.41 $\pm$ 0.00 \\
    \midrule
    3B & 0.00 & 0.00 & 0.03 & 0.00 & 0.00 & 0.00 \\
    3B (F) & 0.12 $\pm$ 0.02 & 0.39 $\pm$ 0.01 & 0.68 $\pm$ 0.02 & 0.02 $\pm$ 0.00 & 0.18 $\pm$ 0.00 & 0.45 $\pm$ 0.00 \\
    \midrule
    7B & 0.00 & 0.00 & 0.08 & 0.00 & 0.00 & 0.02 \\
    7B (B) & 0.15 $\pm$ 0.02 & 0.46 $\pm$ 0.04 & 0.75 $\pm$ 0.03 & 0.02 $\pm$ 0.00 & 0.17 $\pm$ 0.00 & 0.45 $\pm$ 0.00 \\
    7B (F) & 0.10 $\pm$ 0.02 & 0.35 $\pm$ 0.05 & 0.61 $\pm$ 0.10 & 0.01 $\pm$ 0.00 & 0.14 $\pm$ 0.00 & 0.41 $\pm$ 0.00 \\
    \bottomrule
    
  \end{tabular}
\end{table}
\clearpage
\begin{table}[h]
  \centering
  \caption{\textbf{Per-pattern precision averaged by pattern structure on $n{=}4$ (Full version).}
  Averaged across all successful seeds (\textbf{final} checkpoints for 1.5B/3B and \textbf{best} and \textbf{final} checkpoints for 7B). Averaged by (a) root operator; (b) pattern with root operator $-$; (c) pattern with root operator $\div$. Models struggle more on $-, \div$ as the root operator. Shapes in the form of $[1] - [3]$ or $[1] \div [3]$ are particularly challenging. Within each column, the highest number is \textbf{boldfaced} and the second highest number is \underline{underlined}.}
  \label{tab:root-and-subtree-final}
  \setlength{\tabcolsep}{6pt}

  \begin{subtable}[t]{\linewidth}
    \centering
    \caption{by root operator} 
    \begin{tabular}{@{}l c c c c@{}}
      \toprule
      \textbf{Operator} & \textbf{1.5B} & \textbf{3B} & \textbf{7B} (best) & \textbf{7B} (final) \\
      \midrule
      $+$      & \underline{0.86} $\pm$ 0.08 & \underline{0.85} $\pm$ 0.01 & \textbf{0.91} $\pm$ 0.01 & \underline{0.83} $\pm$ 0.03 \\
      $-$      & 0.83 $\pm$ 0.09 & 0.80 $\pm$ 0.04 & 0.86 $\pm$ 0.06 & 0.76 $\pm$ 0.11 \\
      $\times$ & \textbf{0.90} $\pm$ 0.01 & \textbf{0.86} $\pm$ 0.03 & \textbf{0.91} $\pm$ 0.01 & \textbf{0.84} $\pm$ 0.08 \\
      $\div$   & 0.75 $\pm$ 0.04 & 0.78 $\pm$ 0.06 & 0.83 $\pm$ 0.04 & 0.81 $\pm$ 0.08 \\
      \bottomrule
    \end{tabular}
  \end{subtable}\hfill \vspace{0.1in}
  \begin{subtable}[t]{\linewidth}
    \centering
    \caption{by subtree shape ($-$)}
    \begin{tabular}{@{}l c c c c@{}}
      \toprule
      \textbf{Shape} & \textbf{1.5B} & \textbf{3B} & \textbf{7B} (best) & \textbf{7B} (final) \\
      \midrule
      $\fbox{\scriptsize 3}\!-\!\fbox{\scriptsize 1}$ & \underline{0.81} $\pm$ 0.13 & \underline{0.78} $\pm$ 0.03 & 0.84 $\pm$ 0.04  & 0.72 $\pm$ 0.12 \\
      $\fbox{\scriptsize 2}\!-\!\fbox{\scriptsize 2}$ & \textbf{0.92} $\pm$ 0.01 & \textbf{0.89} $\pm$ 0.00 & \textbf{0.90} $\pm$ 0.07 & \textbf{0.86} $\pm$ 0.10\\
      $\fbox{\scriptsize 1}\!-\!\fbox{\scriptsize 3}$ & 0.73 $\pm$ 0.00 & 0.76 $\pm$ 0.09 & \underline{0.86} $\pm$ 0.12 & \underline{0.82} $\pm$ 0.10 \\
      \bottomrule
    \end{tabular}
  \end{subtable}\hfill \vspace{0.1in}
  \begin{subtable}[t]{\linewidth}
    \centering
    \caption{by subtree shape ($\div$)}
    \begin{tabular}{@{}l c c c c@{}}
      \toprule
      \textbf{Shape} & \textbf{1.5B} & \textbf{3B} & \textbf{7B} (best) & \textbf{7B} (final) \\
      \midrule
      $\fbox{\scriptsize 3}\!\div\!\fbox{\scriptsize 1}$ & \underline{0.77} $\pm$ 0.05 & \underline{0.77} $\pm$ 0.05 & \underline{0.83} $\pm$ 0.05 & 0.83 $\pm$ 0.05 \\
      $\fbox{\scriptsize 2}\!\div\!\fbox{\scriptsize 2}$ & \textbf{0.88} $\pm$ 0.10 & \textbf{0.85} $\pm$ 0.00 & \textbf{0.91} $\pm$ 0.07 & \textbf{0.89} $\pm$ 0.14 \\
      $\fbox{\scriptsize 1}\!\div\!\fbox{\scriptsize 3}$ & 0.61 $\pm$ 0.03 & 0.76 $\pm$ 0.12 & 0.80 $\pm$ 0.17 & 0.80 $\pm$ 0.05 \\
      \bottomrule
    \end{tabular}
  \end{subtable}
\end{table}
\clearpage

\subsection{Effect of Held-out Patterns}
\label{sec:pattern_heldout}

\subsubsection{Generalization to Held-Out Patterns (More Details)}
\label{sec:pattern_heldout_details}
Here, we continue the discussion from \Cref{sec:results:generalization-heldout-tree-alt}. When we remove the pattern $A/B+C$, we must also remove all patterns which directly include $A/B+C$ as a subpattern, or patterns that can collapse into $A/B+C$, once some subtasks have been solved. See \Cref{table:heldout-tree3-list} for the full list.

\begin{table}[h!]
\caption{\textbf{The list of patterns removed, when holding out $A/B+C$.} We additionally remove 6 patterns that contain $A/B+C$ as a subpattern and 8 more that can be collapsed into $A/B+C$ once a subtask has been solved. }
\label{table:heldout-tree3-list}
\centering
\begin{tabular}{|l|l|l|l|l|}
\toprule
Pattern & Explanation & Canonical Pattern & Structure \\ \hline
$A/B+C$ & Itself & $A/B+C$ & $[2] + [1]$\\
$(A/B + C) + D$ & Direct subpattern & $A/B + C + D$ & $[3] + [1]$ \\
$(A/B + C) - D$ & Direct subpattern & $A/B + C - D$ & $[3] - [1]$ \\
$(A/B + C) \times D$ & Direct subpattern & $(A/B + C) \times D$& $[3] \times [1]$ \\
$(A/B + C) / D$ & Direct subpattern & $(A/B + C) / D$ & $[3] \div [1]$\\
$A-(B/C+D)$ & Direct subpattern & $A-B/C-D$ & $[1] - [3]$\\
$A/(B/C+D)$ & Direct subpattern & $A/(B/C+D)$ & $[1] \div [3]$ \\
\midrule
$A/B + (C \times D)$ & Combine $C \times D$ & $A \times B + C / D$ & $[2] + [2]$\\
$A/B + (C/D)$ & Combine $C / D$ & $A / B + C / D$ & $[2] + [2]$\\
$A/(B+C) + D$ & Combine $B+C$ & $A/(B+C) + D$ & $[3] + [1]$\\
$A/(B-C) + D$ & Combine $B-C$ & $A/(B-C) + D$ & $[3] + [1]$\\
$(A+B)/C + D$ & Combine $A+B$ & $(A+B)/C + D$ & $[3] + [1]$ \\
$(A-B)/C + D$ & Combine $A-B$ & $(A-B)/C + D$ & $[3] + [1]$\\
$(A \times B)/C + D$ & Combine $A \times B$ & $A \times B / C + D$ & $[3] + [1]$ \\
$(A/B)/C + D$ & Combine $A/B$ & $A/B/C+D$ & $[3] + [1]$ \\
\bottomrule
\end{tabular}
\end{table}

Next, we test how the RL training affects the model's skills on the heldout patterns. We let 1) the base model; and 2) the final RL checkpoint generate $k=1024$ responses each with a limit of 2048 tokens to every problem in the test set. \footnote{Note that this is 512x the budget we allocate during training, where we generate $k=4$ rollouts each with 1024 tokens.} \Cref{table:heldout-tree3-rollout} shows the number of occurrences of each of the heldout patterns used in the puzzle 1) for the correct target pattern; and 2) across the entire test set. Before the RL training, the base model does not have the skill (or if it does, it has a very weak skill) to decide when to use the heldout patterns. On the other hand, after the RL training, even though the model never received any training signal on the heldout patterns, it learned the skills to identify when to apply the patterns.

\begin{table}[h!]
\caption{\textbf{The number of occurrences of heldout patterns before (\textit{base}) and after (\textit{RL}) training.} \textit{target} refers to the number of occurrences of heldout patterns when they are correctly used on the puzzles that require the usage of the particular pattern, where as \textit{global} refers to the number of occurrences across all puzzles in the test data. The base model rarely produces the heldout patterns to the correct target puzzles, whereas the final RL checkpoint is able to, even though it did not receive any training signal on the heldout patterns.}
\label{table:heldout-tree3-rollout}
\centering 
\begin{tabular}{|l|l|l|l|l|}
\toprule
Pattern & base target & base global & RL target & RL global \\ \hline
$A/B+C$ & 127 & 1,335 & 5,631 & 5,705  \\
$A/B + C + D$ & 13 & 677 & 1,318 & 4,871 \\
$A/B + C - D$ & 99 & 7,094 & 1,457 & 15,368 \\
$(A/B + C) \times D$ & 4 & 140 & 147 & 2,119 \\
$(A/B + C) / D$ & 0 & 29 & 21 & 582 \\
$A-B/C-D$ & 27 & 1,782 & 5,078 & 12,831 \\
$A/(B/C+D)$ & 0 & 19 & 438 & 3,918 \\
$A \times B + C / D$ & 32 & 919 & 6,570 & 9,728 \\
$A / B + C / D$ & 24 & 914 & 7,159 & 9,897 \\
$A/(B+C)+D$ & 0 & 18 & 619 & 2,445 \\
$A/(B-C)+D$ & 1 & 96 & 73 & 2,966 \\
$(A+B)/C+D$ & 27 & 1,422 & 294 & 4,468 \\
$(A-B)/C+D$ & 107 & 11,018 & 269 & 6,358 \\
$A \times B / C + D$ & 49 & 3,126 & 1,317 & 6,386 \\
$A/B/C+D$ & 3 & 144 & 543 & 1,056 \\
\bottomrule
\end{tabular}
\end{table}

\subsubsection{What Patterns are Necessary for the Model to Learn?}
We further explore the effect of heldout patterns during training. We train the \texttt{Qwen2.5-1.5B} model with the following heldout strategies:
\begin{itemize}
    \item Only 3 / Only 4: Train only with patterns from specific puzzle sizes.
    \item No Hard: Train without patterns of shape $[1] - [3]$ or $[1] \div [3]$ (\cref{sec:results:structure-alt})
    \item Random $p$: Train only with randomly selected $p$\% of the patterns 
\end{itemize}

Following \cref{sec:reward_hacking}, we warm-start the training with a larger group size $G=16$ and check how many runs are successful out of 3; i.e., model escapes the formatting-only shortcut after something ``clicks'' (\cref{tab:heldout-phase-transition}). When hard patterns are not a part of the training (``Only 3'' or ``No hard''), the model does not have access to learn from high-entropy, high-reward rollouts and fail to learn. However, training only with $n=4$ patterns still leads to successful training. Additionally, the diversity of the patterns also seems to be helpful for the model --- the probability of a successful run increases with the number of unique patterns in the dataset.

\begin{table}[ht]
    \centering
    \caption{\textbf{Number of successful runs (out of 3) per heldout strategy:} The existence of harder patterns and an overall wide range of patterns seems necessary for the 1.5B model to learn.}
    \begin{tabular}{lccc|ccc}
        \\ \toprule
         & Only 3 & Only 4 & No Hard & Random 30 & Random 40 & Random 50 \\
        \midrule
        \# Successful Runs & 0 & 3 & 0 & 1 & 2 & 3 \\
        \bottomrule \\
    \end{tabular}
    \label{tab:heldout-phase-transition}
\end{table}

\subsubsection{Models Can Generalize to Unseen Patterns (Random 50)}
Even when the model is trained with only a randomly selected 50\% of the patterns, more than 95\% of the patterns are \texttt{present} in the final checkpoint, and the model scores near 70\% average accuracy. The 3B model is able to generalize to more patterns, whereas 4\% of patterns are \texttt{Absent} from the reasoning of the 1.5B model and another 4\% are always applied incorrectly (\Cref{tab:heldout-generalization}). Additionally, the 1.5B model is able to generalize to $n=5$ patterns, but the strength of the generalization is weaker than having been trained on the full set of patterns (\Cref{tab:final-accuracy}). 

\begin{table}[ht]
    \centering
    \caption{\textbf{Final performance of models trained without random 50\% of patterns:} \strictacc{} (All@), \avgacc{} (Avg), and \passk{} (Pass@) of models. Models can generalize to most unseen patterns. 3B model is able to generalize to more diverse patterns than 1.5B, but has a lower \strictacc{}.}
    \begin{tabular}{lccc|cc|ccc}
        \\ \toprule
        & \multicolumn{5}{c}{$n{=}4$} & \multicolumn{3}{c}{$n{=}5$} \\
        \cmidrule(lr){2-6}\cmidrule(lr){7-9}
        & All@ & Avg & Pass@ & Absent (\%) & Incorrect (\%) & All@ & Avg & Pass@ \\
        \midrule
        1.5B & 0.47 & 0.68 & 0.82 & 4 & 4 & 0.06 & 0.25 & 0.52 \\
        3B & 0.35 & 0.69 & 0.88 & 1 & 2 & - & - & - \\
        \bottomrule \\
    \end{tabular}
    \label{tab:heldout-generalization}
\end{table}

\clearpage

\subsection{Prefix Notation Alleviates the Lookahead Bottleneck}
\label{sec:prefix}
The lookahead bottleneck we identify in \Cref{sec:results:structure-alt,sec:results:right-heavy-alt} assumes that the puzzles in \countdown{} are in an infix notation, where an operator is between the operands. We hypothesize that right-heavy patterns are particularly difficult (especially in comparison to left-heavy patterns) because generating the correct operator requires planning ahead for the longer right operand. 

In this section, we now consider a prefix notation, where an operator is emitted before both operands. Under this notation, generating the correct operator requires planning ahead for \textit{both} operands, regardless of whether the pattern is left-heavy or right-heavy. We train a \texttt{Qwen2.5-3B} model with the exact same dataset as in the main text, but we guide the model to use the desired functions (add, sub, mul, div) with the following system prompt:

\begin{tcolorbox}[title=Prompt Template for Prefix Notation] \label{prompt-prefix}
A conversation between User and Assistant. The user asks a question, and the Assistant solves it. The assistant first thinks about the reasoning process in the mind and then provides the user with the answer.

User: Using the numbers {numbers}, create an equation that equals {target}. You must use the functions add(x,y), sub(x,y), mul(x,y), div(x,y) and each number can only be used once. Show your work in <think> </think> tags. And return the final answer in <answer> </answer> tags, for example <answer> div(add(1, 2), 3) </answer>.

Assistant: Let me solve this step by step.

<think>
\end{tcolorbox}

\begin{figure}[h]
  \centering
  \includegraphics[width=0.85\linewidth]{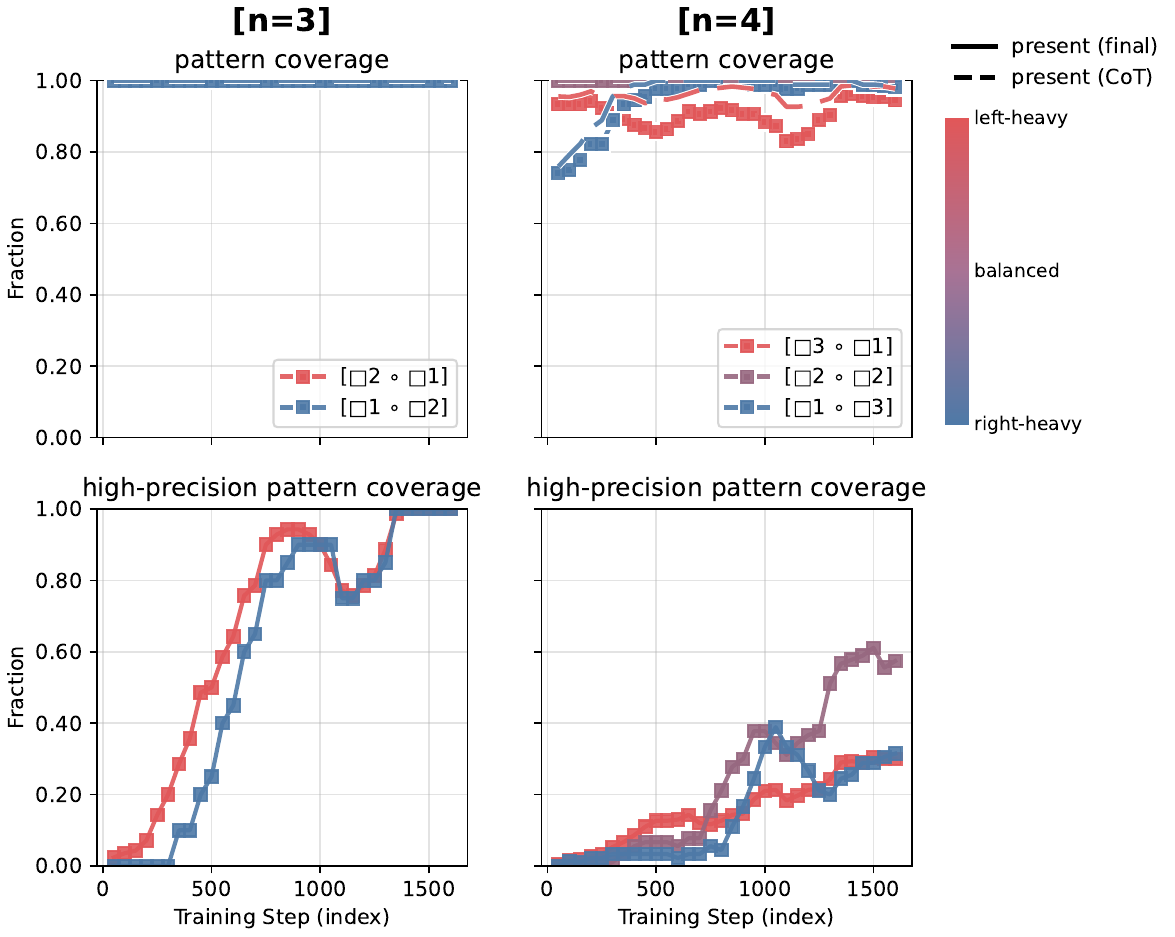}
  \caption{\textbf{Learning dynamics at a glance (Qwen2.5-3B, prefix notation)}. We see a different hierarchy of difficulty by subtree structure:  balanced $<$ left-heavy $\approx$ right-heavy. This provides additional evidence for our hypothesis that lookahead bottleneck causes right-heavy patterns to be more difficult for models to learn.}
  \label{fig:subtrees-qwen2.5-3b-prefix}
\end{figure}

\Cref{fig:subtrees-qwen2.5-3b-prefix} shows that balanced patterns (shallower depth for a fixed size $n$) are still easier to learn, whereas the gap between left-heavy and right-heavy patterns mostly disappears. This provides additional support for our hypothesis that the lookahead bottleneck is the primary reason why right-heavy patterns are harder for models to learn.

\clearpage
\subsection{Isolating the Effect of RL}
\label{sec:sft}

To test whether the effect of the composition structure on the speed of skill acquisition is an artifact of the task or an effect of RL post-training, we additionally compare with supervised finetuning (SFT). To prepare the answer column, we replace the placeholder symbols in the desired pattern with the correct list of numbers. For the optional chain of thought text, we 1) randomly select an integer $k \in [2, n]$ based on the puzzle size $n$; 2) for each of $k$ incorrect attempts, randomly choose a pattern different from the desired pattern and replace the placeholder symbols with a randomly shuffled list of the input numbers; 3) evaluate the expression and append ``\texttt{(Incorrect)}''; 4) append the correct expression and append ``\texttt{(Correct)}.'' The CoT is explicitly designed to be random to remove any bias from hand-coded heuristics. See \Cref{prompt-sft} for example prompts. 

\paragraph{Final accuracies.} \Cref{tab:sft} reports the final accuracies of \texttt{Qwen2.5-1.5B} trained with SFT. The model performs better without the CoT, most likely because including the random patterns confuses the model. Note that these results are only provided as reference and are not a core part of the analysis. Instead, we focus on how the order of skill acquisition shifts when the model is trained on the same task, without access to its own exploration.

\paragraph{SFT exhibits different learning dynamics by pattern structure.} 
The model trained without CoT fails to reliably learn most $n=4$ patterns. To probe the order in which the model learns the patterns to an intermediate level, we instead relax the threshold of high-precision to 50\% (\Cref{fig:subtrees-Qwen2.5-1.5B-SFT-50}). Even though the coverage of patterns for $n=5$ patterns decrease in a similar order as in RL: right-heavy > left-heavy > balanced, the model instead masters $n=4$ patterns (in-distribution) in the exact opposite order: right-heavy < left-heavy < balanced. This highlights that the clear hierarchy of difficulty caused by the subtree structure is inherent to the RL training, not an artifact of the dataset. 

\begin{table}[ht]
    \centering
    \caption{\textbf{Final performance of SFT-trained models.} \passk{} on $n{=}3,4$ (trained) and $n{=}5$ (held out). We present the performance of the base model and the GRPO-trained model as a basis for comparison.}
    \begin{tabular}{l *{3}{c}}
    \toprule
    Training & \multicolumn{1}{c}{$n{=}3$} & \multicolumn{1}{c}{$n{=}4$} & \multicolumn{1}{c}{$n{=}5$}\\
        \midrule
        SFT+No CoT & 0.92 & 0.82 & 0.33 \\
        SFT+Random CoT& 0.59 & 0.25 & 0.04 \\
        \midrule
        Base Model & 0.42 & 0.00 & 0.02 \\
        GRPO & 1.00 & 0.94 & 0.68 \\
        \bottomrule
    \end{tabular}
    \label{tab:sft}
\end{table}

\subsubsection{Training}
We train with batch size 256, prompt and response length 1024, max learning rate $10^{-6}$ with a linear warmup of 0.03 and cosine decay to zero. For all other hyperparmeters, we use the default values from the \texttt{verl} package. We train the \texttt{Qwen2.5-1.5B} model with and without the random chain of thought text, each with 1 random seed for dataset shuffling. 

\subsubsection{Learning dynamics}
\begin{figure}[h]
  \centering

  \includegraphics[width=\textwidth]{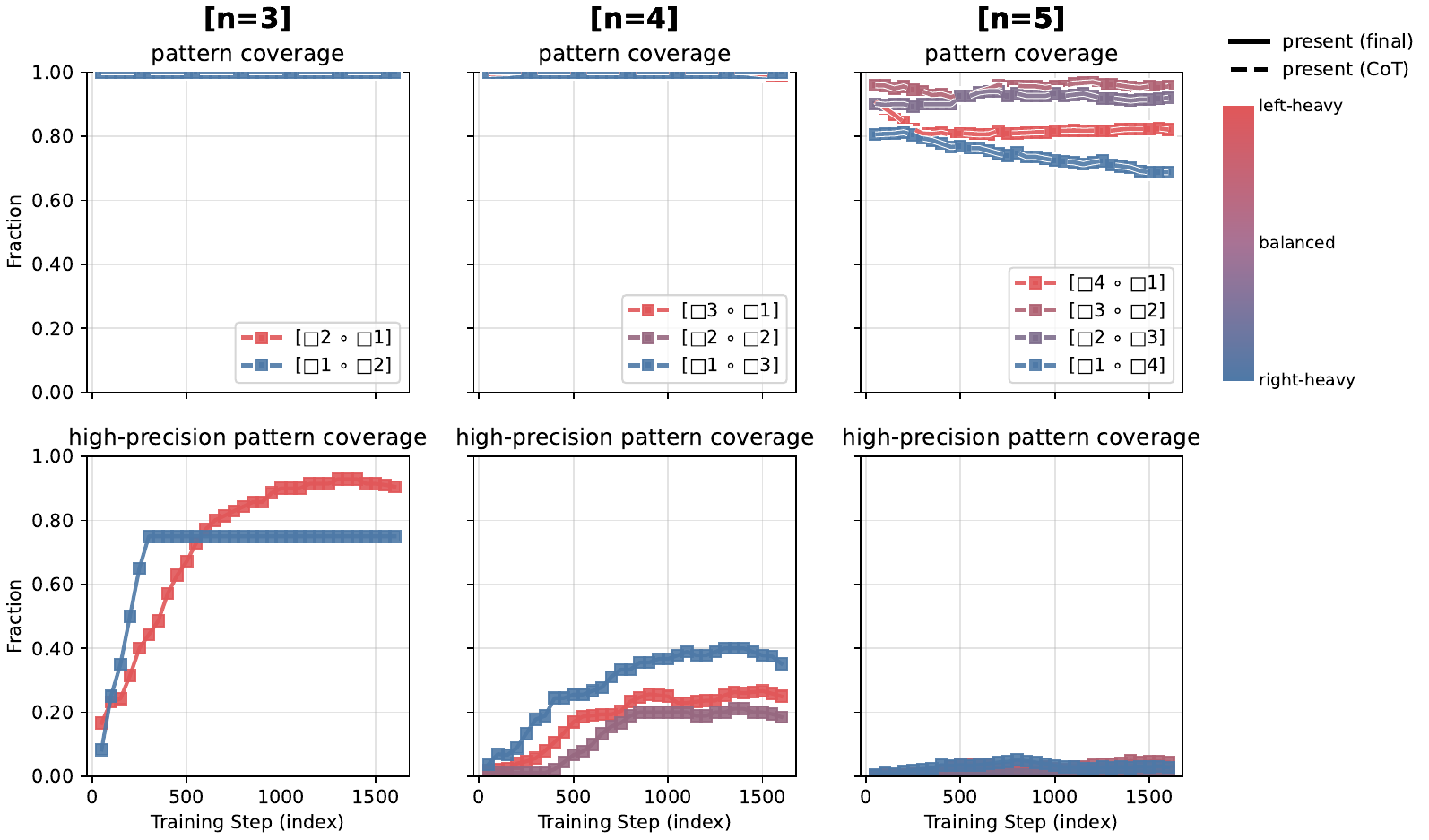}

  \caption{\textbf{Learning dynamics at a glance (Qwen2.5-1.5B with SFT)}. Here, the high-precision threshold has been lowered to 50\%. Even though the coverage of $n=5$ patterns (OOD) decrease in the following order: balanced > left-heavy > right-heavy, the model masters $n=4$ patterns (in-distribution) in the opposite order: right-heavy > left-heavy > balanced. This shows a fundamental difference in the learning dynamics from RL.}
  \label{fig:subtrees-Qwen2.5-1.5B-SFT-50}
\end{figure}

\begin{figure}[h]
  \centering

  \includegraphics[width=\textwidth]{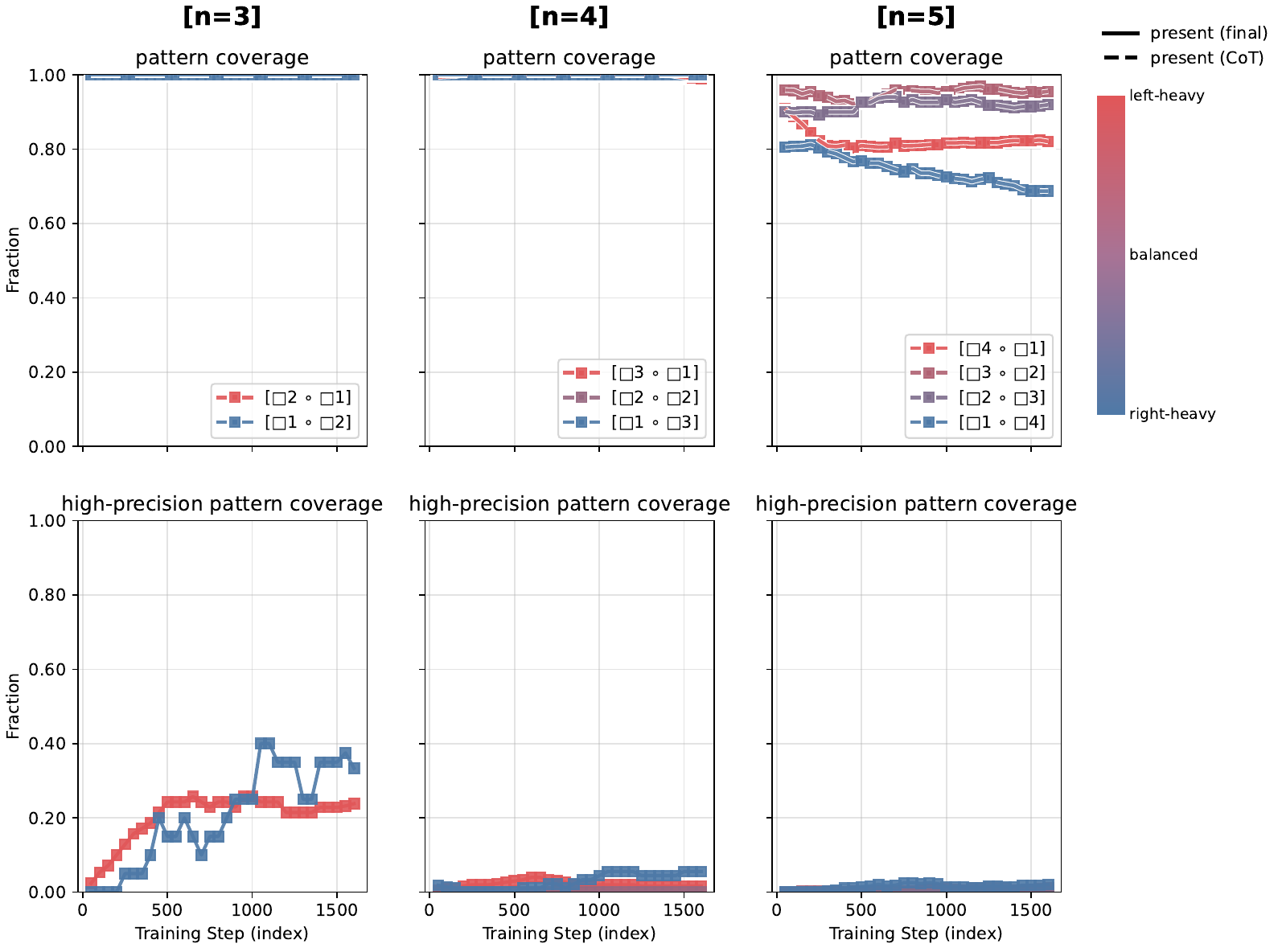}

  \caption{\textbf{Learning dynamics at a glance (Qwen2.5-1.5B with SFT)}. The model does not reliably learn any $n=4$ pattern.}
  \label{fig:subtrees-Qwen2.5-1.5B-SFT}
\end{figure}

\begin{figure}[h]
  \centering

  \includegraphics[width=\textwidth]{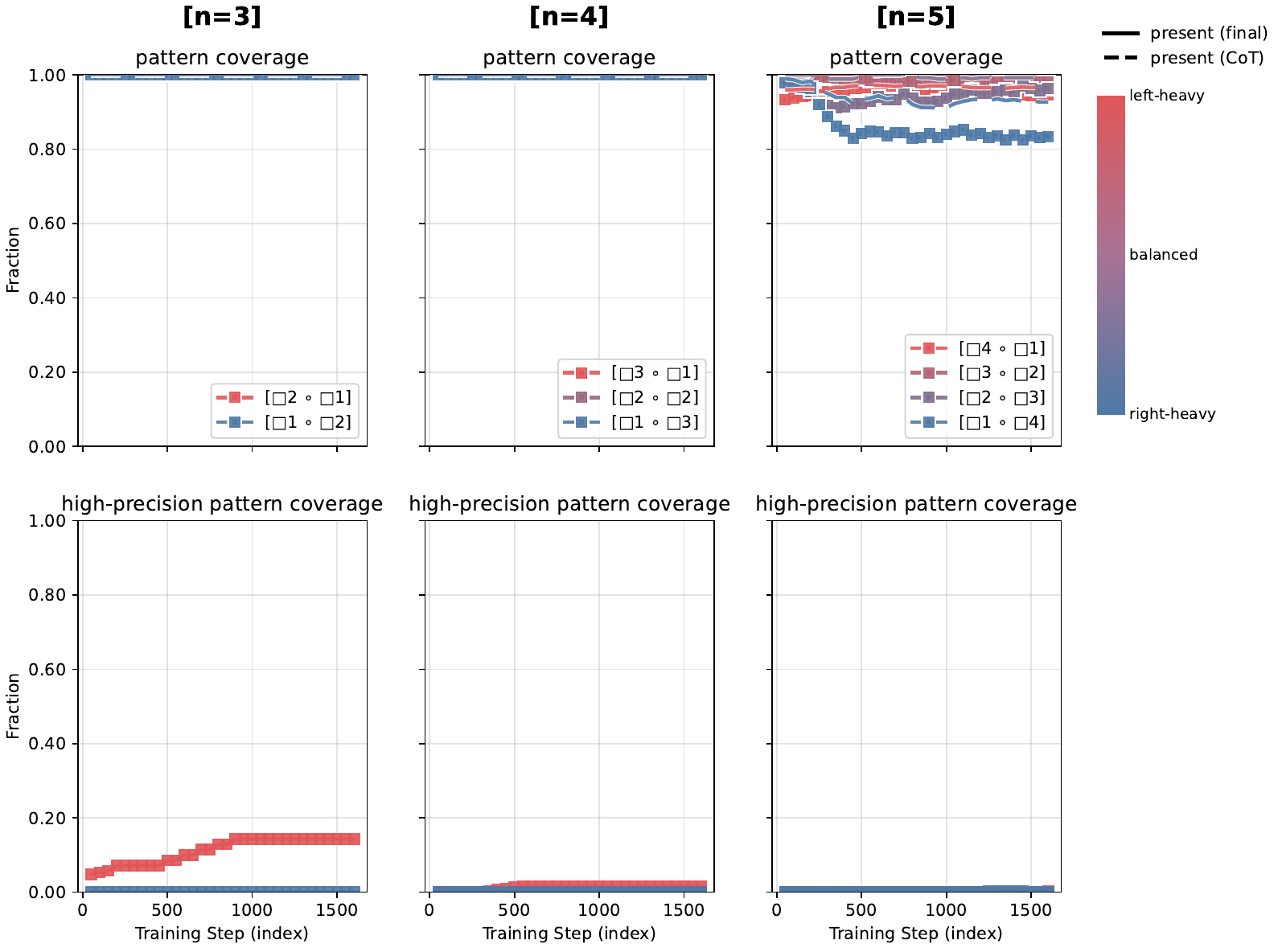}

  \caption{\textbf{Learning dynamics at a glance (Qwen2.5-1.5B with SFT+CoT)}. The model does not reliably learn any $n=4$ pattern.}
  \label{fig:subtrees-Qwen2.5-1.5B-SFT-CoT}
\end{figure}

In \Cref{fig:subtrees-Qwen2.5-1.5B-SFT-50,fig:subtrees-Qwen2.5-1.5B-SFT,fig:subtrees-Qwen2.5-1.5B-SFT-CoT}, we present the learning dynamics plots for the \texttt{Qwen2.5-1.5B} model trained with SFT (equivalent to \Cref{fig:subtrees}).

\clearpage

\subsection{Extension to Formal Mathematical Reasoning (More Detailed)}
\label{app:lean}

Lean~4 (\citep{lean4}) is an interactive theorem prover in which proofs are constructed by writing \emph{tactics} that incrementally transform the current proof state until all goals are discharged. Many proofs naturally decompose into subtasks: for an iff or conjunction, the model must prove both directions; for a case analysis, each case is a separate subtask. The \texttt{constructor} tactic, for example, opens both directions of an iff as two child goals, and the proof writer then dispatches each one. This hierarchical structure makes Lean a natural setting for the same compositional analysis we apply in \countdown{}: each binary-split tactic produces a \emph{left} subtask and a \emph{right} subtask, and the model's effort can be quantified as the number of tactics it spends on each.

\subsubsection{Choice of tactics}
We limit our analysis to proofs that open with a binary-split tactic at the top level of the proof body. Specifically, we include the following tactics:
\begin{itemize}
    \item \texttt{constructor} and \texttt{fconstructor}
    \item \texttt{refine $\langle$?\_, ?\_$\rangle$}
    \item \texttt{apply Iff.intro} and \texttt{apply And.intro}
    \item \texttt{refine Iff.intro ?\_ ?\_} and \texttt{refine And.intro ?\_ ?\_}
\end{itemize}
We deliberately exclude case-analysis tactics such as \texttt{induction},
\texttt{cases}, and \texttt{rcases}, since these often produce a trivial base
case and a substantial inductive step, a confounder that would inflate the number of right-heavy patterns. 

\begin{tcolorbox}[title=Example Proof: \texttt{Set.preimage\_subset\_preimage\_iff} (mathlib4)] \label{ex:lean-balanced}
\begin{verbatim}
theorem preimage_subset_preimage_iff {f : $\beta$ → $\alpha$}
  (hs : s $\subseteq$ range f) :
  f$^{-1}$' s $\subseteq$ f$^{-1}$' t $\leftrightarrow$ s $\subseteq$ t 
:= by constructor
  -- forward branch (3 tactics, $w_L$ = 3)
  intro h x
  apply h
  intro h x hx
  -- backward branch (2 tactics, $w_R$ = 2)
  rcases hs hx with ⟨y, rfl⟩
  exact h hx
\end{verbatim}
The top-level \texttt{constructor} produces two subgoals; subsequent
tactics are partitioned into the left and right branches based on which
subgoal they discharge. With $w_L = 3$ and $w_R = 2$, this proof has
$r = 3/(3+2) = 0.6$ and is classified as \emph{left-heavy}
(see \Cref{sec:lean:def}).
\end{tcolorbox}

\subsubsection{Definition of pattern structure}
\label{sec:lean:def}

For a proof body that opens with a binary-split tactic, we count the number of
top-level tactics used in each of the two immediate subtasks and write them
as $(w_L, w_R)$. Subtasks are delimited by the Lean~4 bullet syntax (\texttt{$\cdot$}) or by \texttt{case}-tagged blocks; nested \texttt{by}-blocks within a single \texttt{have}/\texttt{show} are counted as one tactic at the parent level. Comments and blank lines are ignored. We let
\[
    r \;=\; \frac{w_L}{w_L + w_R},
\]
and categorize the shape of each proof as
\[
    \text{shape}(r) \;=\;
    \begin{cases}
        \text{left-heavy} & \text{if } r \geq 0.6 \\
        \text{right-heavy} & \text{if } r \leq 0.4 \\
        \text{balanced} & \text{otherwise.}
    \end{cases}
\]
The interval $(0.4, 0.6)$ for the balanced bucket matches the convention used
for $n=5$ \countdown{} puzzles. In \Cref{app:lean-threshold}, we verify that the results are invariant to the choice of the threshold values.

\subsubsection{Dataset}
\paragraph{Distribution of the SFT dataset.}
We take 323K samples ($19\%$) from the full dataset (1.74M samples) and find that roughly 47K ($14\%$ of the sampled data) use binary-split tactics. Among those, $51\%$ use balanced proofs, $14\%$ use left-heavy proofs, and $35\%$ use right-heavy proofs.

\paragraph{Preparing the in-distribution dataset.}
We sample 100 questions from \texttt{Goedel-LM} / \texttt{SFT\_dataset\_v2} \citep{goedelv2} while ensuring that the dataset's structural distribution
among proofs with binary-split tactics is maintained: 14 left-heavy, 51 balanced, and 35 right-heavy.

\paragraph{Preparing the out-of-distribution dataset.}
We filter for questions in the LeanDojo 4 Benchmark \citep{yang2023leandojo} whose mathlib4 \citep{mathlib2020} reference solutions use the binary-split tactics. Next, among questions whose reference solutions have balanced / left-heavy / right-heavy proofs, we respectively sample 43 / 46 / 46 questions.

\subsubsection{Evaluation}
\label{sec:lean:eval}
\paragraph{Generation hyperparameters.}
We run \texttt{Goedel-Prover-V2-32B} with $k=32$ samples per question, temperature $0.6$, top-$p$ $0.95$, and maximum new tokens 8192. The prompt is wrapped with the model's Qwen3 chat template with the user prompt directly from the SFT data.

\paragraph{Parsing the proof body.}
The model's raw output typically contains a chain-of-thought paragraph
followed by one or more fenced code blocks. We extract the proof body by
(i)~locating all fenced blocks with an explicit language tag
(\texttt{lean}, \texttt{lean4}, \texttt{tactic}, \texttt{tactics}),
(ii)~preferring the final \texttt{lean*} block over any
\texttt{tactics*} scratch block, and (iii)~taking the substring after the
final \texttt{:= by} marker inside the chosen block. Outputs that do not
contain a parseable Lean code block are treated as empty proof bodies
and are discarded.

\subsubsection{Choice of threshold}
\label{app:lean-threshold}
We test whether the results in \Cref{sec:lean} are sensitive to the choice of threshold values for determining what counts as balanced, left-heavy, or right-heavy. In \Cref{tab:lean-threshold}, we present the result when re-categorizing the proofs with a different threshold: (1/3, 2/3). The results still hold.

  \begin{table}[ht]
      \centering   
      \caption{\textbf{Distribution of proofs by compositional structure (\%) under different thresholds (1/3, 2/3).} The results from \Cref{tab:lean} still hold.}
      \label{tab:lean-threshold}
      \begin{tabular}{@{}l r@{\;}l r@{\;}l r@{\;}l@{}}
          \\
          \toprule 
          & \multicolumn{2}{c}{balanced}  
          & \multicolumn{2}{c}{left-heavy}
          & \multicolumn{2}{c}{right-heavy} \\
          \midrule 
          Present in SFT dataset
              & 58 &
              & 11 &
              & 31 & \\
          Generated by RL model (ID)
              & 65 & \textcolor{deltaPos}{+7}
              & 11 &
              & 25 & \textcolor{deltaNeg}{-6} \\     
          Generated by RL model (OOD)
              & 77 & \textcolor{deltaPos}{+19}
              & 14 & \textcolor{deltaPos}{+3}
              & 9 & \textcolor{deltaNeg}{-22} \\   
          \bottomrule
      \end{tabular}
  \end{table}  \clearpage
\section{Additional Ablations}
\label{appendix:ablations}


\begin{figure}[!h]
    \centering
    \includegraphics[width=0.95\linewidth]{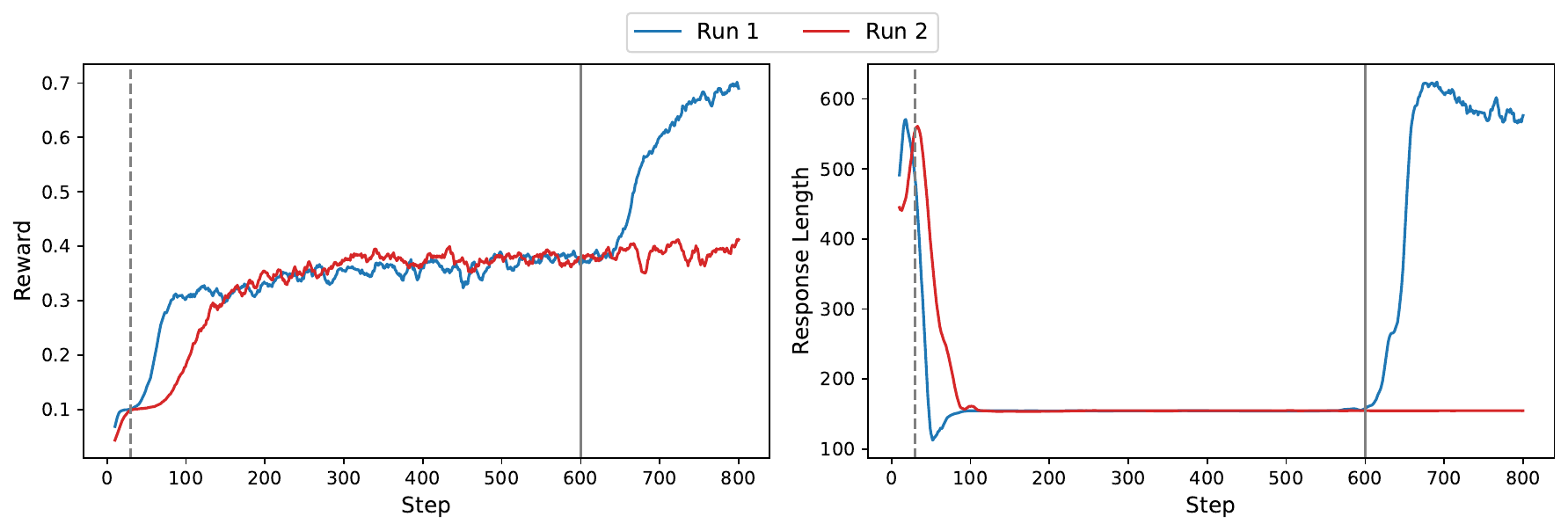}
    \caption{\textbf{Qwen2.5-1.5B often learns a formatting-only shortcut:} \textbf{(Left)} training reward; \textbf{(Right)} training response length of two distinct training runs. Once the model learns to collect the formatting reward (grey dotted vertical), the model attempts the puzzle without chain of thought, characterized by a sharp decrease in the response length. During some but not all runs (blue), we observe a phase transition (grey solid vertical), where the model starts generating chain of thought again.}
    \label{fig:reward-hacking}
\end{figure}

\subsection{A Formatting-Only Shortcut, and How to Avoid It}
\label{sec:reward_hacking}

Each answer can receive 0.1 points for correct formatting, plus up to 0.9 points for correctness.
Smaller models (1.5B and 3B) often exploit the format-only reward first (\Cref{fig:reward-hacking}), whereas the larger 7B model does not.
This shows that pretraining endows the larger model with more advanced reasoning circuits.
For example, if it can collect reward $0.9$ with probability $>0.1$ from the start, then it has no incentive to overfit to the formatting reward.

\paragraph{Intervention (large-to-small group size schedule)}
To stabilize training, 
we warm-start training with a larger group size 
to encourage exploration of different patterns
\footnote{We find that $G=16$ for the 1.5B and $G=8$ for the 3B model are sufficient to induce the phase transition.} and reduce the group size to the default value of $G=4$ whenever the models make the phase transition and exit the shortcut. 

\subsection{Effect of Standard Deviation Normalization}
\label{sec:stddev}
For our main experiments, we do not apply std normalization to the group advantage, following \citet{liu2025understanding}. Here, we present the results when we do normalize the advantages by the standard deviation in each group \citep{shao2024deepseekmathpushinglimitsmathematical}. All results are from the 1.5B model.

\subsubsection{Negative Gradients Induce Longer Responses}
\begin{figure}[!h]
    \centering
    \includegraphics[width=0.95\linewidth]{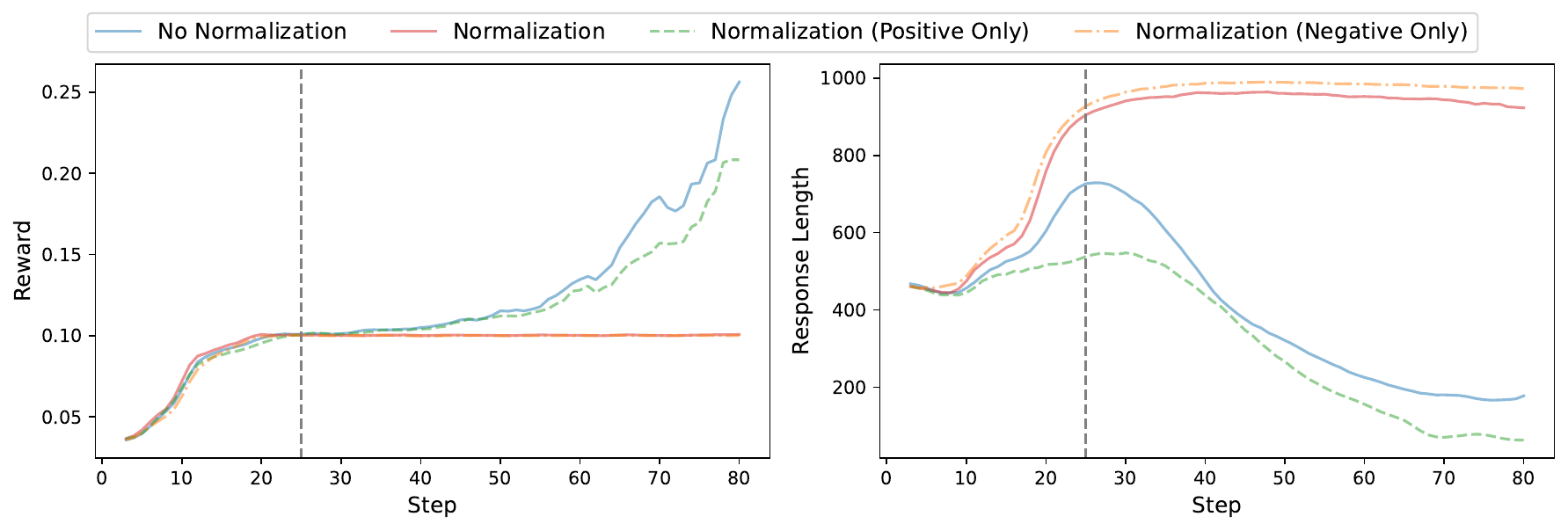}
    \caption{\textbf{When normalizing group rewards by standard deviation, negative gradients make responses longer:} \textbf{(Left)} training reward; \textbf{(Right)} training response length. In the first 25 steps, training reward increases to 0.1 at the same pace across all runs. However, with standard deviation normalization (red), the response length peaks high, and the training reward stalls. Training with only positive gradients (green dotted) and negative gradients (yellow dotted) respectively mimic the behavior of training without (blue) and with (red) normalization. }
    \label{fig:stddev}
\end{figure}

When initializing training with the normalization, the response length increases faster and stays higher, compared to when training without normalization (\Cref{fig:stddev}). We observe a similar trend with a different model (\Cref{appendix:llama}).

To identify the cause of the phenomenon, we separately train with positive gradients only (set the group advantage $A_i = \max(A_i, 0)$ after the standard deviation normalization) and negative gradients only ($A_i = \min(A_i, 0)$). The response length of the positive-only training run follows the trend of training without the normalization, whereas negative-only training mimics the trend of training with the normalization. This shows that the negative gradients are the sole cause behind the initial increase in the response length.

This behavior can be understood as the effect of the negative gradients on the token probabilities. Initially, weak models do not collect the final output reward, and the possible rewards are within $\{ 0, 0.1\}$. However, when the rewards are normalized by the standard deviation, the absolute value of the advantage becomes at most 20 times larger. Updates using the negative gradients greatly reduces the probability of all tokens generated, and in particular, the generation of the EOS character (which usually occurs after the \texttt{<answer></answer>} tags). After the probability of generating EOS token is lowered sufficiently, the model tends to continue generating tokens after the \texttt{<answer></answer>} tags.

\subsubsection{Standard Deviation Normalization Leads to Worse Performance}
We additionally notice that normalizing by the standard deviation leads to a worse performance at the end of the training (\Cref{tab:stddev-perf}).

\begin{table}[ht]
    \centering
    \caption{\textbf{Effect of standard deviation normalization on final performance.} \strictacc{} (All@), \avgacc{} (Avg), and \passk{} (Pass@) with or without standard deviation normalization during group rewards computation. Normalizing generally hurts final performance.}
    \begin{tabular}{lccc|ccc}
        \\ \toprule
        & \multicolumn{3}{c|}{No Normalization} & \multicolumn{3}{c}{Normalization} \\
        \midrule
        & All@ & Avg & Pass@ & All@ & Avg & Pass@ \\
        \midrule
        1.5B & 0.66 & 0.85 & 0.94 & 0.36 & 0.60 & 0.86 \\
        3B & 0.61 & 0.84 & 0.95 & 0.66 & 0.86 & 0.96 \\
        \bottomrule \\
    \end{tabular}
    \label{tab:stddev-perf}
\end{table}

\subsection{Effect of Number of Rollouts}
\label{sec:rollout}
In \cref{sec:reward_hacking}, we increase the number of rollouts only during the initial phase of training to induce the phase transition necessary for learning. Here, we explore maintaining the increased number of rollouts ($G=16$ for the 1.5B model and $G=8$ for the 3B model) until the end of the training. 

We notice that entropy remains slightly higher than training with a small group size ($G=4$) and the training reward may start decreasing before 1 epoch (around 0.5 epoch for 1.5B model and 0.75 epoch for 3B model). We stop training whenever we observe a decline in the training reward, and evaluate the final performance. The performance of these models is slightly worse than training switching to a smaller rollout mid-training and significantly worse than fully training with $G=4$ (\cref{tab:rollout-perf}). Other than the initial training stability, we observe no benefit of training with a larger group size.

\begin{table}[ht]
    \centering
    \caption{\textbf{Effect of group size on final performance.} \strictacc{} (All@), \avgacc{} (Avg), and \passk{} (Pass@) on different choices of group sizes. A larger number of rollouts generally hurts final performance.}
    \begin{tabular}{lccc|ccc|ccc}
        \\ \toprule
        & \multicolumn{3}{c}{Small} & \multicolumn{3}{c}{Large} & \multicolumn{3}{c}{Switch} \\
        \midrule
        & All@ & Avg & Pass@ & All@ & Avg & Pass@ & All@ & Avg & Pass@ \\
        \midrule
        1.5B & 0.66 & 0.85 & 0.94 & 0.46 & 0.78 & 0.93 & 0.46 & 0.79 & 0.95 \\
        3B & 0.61 & 0.84 & 0.95 & 0.46 & 0.78 & 0.94 & 0.48 & 0.83 & 0.98 \\
        \bottomrule \\
    \end{tabular}
    \label{tab:rollout-perf}
\end{table}

\subsection{Choice of RL Algorithm}
\label{sec:ppo}
Here, we investigate the effect of the RL algorithm. We replace the GRPO training with PPO \citep{schulman2017proximalpolicyoptimizationalgorithms}. Due to higher computational constraints, we only train with 1 seed. 

1.5B does not escape the formatting-only shortcut as in \Cref{sec:reward_hacking}. 3B and 7B improve training accuracy faster with respect to the same number of gradient updates. However, training entropy decreases much faster during PPO training, which translates to a lower \passk{} but a higher \strictacc{} (\Cref{tab:ppo}). Compared to GRPO, PPO tends to push the model behavior on each pattern to one extreme --- there are simultaneously more 1) \texttt{Absent} patterns and 2) patterns with precision of 100\%. 

\begin{table}[ht]
    \centering
    \caption{\textbf{Effect of RL algorithm on final performance.} \strictacc{} (All@), \avgacc{} (Avg), and \passk{} (Pass@) of models. Using PPO instead of GRPO improves \strictacc{} at the cost of \passk{}.}
    \begin{tabular}{lccc|ccc}
        \\ \toprule
        & \multicolumn{3}{c}{GRPO} & \multicolumn{3}{c}{PPO} \\
        \midrule
        & All@ & Avg & Pass@ & All@ & Avg & Pass@ \\
        \midrule
        3B & 0.61 & 0.84 & 0.95 & 0.65 & 0.81 & 0.90 \\
        7B & 0.62 & 0.83 & 0.94 & 0.76 & 0.85 & 0.93 \\
        \bottomrule
    \end{tabular}
    \label{tab:ppo}
\end{table}

\clearpage  \clearpage
\section{Results on a Different Model Family (Qwen3)}
\label{appendix:qwen3}
\begin{figure}[h]
  \centering

  \includegraphics[width=\textwidth]{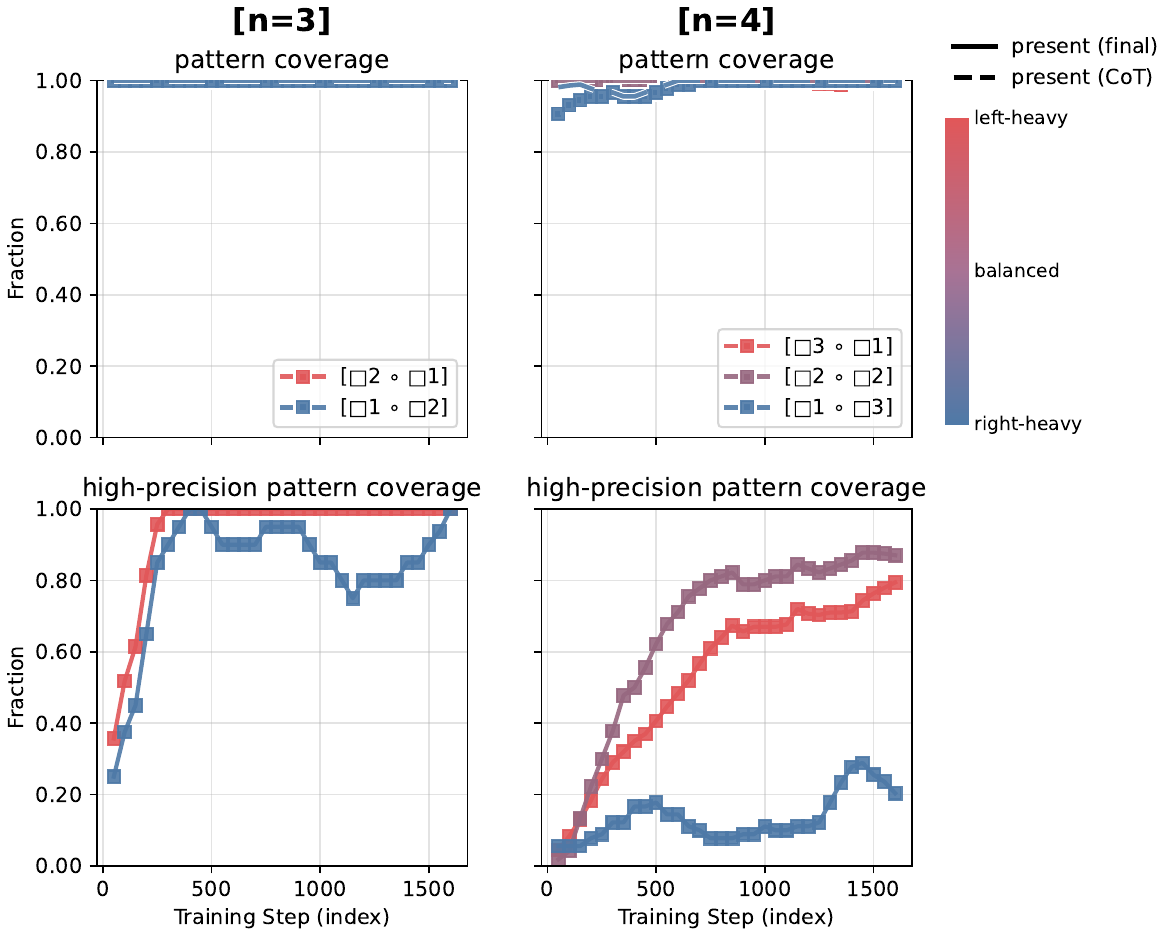}

  \caption{\textbf{Learning dynamics at a glance (Qwen3-1.7B-Base)}. The same conclusion from \Cref{fig:subtrees} holds about the hierarchy of difficulty by subtree structure: balanced $<$ left-heavy $<$ right-heavy.}
  \label{fig:subtrees-Qwen3-1.7B}
\end{figure}

We additionally evaluate on \texttt{Qwen3-1.7B-Base} \citep{yang2025qwen3technicalreport}. \Cref{fig:subtrees-Qwen3-1.7B} shows that the same result holds from \Cref{sec:results:structure-alt}.

\clearpage

\clearpage

\section{Results on a Different Model Family (Llama)}
\label{appendix:llama}
We additionally conduct some ablations studies on the \texttt{Llama-3.2} models (1B and 3B) and a \texttt{Llama-3.1} model (8B). We take the chat template from the \texttt{Instruct} variant but discard most of the system prompt, since the original version is too verbose and is out-of-distribution for the base model. See \Cref{prompt-chat-template-llama} for the exact chat template. We apply the same set of training hyperparameters.

\subsection{Formatting-only Shortcut}
Across all model sizes (even the 8B model), \texttt{Llama-3.2} and \texttt{Llama-3.1} models also find a similar formatting-only shortcut as outlined in \cref{sec:reward_hacking}. \texttt{Llama-3.1-8B} is able to escape the shortcut without a manual intervention ($G=4$ rollouts throughout training), but the \texttt{Llama-3.2} models are unable to escape the shortcut. By applying the group size switching strategy ($G = 16$ for \texttt{Llama-3.2-1B} and $G=8$ for \texttt{Llama-3.2-3B}), the models are able to escape the shortcut on 2 out of 3 runs each. This shows that the mitigation strategy for the shortcut is applicable to another model family, but the strength of its effect may depend on the performance of the base model. 

\subsection{\texttt{Llama} Models Do Not Properly Learn}
Even after escaping the formatting-only shortcut, the \texttt{Llama} models do not properly learn to use chain-of-thought to improve its answers. Instead, the models generally guess one expression and pad it with repeated answers or gibberish text. 

\subsection{After SFT on Basic Formatting, RL Shows the Same Trend}
\begin{figure}[h]
  \centering

  \includegraphics[width=\textwidth]{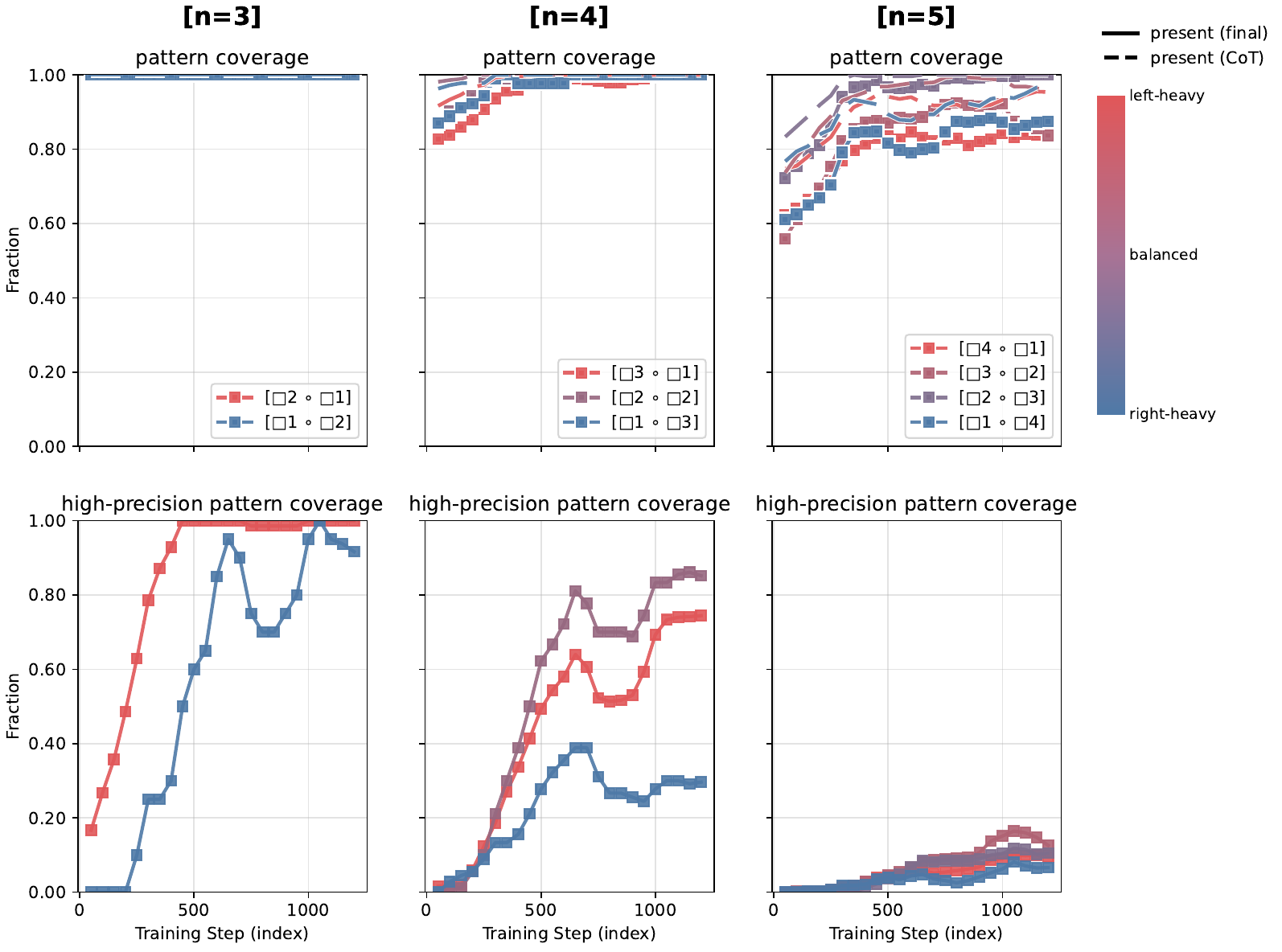}

  \caption{\textbf{Learning dynamics at a glance (Llama-3.2-3B after lightweight SFT)}. The same conclusion from \Cref{fig:subtrees} holds about the hierarchy of difficulty by subtree structure: balanced $<$ left-heavy $<$ right-heavy.}
  \label{fig:subtrees-Llama-3.2-3B}
\end{figure}

We speculate that the \texttt{Llama} base models may require a little more training on the correct formatting (including the \texttt{<think>} and \texttt{<answer>} tags). We SFT the \texttt{Llama-3.2-3B} model on examples from out dataset to teach the formatting.\footnote{We apply the random CoT template and the hyperparameters in \Cref{sec:sft}. We manually stop training after 50 gradient updates (before the learning rate reaches max value) when the training loss decreases below 1.} RL training on top of this warmed-up checkpoint shows a similar learning dynamics as in the main part of the paper (\Cref{fig:subtrees-Llama-3.2-3B}).

\subsection{Effect of Standard Deviation Normalization}
\begin{figure}[!h]
    \centering
    \includegraphics[width=0.95\linewidth]{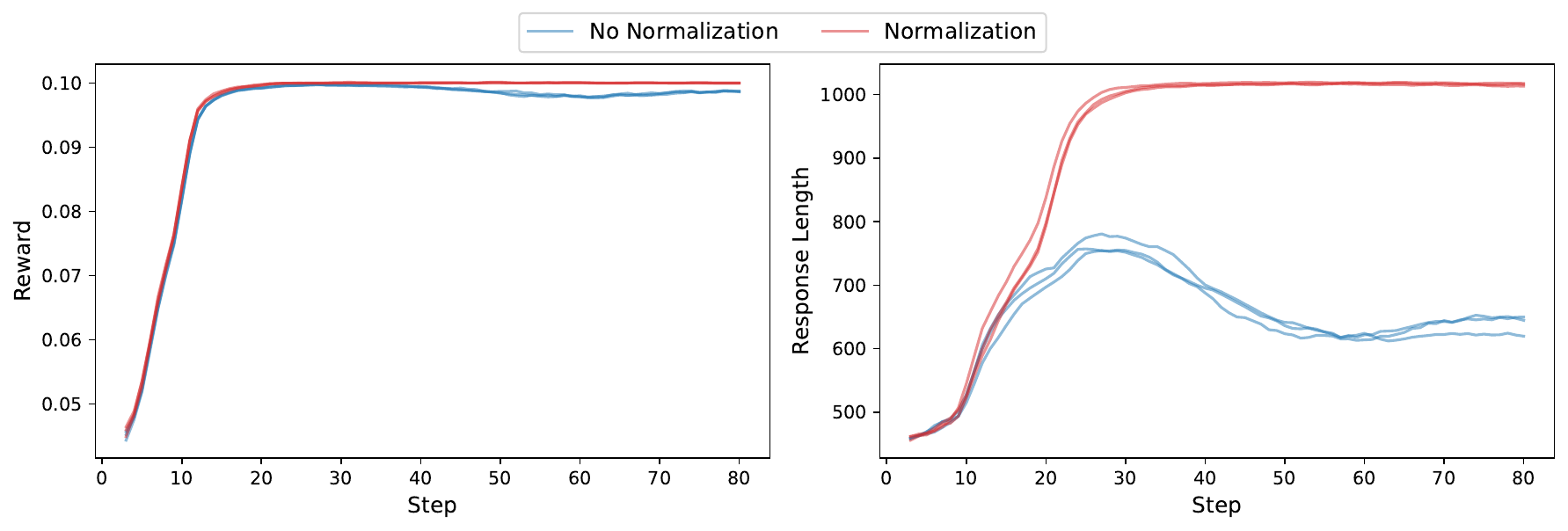}
    \caption{\textbf{Effect of standard deviation normalization on response length:} We reproduce the results from \Cref{fig:stddev}. }
    \label{fig:stddev-llama}
\end{figure}

We train with standard deviation normalization and reproduce its effect on the response length as outlined in \Cref{sec:stddev} and report the result in \Cref{fig:stddev-llama}.

\clearpage

\clearpage  \clearpage
\section{Prompts Used For Training}

\begin{tcolorbox}[title=Prompt Template for RL] \label{prompt-rl}
A conversation between User and Assistant. The user asks a question, and the Assistant solves it. The assistant first thinks about the reasoning process in the mind and then provides the user with the answer.

User: Using the numbers \{numbers\}, create an equation that equals \{target\}. You can use basic arithmetic operations (+, -, *, /) and each number can only be used once. Show your work in <think> </think> tags. And return the final answer in <answer> </answer> tags, for example <answer> (1 + 2) / 3 </answer>.

Assistant: Let me solve this step by step.

<think>
\end{tcolorbox}

\begin{tcolorbox}[title=Prompt Template for SFT] \label{prompt-sft}
A conversation between User and Assistant. The user asks a question, and the Assistant solves it. The assistant first thinks about the reasoning process in the mind and then provides the user with the answer.

User: Using the numbers \{numbers\}, create an equation that equals \{target\}. You can use basic arithmetic operations (+, -, *, /) and each number can only be used once. Show your work in <think> </think> tags. And return the final answer in <answer> </answer> tags, for example <answer> (1 + 2) / 3 </answer>.

Assistant: Let me solve this step by step.

<think>Using the numbers \{numbers\}, create an equation that equals \{target\}.\{CoT text\}</think>

<answer>\{Solution text\}</answer>
\end{tcolorbox}

\begin{tcolorbox}[title=Example CoT Text for SFT] \label{prompt-sft-cot}
8 / 8 + 3 / 3 = 1 (Incorrect). 3 + 8 + 8 - 3 = 16 (Incorrect). 3 * 3 / 8 - 8 = -6.87 (Incorrect). 8 / (3 - 8 / 3) = 24 (Correct).
\end{tcolorbox}

\begin{tcolorbox}[title=Example Solution Text for SFT] \label{prompt-sft-sol}
8 / (3 - 8 / 3)
\end{tcolorbox}

\begin{tcolorbox}[title=Chat Template for \texttt{Llama} Models] \label{prompt-chat-template-llama}
<|begin\_of\_text|><|start\_header\_id|>user<|end\_header\_id|> \\ \\
\{User prompt\}<|eot\_id|><|start\_header\_id|>assistant<|end\_header\_id|> \\ \\
\end{tcolorbox}

\clearpage

 \clearpage
\section{Alternate Ways to Canonicalize Patterns}
\label{appendix:alt-patterns}
A central claim of this work is that the compositional structure of a problem solution (specifically balanced vs. unbalanced) dictates its difficulty for the model. This analysis, however, relies on our specific method for mapping syntactically diverse expressions to unique canonical operators. A valid concern is whether our findings are an artifact of this mapping. For instance, our normalization rules (\Cref{appendix:pattern}) resolve ambiguities by systematically preferring left-heavy structures (e.g., mapping both $9 \times (3 + 2)$ and $(3 + 2) \times 9$ to the canonical form $(A+B) \times C$). This choice could interact with an intrinsic bias of the model towards left-to-right sequential generation, thereby confounding the analysis of structural difficulty. To isolate the effect of structure from our analytical choices, we conduct two ablations to test the robustness of our conclusions.

\begin{figure}[h]
    \centering
    \includegraphics[width=\linewidth]{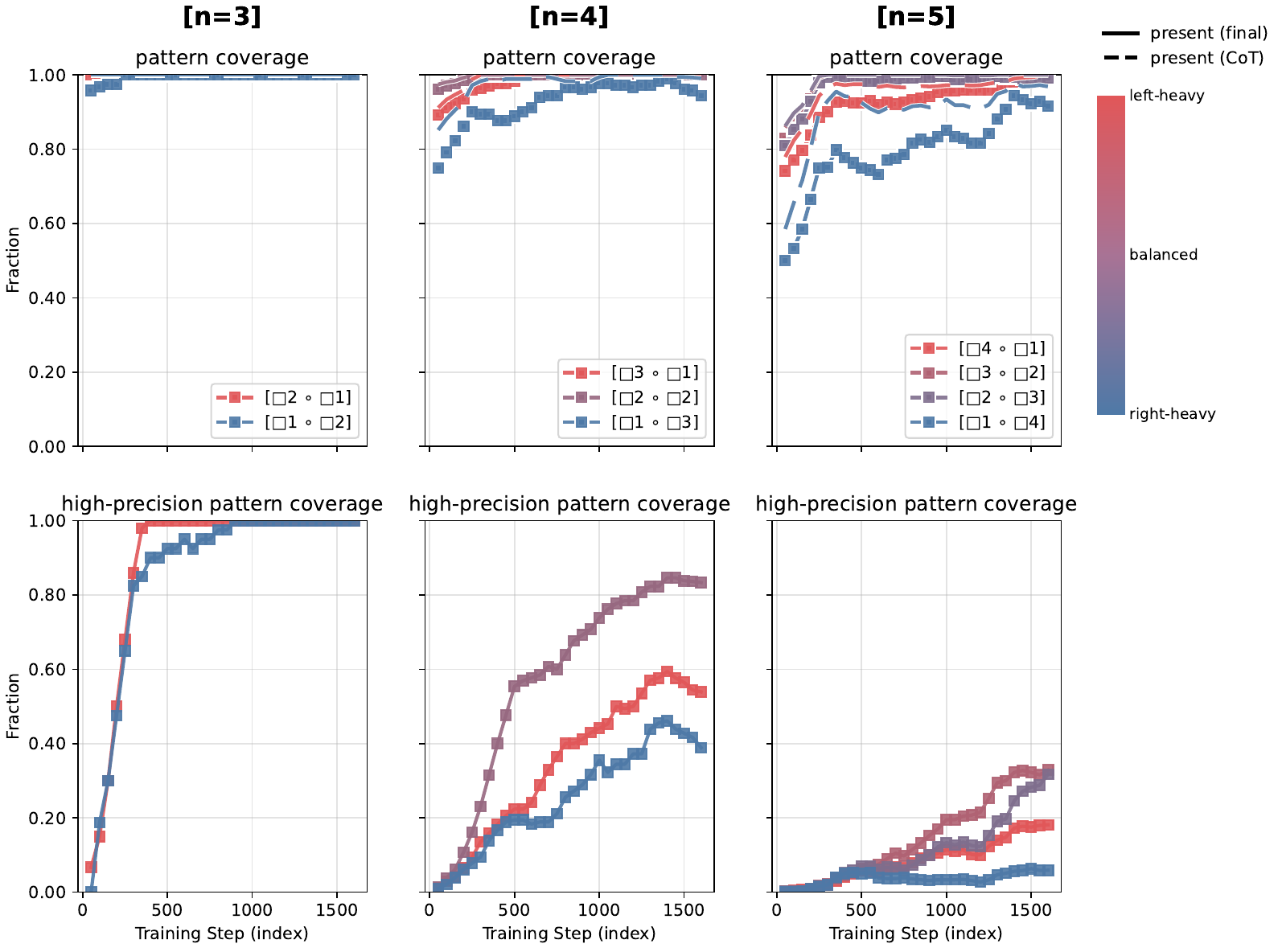}
    \caption{\textbf{Learning dynamics at a glance (Qwen2.5-1.5B)} with \emph{reversed} canonicalization. The same conclusion from \Cref{fig:subtrees} holds about the hierarchy of difficulty by subtree structure: balanced $<$ left-heavy $<$ right-heavy. }
    \label{fig:subtrees-qwen2.5-1.5b-reverse-canon}
\end{figure}

\paragraph{Reversed Canonicalization Preference}
First, we investigate whether our findings hold under an alternative normalization scheme. We invert the rule that resolves ambiguity based on subtree weight, now enforcing a preference for \emph{right-heavy} structures. For example, under this new canonicalization scheme, both $9 \times (3 + 2)$ and $(3 + 2) \times 9$ map to the canonical form $A \times (B+C)$. We then redo the analysis. The results, shown in \Cref{fig:subtrees-qwen2.5-1.5b-reverse-canon}, indicate that our central observations are unaffected by this change. We observe the same hierarchy of difficulty as reported in \Cref{sec:results:structure-alt}: balanced patterns are mastered most reliably, while the structures that were originally classified as right-heavy (and additionally structures now included in the right-heavy grouping) remain the most challenging. This suggests that the difficulty arises from the procedural requirement to commit to an operator before a complex subroutine is generated, rather than an arbitrary choice of a left- or right-associative canonical representation.

\begin{figure}[h]
    \centering
    \includegraphics[width=\linewidth]{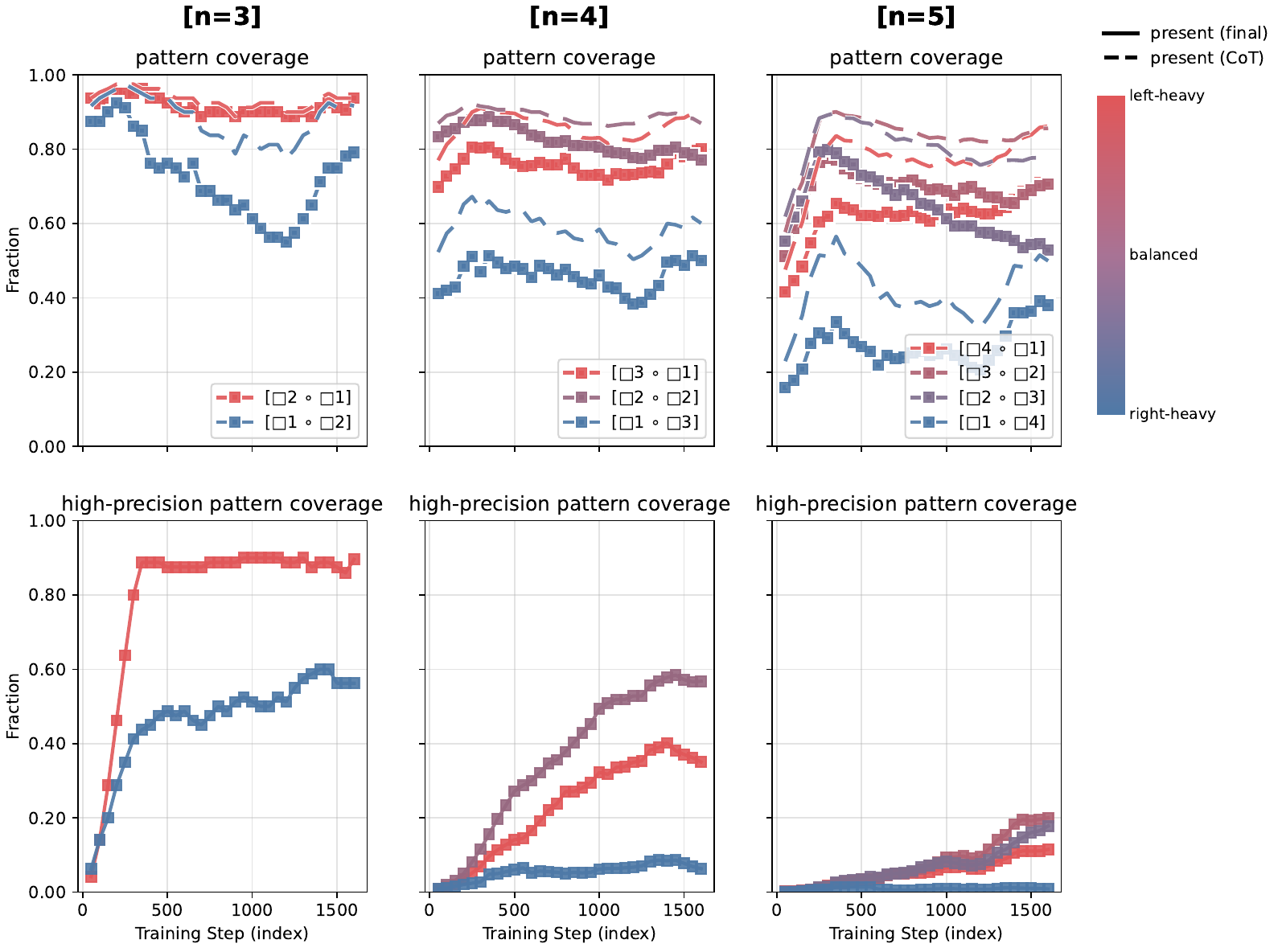}
    \caption{\textbf{Learning dynamics at a glance (Qwen2.5-1.5B)} \emph{without} canonicalization. While coverage decreases throughout training (models converge on one of multiple equivalent expressions), the same conclusion from \Cref{fig:subtrees} holds about the hierarchy of difficulty by subtree structure: balanced $<$ left-heavy $<$ right-heavy. }
    \label{fig:subtrees-qwen2.5-1.5b-no-canon}
\end{figure}

\paragraph{Analysis Without Canonicalization}
Second, we remove the canonicalization process entirely and analyze the raw expressions generated by the model. In this setting, a single problem may have multiple correct solutions corresponding to different tree shapes (e,g, $9 \times (3 + 2)$ would map to $A \times (B+C)$ and $(3 + 2) \times 9$ would map to $(A+B) \times C$. We group all valid expressions generated across the evaluation set by their unnormalized tree shapes (left-heavy, balanced, or right-heavy), and measure the per-shape pattern coverage and high-precision pattern coverage. As shown in \Cref{fig:subtrees-qwen2.5-1.5b-no-canon}, the structural difficulty persists. However, a model's intrinsic bias toward generating solutions in a particular syntactic form can artificially deflate the measured performance of any structural group that includes alternative, algebraically equivalent forms. For example, if a model only ever generates structures like $(A+B)+C$, but not $A+(B+C)$, then the coverage for $[1]\circ[2]$ goes down even though the model can correctly compose two additions.

\clearpage \clearpage
\section{Additional Plots}
\label{appendix:plots}
\begin{figure}[h]
  \centering

  \includegraphics[width=\linewidth]{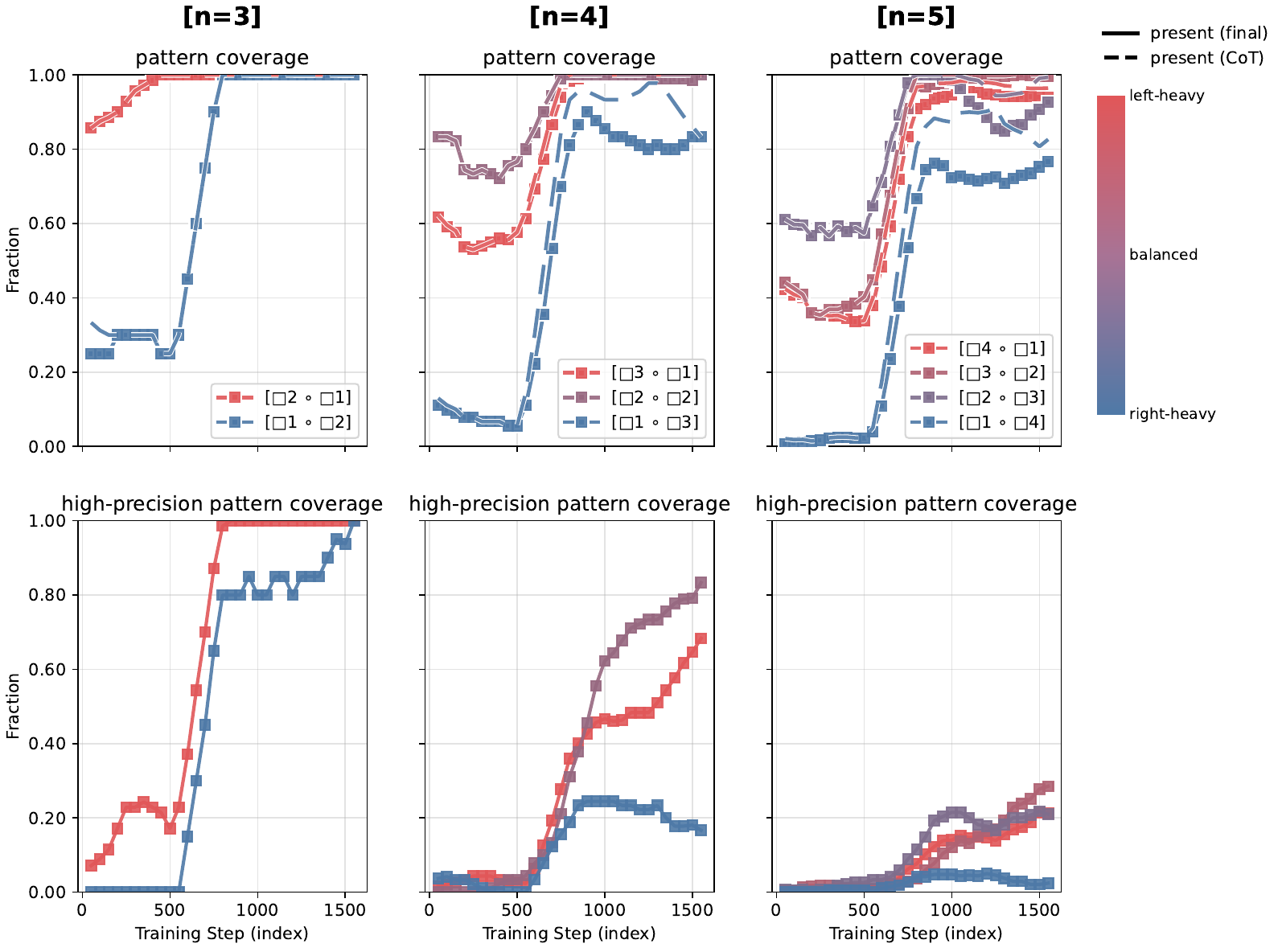}

  \caption{\textbf{Learning dynamics at a glance (Qwen2.5-1.5B, another seed)}. The same conclusion from \Cref{fig:subtrees} holds about the hierarchy of difficulty by subtree structure: balanced $<$ left-heavy $<$ right-heavy. }
  \label{fig:subtrees-qwen2.5-1.5b-seed1}
\end{figure}

\begin{figure}[h]
  \centering

  \includegraphics[width=\linewidth]{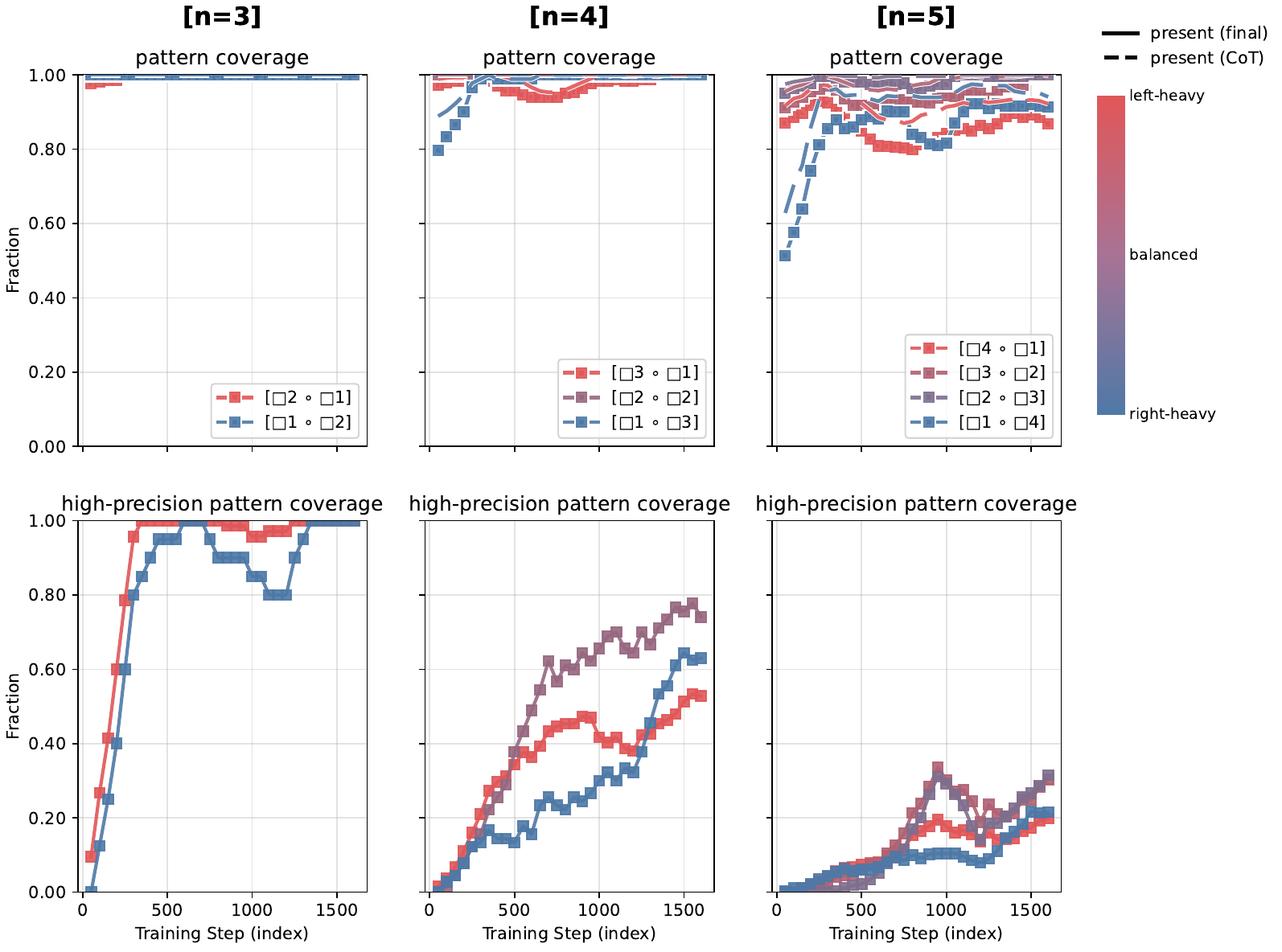}

  \caption{\textbf{Learning dynamics at a glance (Qwen2.5-3B)}. The same conclusion from \Cref{fig:subtrees} holds about the hierarchy of difficulty by subtree structure: balanced $<$ left-heavy $<$ right-heavy. }
  \label{fig:subtrees-qwen2.5-3b}
\end{figure}

\begin{figure}[h]
  \centering

  \includegraphics[width=\linewidth]{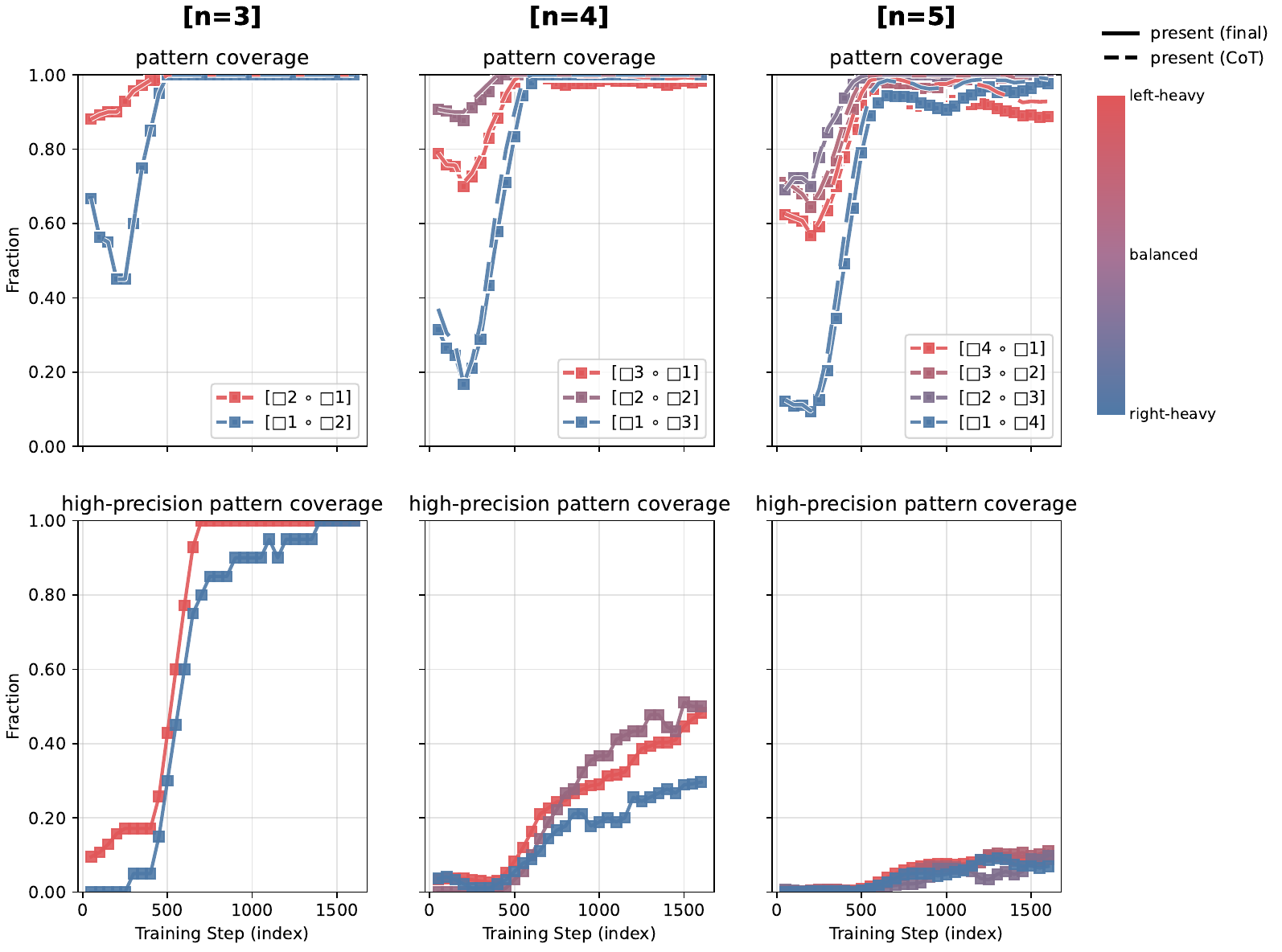}

  \caption{\textbf{Learning dynamics at a glance (Qwen2.5-3B, another seed)}. The same conclusion from \Cref{fig:subtrees} holds about the hierarchy of difficulty by subtree structure: balanced $<$ left-heavy $<$ right-heavy. }
  \label{fig:subtrees-qwen2.5-3b-seed2}
\end{figure}

\begin{figure}[h]
  \centering

  \includegraphics[width=\linewidth]{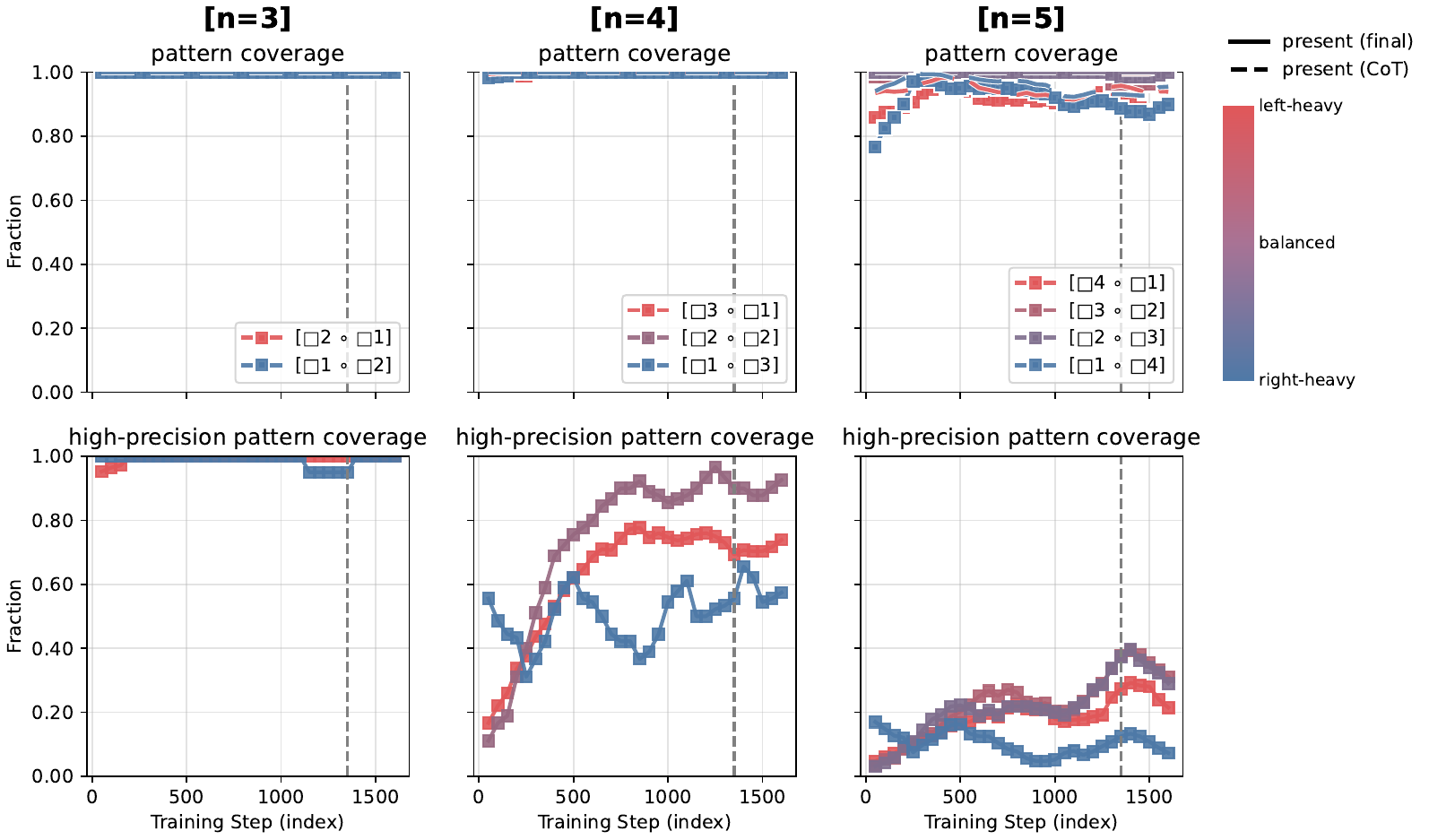}

  \caption{\textbf{Learning dynamics at a glance (Qwen2.5-7B)}. The same conclusion from \Cref{fig:subtrees} holds about the hierarchy of difficulty by subtree structure: balanced $<$ left-heavy $<$ right-heavy. We take the \textbf{best} checkpoint at 1350 steps (grey dotted), after which training reward starts declining.}
  \label{fig:subtrees-qwen2.5-7b}
\end{figure}

\begin{figure}[h]
  \centering

  \includegraphics[width=\linewidth]{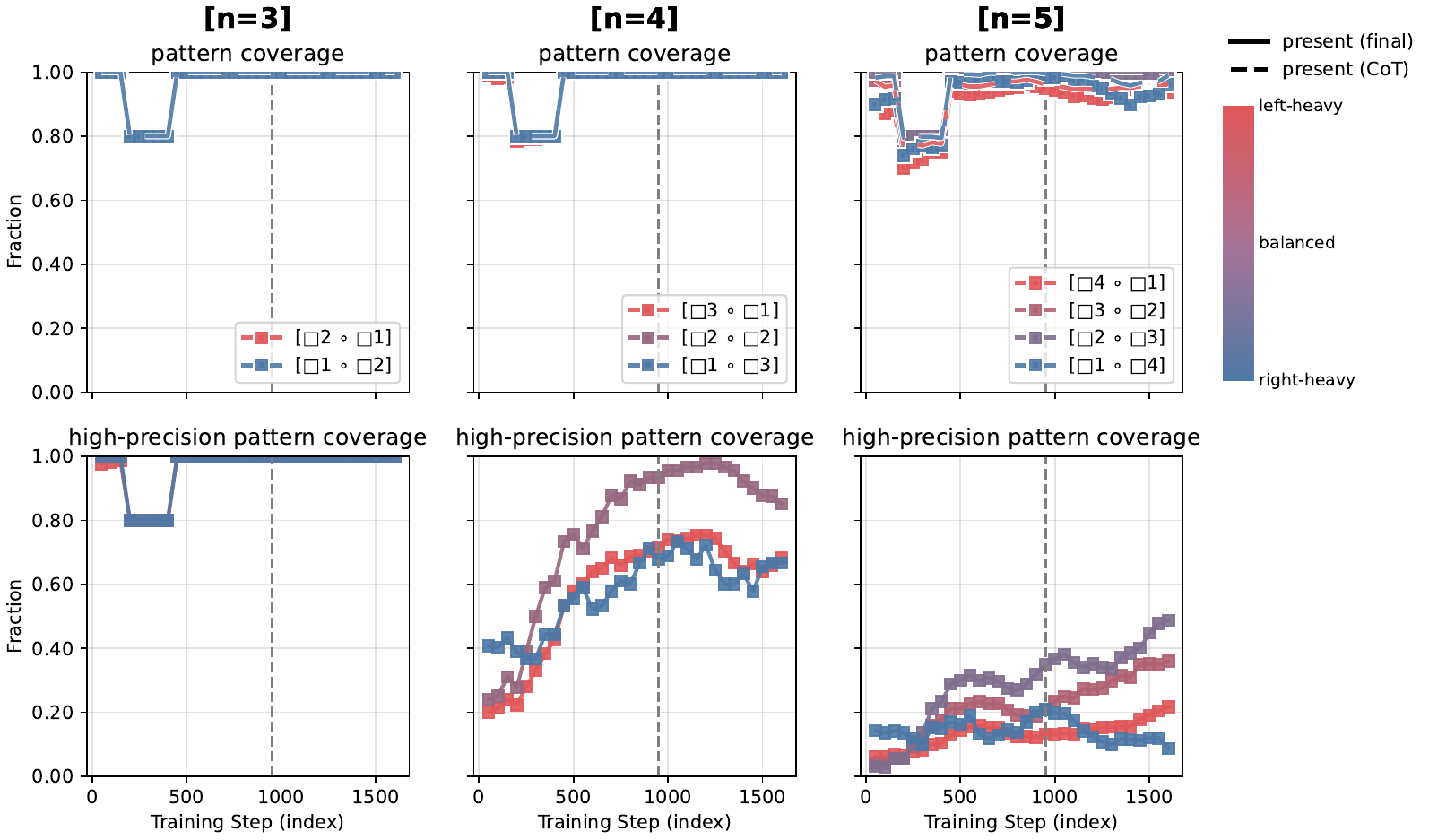}

  \caption{\textbf{Learning dynamics at a glance (Qwen2.5-7B, another seed)}. The same conclusion from \Cref{fig:subtrees} holds about the hierarchy of difficulty by subtree structure: balanced $<$ left-heavy $<$ right-heavy. We take the \textbf{best} checkpoint at 950 steps (grey dotted), after which training reward starts declining.}
  \label{fig:subtrees-qwen2.5-7b-seed1}
\end{figure}

\begin{figure}[h]
  \centering

  \includegraphics[width=\linewidth]{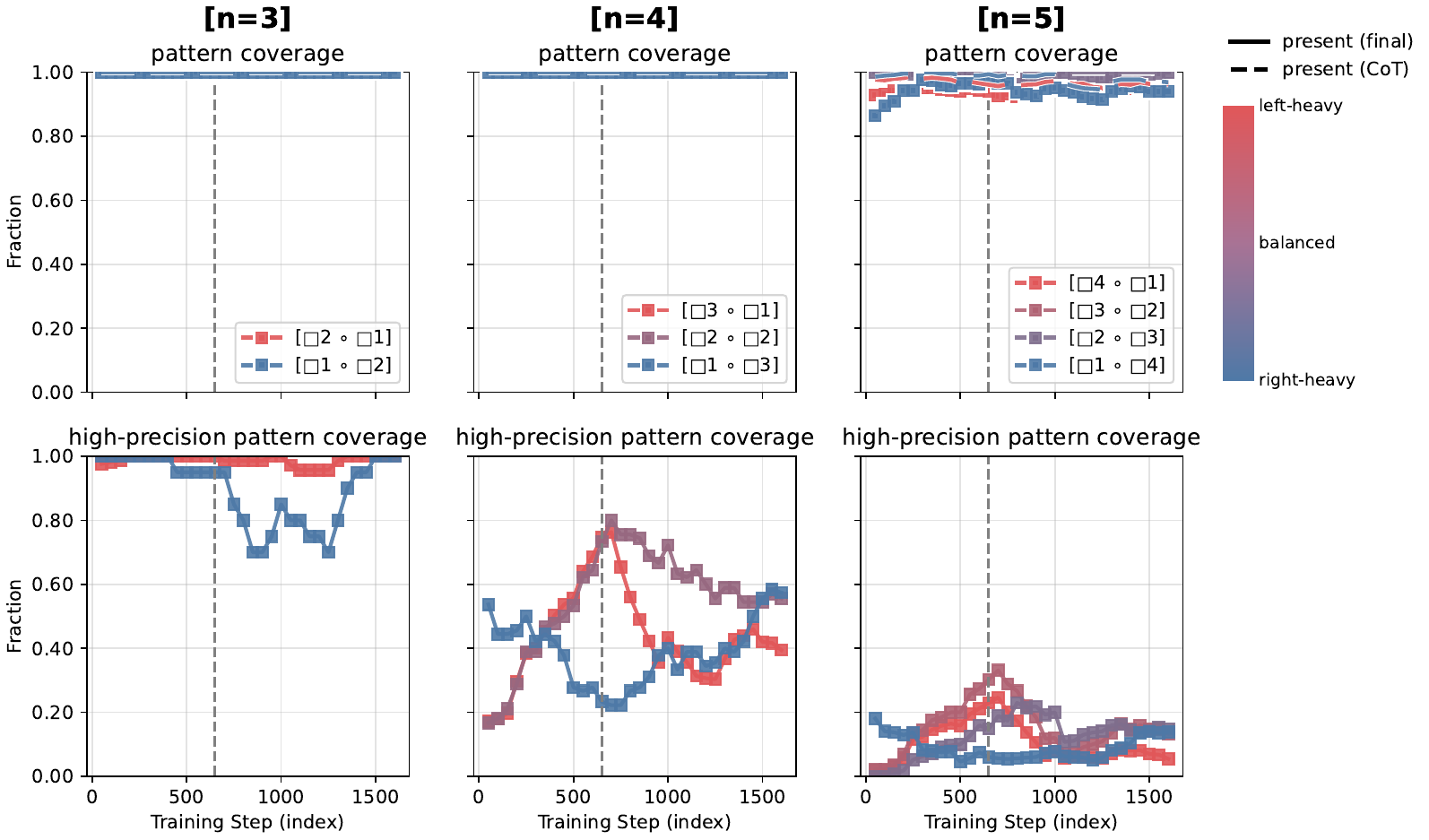}

  \caption{\textbf{Learning dynamics at a glance (Qwen2.5-7B, another seed)}. The same conclusion from \Cref{fig:subtrees} holds about the hierarchy of difficulty by subtree structure: balanced $<$ left-heavy $<$ right-heavy. We take the \textbf{best} checkpoint at 650 steps (grey dotted), after which training reward starts declining.}
  \label{fig:subtrees-qwen2.5-7b-seed2}
\end{figure}

In \Cref{fig:subtrees-qwen2.5-1.5b-seed1,fig:subtrees-qwen2.5-3b,fig:subtrees-qwen2.5-3b-seed2,fig:subtrees-qwen2.5-7b,fig:subtrees-qwen2.5-7b-seed1,fig:subtrees-qwen2.5-7b-seed2}, we present the learning dynamics plots for the \texttt{Qwen2.5-3B/7B} models and a different training run for the \texttt{Qwen2.5-1.5B} model (equivalent to \Cref{fig:subtrees}).

\begin{figure}[h]
  \centering

  \includegraphics[width=\linewidth]{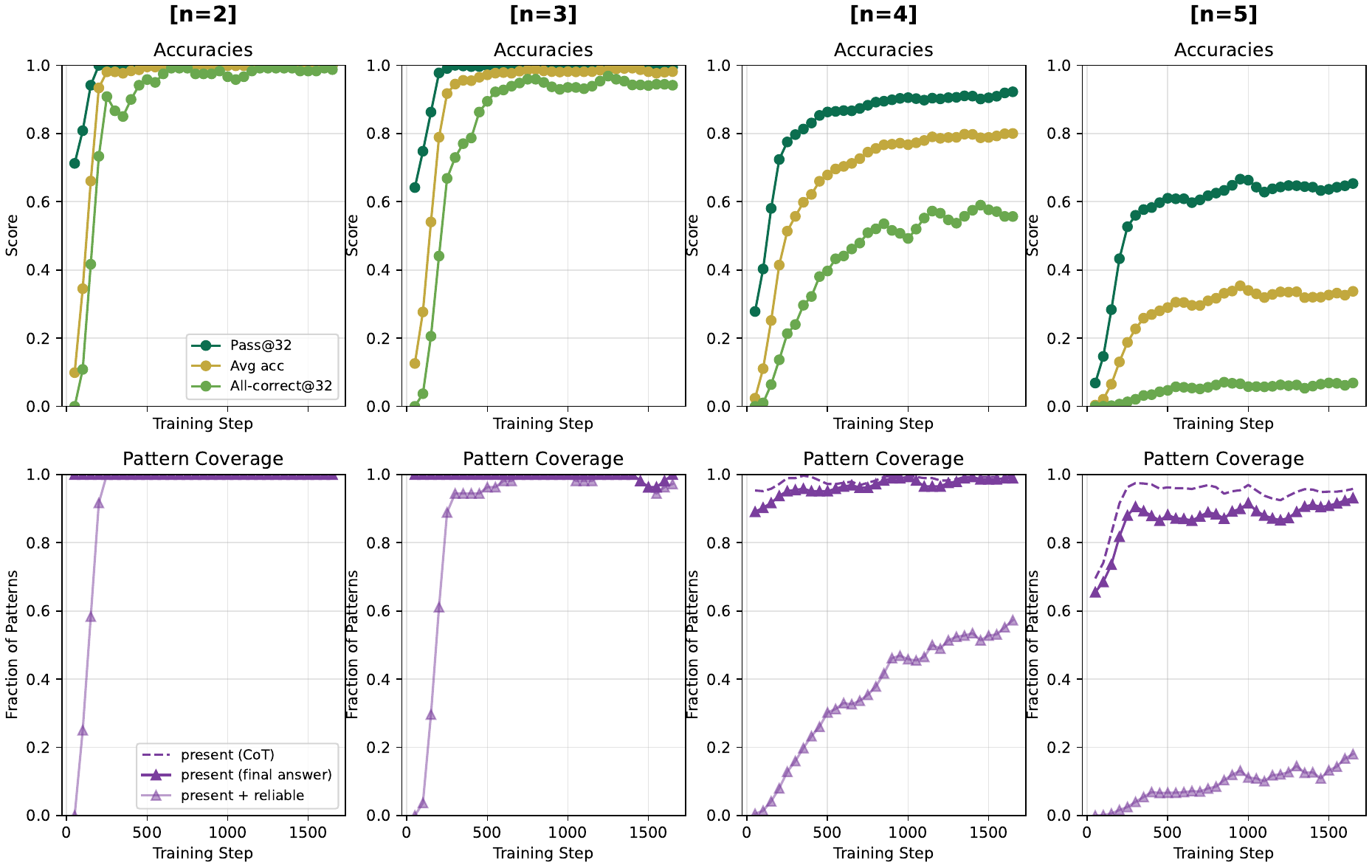}
  \vspace{0.5cm}

  \includegraphics[width=\linewidth]{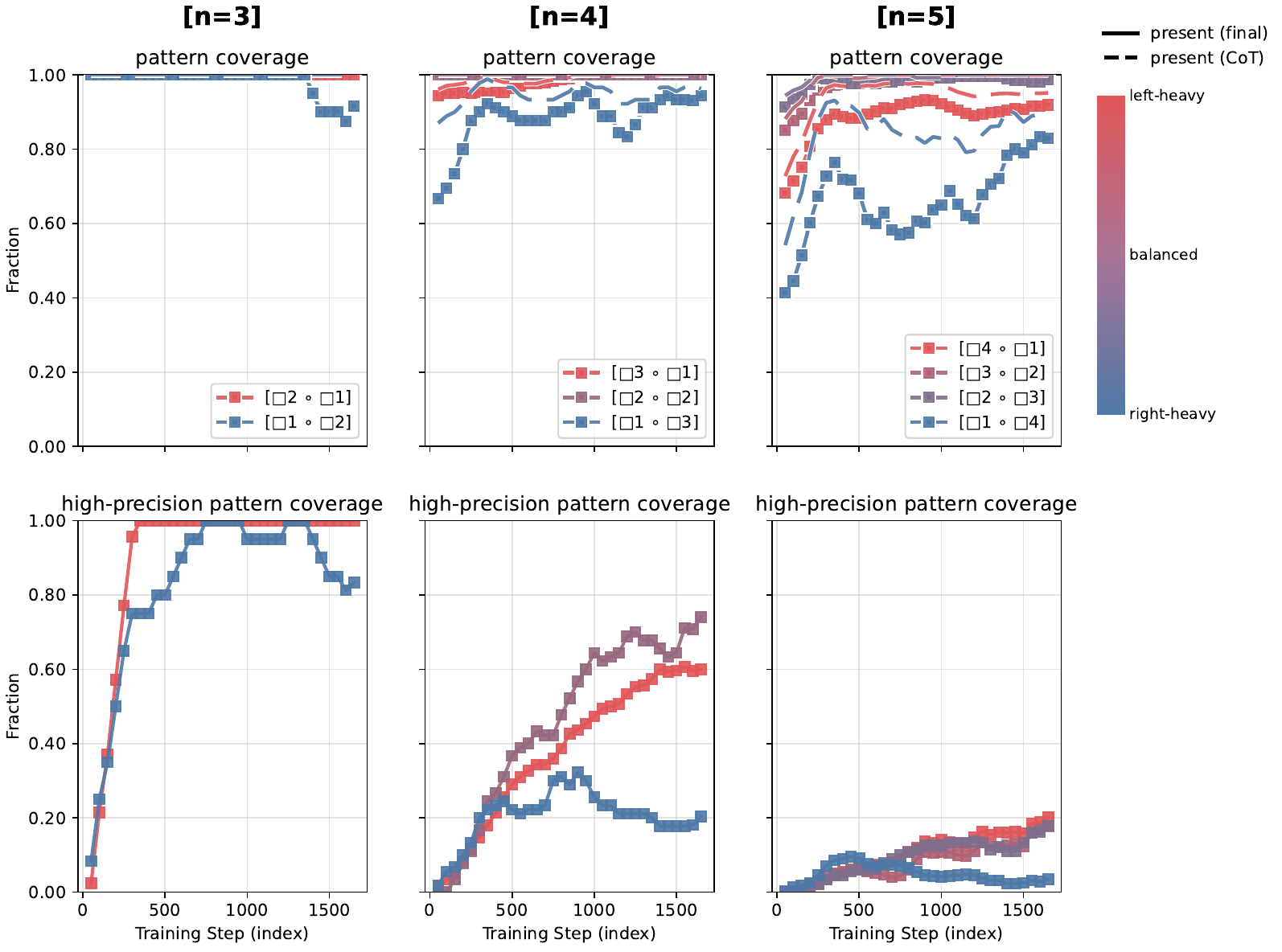}

  \caption{\textbf{Learning dynamics at a glance (Qwen2.5-1.5B)} trained on $n \in \{2,3,4\}$. The same conclusions from \Cref{fig:learning-dynamics,fig:subtrees} hold: models exhibit length generalization and master patterns progressively following a hierarchy of difficulty by subtree structure: balanced $<$ left-heavy $<$ right-heavy. }
  \label{fig:subtrees-qwen2.5-1.5b-bal234}
\end{figure}

In \Cref{fig:subtrees-qwen2.5-1.5b-bal234}, we present the learning dynamics plots for the \texttt{Qwen2.5-1.5B} model trained on $n \in \{2, 3, 4\}$ (equivalent to \Cref{fig:learning-dynamics,fig:subtrees}). For each of 4 patterns for $n = 2$, we generated 1000 examples for training and 10 examples for testing in the same way as in \Cref{sec:generating-balanced-dataset}.

\begin{figure}[h]
  \centering

  \includegraphics[width=\linewidth]{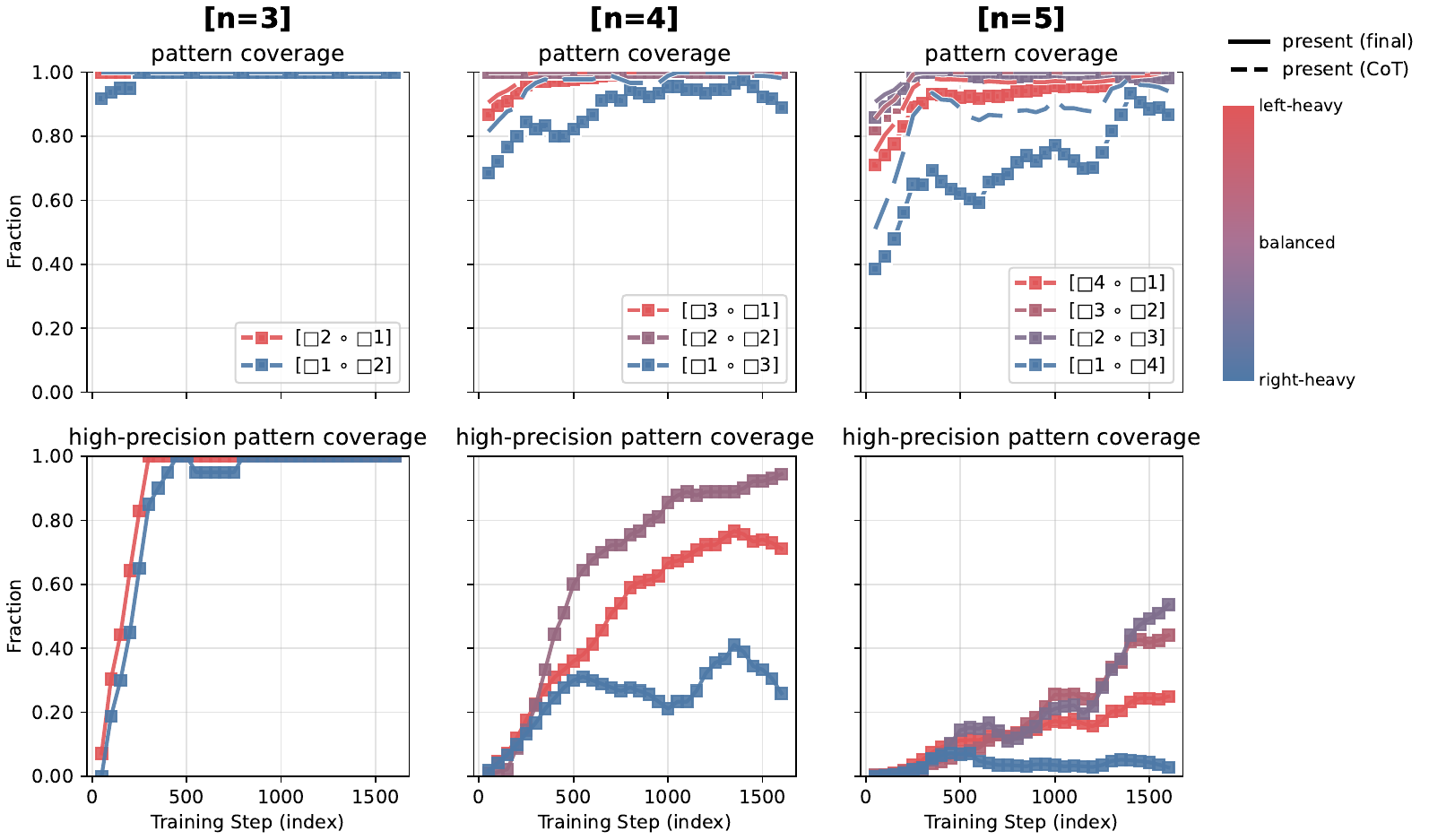}
  \includegraphics[width=\linewidth]{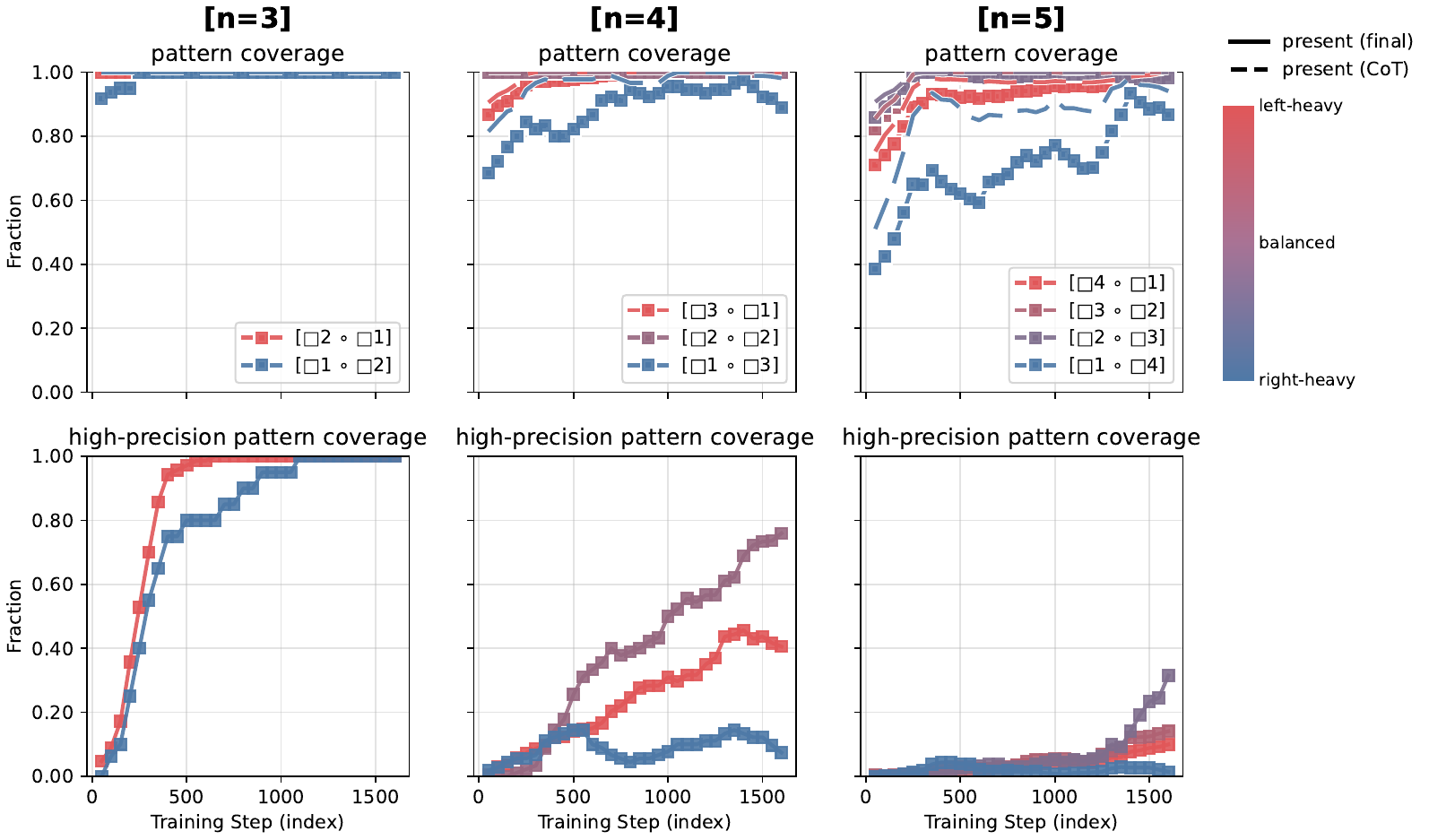}

  \caption{\textbf{Learning dynamics at a glance (Qwen2.5-1.5B)} but high-precision defined as precision $\geq 70\%$ (top); $\geq 90\%$ (bottom). The same conclusion from \Cref{fig:subtrees} holds about the hierarchy of difficulty by subtree structure: balanced $<$ left-heavy $<$ right-heavy.}
  \label{fig:subtrees-qwen2.5-1.5b-diff-threshold}
\end{figure}

In \Cref{fig:subtrees-qwen2.5-1.5b-diff-threshold}, we present the learning dynamics plots for the \texttt{Qwen2.5-1.5B} model (equivalent to \Cref{fig:subtrees}) when the high-precision threshold is changed to 70\% or 90\%.

\begin{figure}[h]
  \centering

  \includegraphics[width=\linewidth]{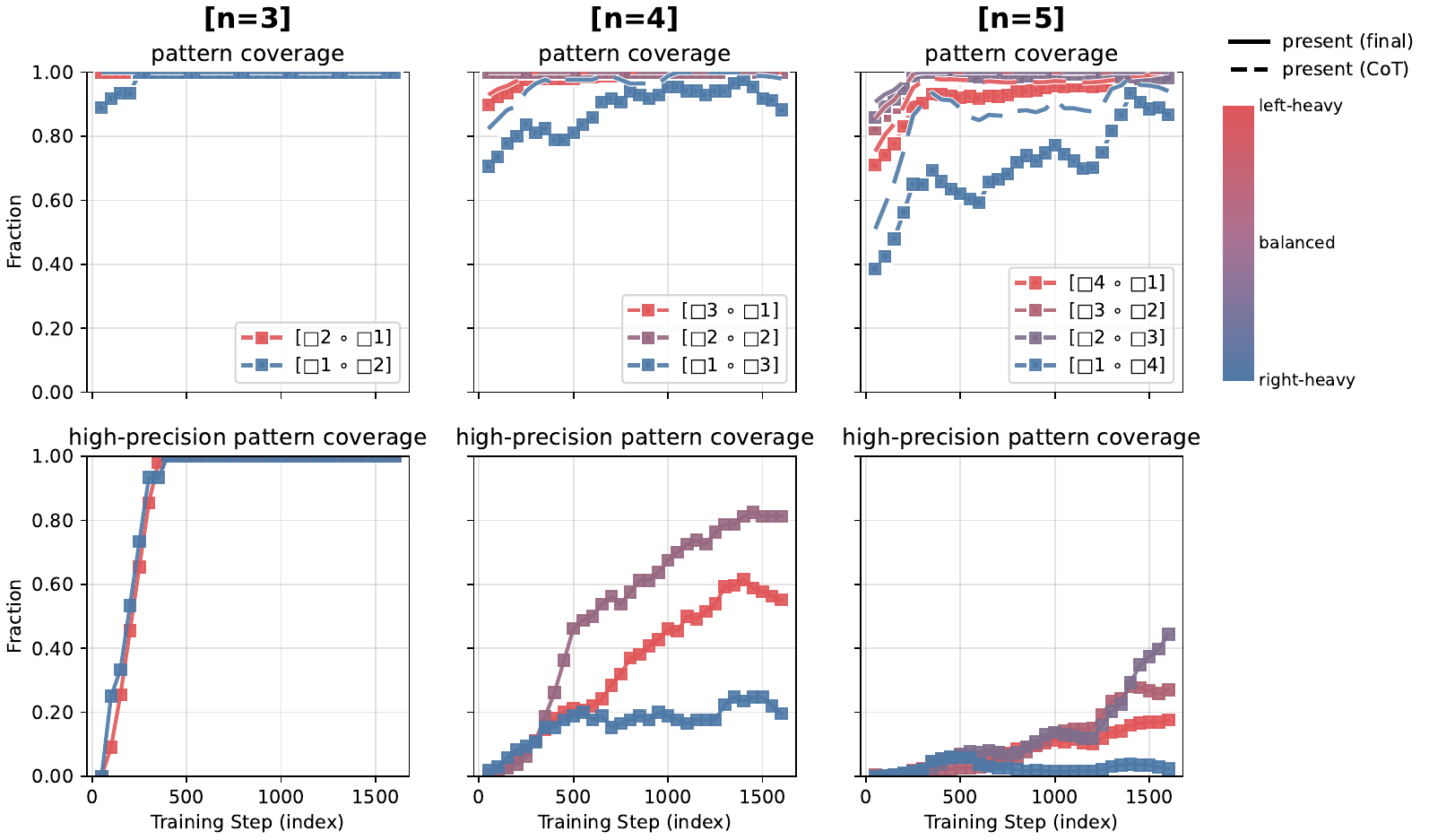}
  \caption{\textbf{Learning dynamics at a glance (Qwen2.5-1.5B, undersampled patterns removed)}. We remove the 15 undersampled patterns (\Cref{table:balanced_num_examples}) from the analysis. The same conclusion from \Cref{fig:subtrees} holds about the hierarchy of difficulty by subtree structure: balanced $<$ left-heavy $<$ right-heavy.}
  \label{fig:subtrees-qwen2.5-1.5b-remove-undersampled}
\end{figure}

In \Cref{fig:subtrees-qwen2.5-1.5b-remove-undersampled}, we present the learning dynamics plots for the \texttt{Qwen2.5-1.5B} model (equivalent to \Cref{fig:subtrees}) when we remove the 15 undersampled patterns (\Cref{table:balanced_num_examples}).

\clearpage

\clearpage

\clearpage

\clearpage

\clearpage

\clearpage

\begin{figure}[h]
  \centering
  \includegraphics[width=0.93\linewidth]{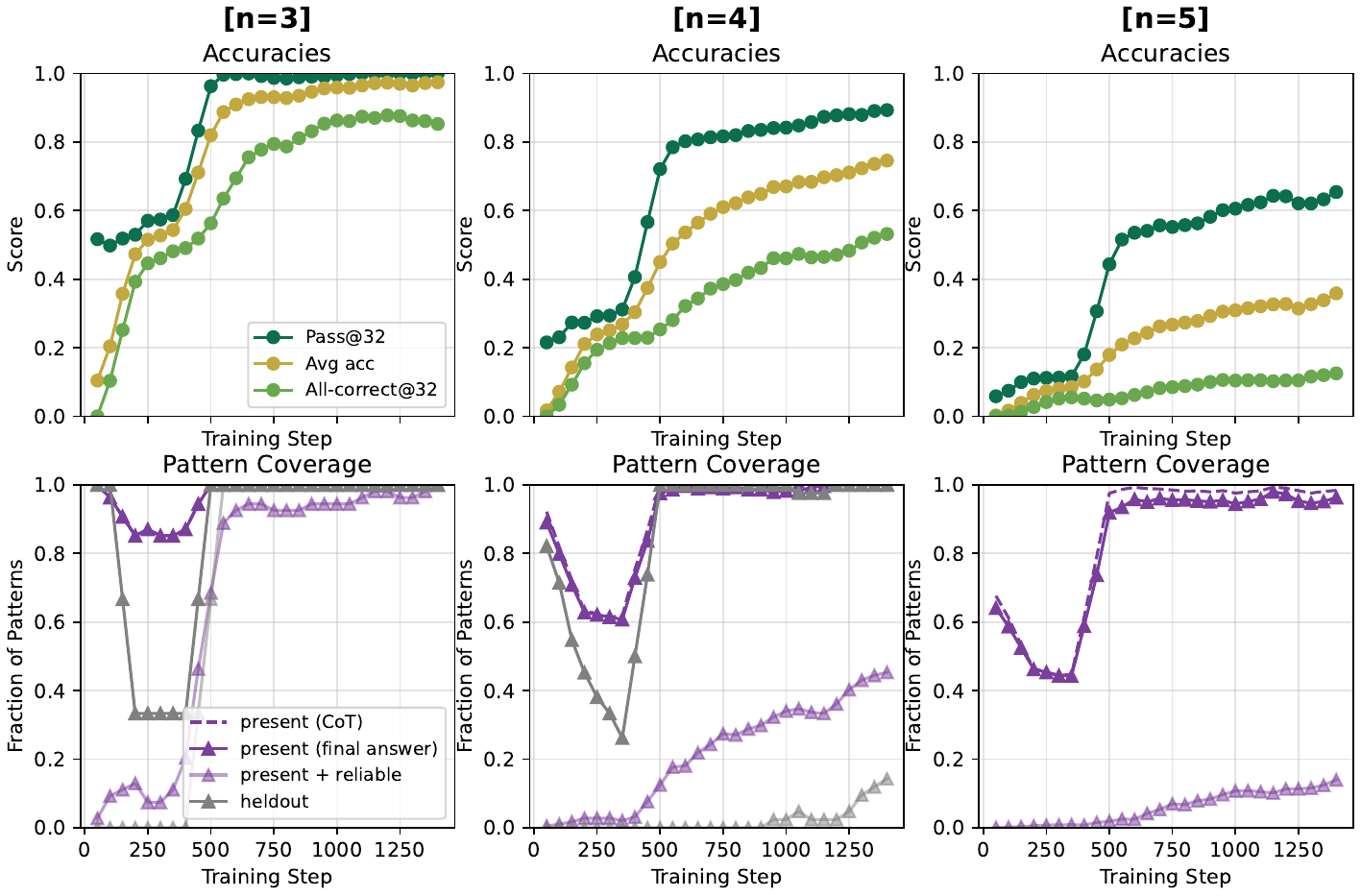}
  \caption{\textbf{Generalizes to entire families of held-out compositional patterns (Qwen2.5-1.5B, another seed).} The same conclusion from \Cref{fig:subtrees-heldout-accs}: the model can reuse learned substructures to assemble novel operator-tree shapes it has never been explicitly trained on.}
  \label{fig:subtrees-heldout-accs-seed2}
\end{figure}

\begin{figure}[h]
  \centering
  \includegraphics[width=0.93\linewidth]{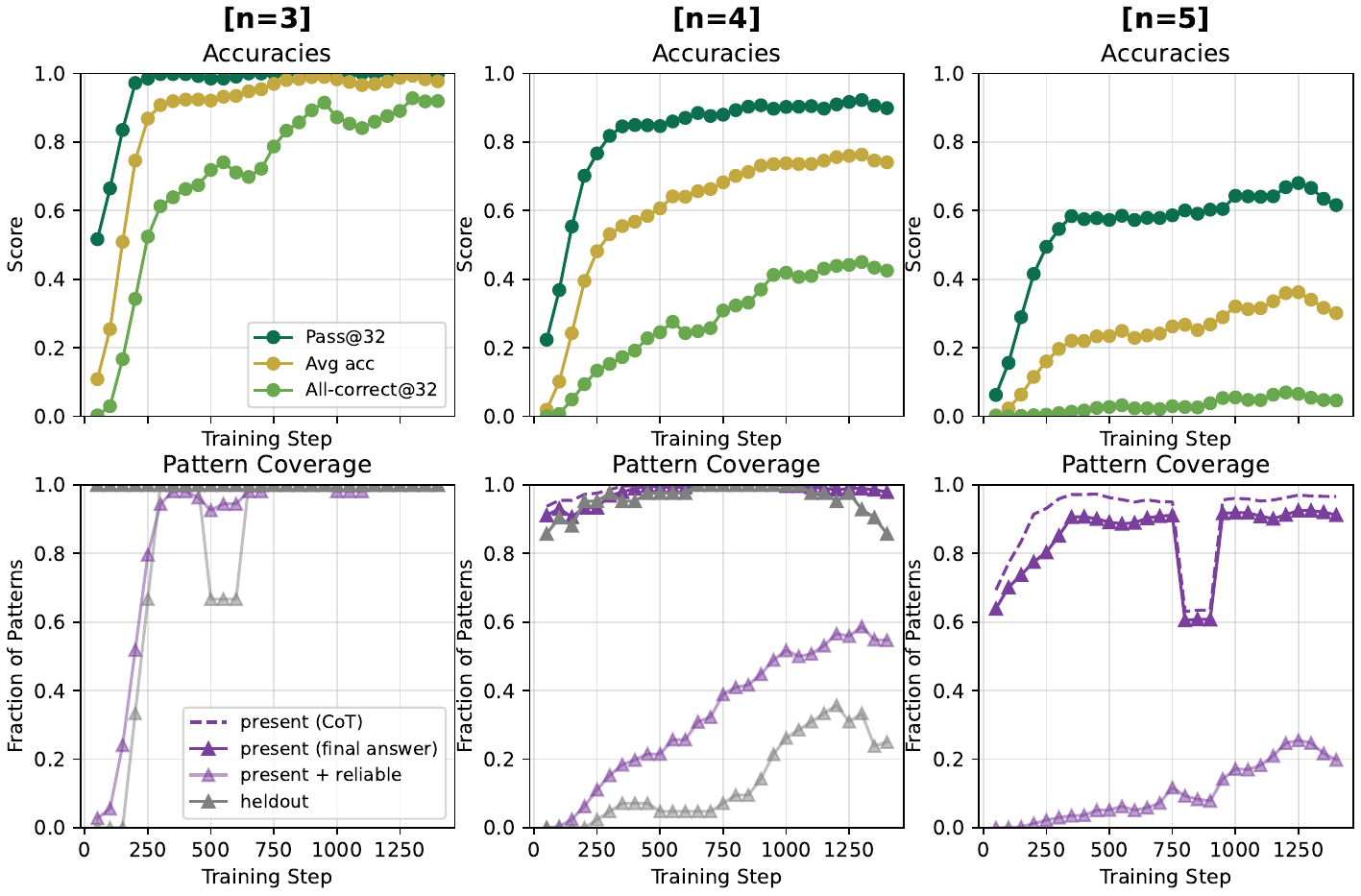}
  \caption{\textbf{Generalizes to entire families of held-out compositional patterns (Qwen2.5-1.5B, another seed).} The same conclusion from \Cref{fig:subtrees-heldout-accs}: the model can reuse learned substructures to assemble novel operator-tree shapes it has never been explicitly trained on.}
  \label{fig:subtrees-heldout-accs-seed3}
\end{figure}

\clearpage

\clearpage

\clearpage

\clearpage \clearpage
\clearpage
\section{List of Patterns}

\begin{table}[!h]
\caption{Expressions and their Canonical Forms (Patterns) (Page 1 of 4)}
\label{table:patterns_list_1}
\centering
\begin{tabular}{|l|l|l|l|}
\toprule
Expression & Pattern & Expression & Pattern \\ \hline
$(A+B)+C$ & $A+B+C$ & $A+(B+C)$ & $A+B+C$  \\ \hline
$(A+B)-C$ & $A+B-C$ & $A+(B-C)$ & $A+B-C$  \\ \hline
$(A+B) \times C$ & $(A+B) \times C$ & $A+(B \times C)$ & $A \times B+C$  \\ \hline
$(A+B)/C$ & $(A+B)/C$ & $A+(B/C)$ & $A/B+C$  \\ \hline
$(A-B)+C$ & $A+B-C$ & $A-(B+C)$ & $A-B-C$  \\ \hline
$(A-B)-C$ & $A-B-C$ & $A-(B-C)$ & $A+B-C$  \\ \hline
$(A-B) \times C$ & $(A-B) \times C$ & $A-(B \times C)$ & $A-B \times C$  \\ \hline
$(A-B)/C$ & $(A-B)/C$ & $A-(B/C)$ & $A-B/C$  \\ \hline
$(A \times B)+C$ & $A \times B+C$ & $A \times (B+C)$ & $(A+B) \times C$  \\ \hline
$(A \times B)-C$ & $A \times B-C$ & $A \times (B-C)$ & $(A-B) \times C$  \\ \hline
$(A \times B) \times C$ & $A \times B \times C$ & $A \times (B \times C)$ & $A \times B \times C$  \\ \hline
$(A \times B)/C$ & $A \times B/C$ & $A \times (B/C)$ & $A \times B/C$  \\ \hline
$(A/B)+C$ & $A/B+C$ & $A/(B+C)$ & $A/(B+C)$  \\ \hline
$(A/B)-C$ & $A/B-C$ & $A/(B-C)$ & $A/(B-C)$  \\ \hline
$(A/B) \times C$ & $A \times B/C$ & $A/(B \times C)$ & $A/B/C$  \\ \hline
$(A/B)/C$ & $A/B/C$ & $A/(B/C)$ & $A \times B/C$  \\ \hline
$A+(B+(C+D))$ & $A+B+C+D$ & $A+(B+(C-D))$ & $A+B+C-D$  \\ \hline
$A+(B+(C \times D))$ & $A \times B+C+D$ & $A+(B+(C/D))$ & $A/B+C+D$  \\ \hline
$A+(B-(C+D))$ & $A+B-C-D$ & $A+(B-(C-D))$ & $A+B+C-D$  \\ \hline
$A+(B-(C \times D))$ & $A+B-C \times D$ & $A+(B-(C/D))$ & $A+B-C/D$  \\ \hline
$A+(B \times (C+D))$ & $(A+B) \times C+D$ & $A+(B \times (C-D))$ & $(A-B) \times C+D$  \\ \hline
$A+(B \times (C \times D))$ & $A \times B \times C+D$ & $A+(B \times (C/D))$ & $A \times B/C+D$  \\ \hline
$A+(B/(C+D))$ & $A/(B+C)+D$ & $A+(B/(C-D))$ & $A/(B-C)+D$  \\ \hline
$A+(B/(C \times D))$ & $A/B/C+D$ & $A+(B/(C/D))$ & $A \times B/C+D$  \\ \hline
$A-(B+(C+D))$ & $A-B-C-D$ & $A-(B+(C-D))$ & $A+B-C-D$  \\ \hline
$A-(B+(C \times D))$ & $A-B \times C-D$ & $A-(B+(C/D))$ & $A-B/C-D$  \\ \hline
$A-(B-(C+D))$ & $A+B+C-D$ & $A-(B-(C-D))$ & $A+B-C-D$  \\ \hline
$A-(B-(C \times D))$ & $A \times B+C-D$ & $A-(B-(C/D))$ & $A/B+C-D$  \\ \hline
$A-(B \times (C+D))$ & $A-(B+C) \times D$ & $A-(B \times (C-D))$ & $A-(B-C) \times D$  \\ \hline
$A-(B \times (C \times D))$ & $A-B \times C \times D$ & $A-(B \times (C/D))$ & $A-B \times C/D$  \\ \hline
$A-(B/(C+D))$ & $A-B/(C+D)$ & $A-(B/(C-D))$ & $A-B/(C-D)$  \\ \hline
$A-(B/(C \times D))$ & $A-B/C/D$ & $A-(B/(C/D))$ & $A-B \times C/D$  \\ \hline
$A \times (B+(C+D))$ & $(A+B+C) \times D$ & $A \times (B+(C-D))$ & $(A+B-C) \times D$  \\ \hline
$A \times (B+(C \times D))$ & $(A \times B+C) \times D$ & $A \times (B+(C/D))$ & $(A/B+C) \times D$  \\ \hline
$A \times (B-(C+D))$ & $(A-B-C) \times D$ & $A \times (B-(C-D))$ & $(A+B-C) \times D$  \\ \hline
$A \times (B-(C \times D))$ & $(A-B \times C) \times D$ & $A \times (B-(C/D))$ & $(A-B/C) \times D$  \\ \hline
$A \times (B \times (C+D))$ & $(A+B) \times C \times D$ & $A \times (B \times (C-D))$ & $(A-B) \times C \times D$  \\ \hline
$A \times (B \times (C \times D))$ & $A \times B \times C \times D$ & $A \times (B \times (C/D))$ & $A \times B \times C/D$  \\ \hline
$A \times (B/(C+D))$ & $A \times B/(C+D)$ & $A \times (B/(C-D))$ & $A \times B/(C-D)$  \\ \hline
$A \times (B/(C \times D))$ & $A \times B/C/D$ & $A \times (B/(C/D))$ & $A \times B \times C/D$  \\ \hline
$A/(B+(C+D))$ & $A/(B+C+D)$ & $A/(B+(C-D))$ & $A/(B+C-D)$  \\ \hline
$A/(B+(C \times D))$ & $A/(B \times C+D)$ & $A/(B+(C/D))$ & $A/(B/C+D)$  \\ \hline
$A/(B-(C+D))$ & $A/(B-C-D)$ & $A/(B-(C-D))$ & $A/(B+C-D)$  \\ \hline
$A/(B-(C \times D))$ & $A/(B-C \times D)$ & $A/(B-(C/D))$ & $A/(B-C/D)$  \\ \hline
$A/(B \times (C+D))$ & $A/(B+C)/D$ & $A/(B \times (C-D))$ & $A/(B-C)/D$  \\ \hline
$A/(B \times (C \times D))$ & $A/B/C/D$ & $A/(B \times (C/D))$ & $A \times B/C/D$  \\ \hline
$A/(B/(C+D))$ & $(A+B) \times C/D$ & $A/(B/(C-D))$ & $(A-B) \times C/D$  \\ \hline
$A/(B/(C \times D))$ & $A \times B \times C/D$ & $A/(B/(C/D))$ & $A \times B/C/D$  \\ \hline
$A+((B+C)+D)$ & $A+B+C+D$ & $A+((B-C)+D)$ & $A+B+C-D$  \\ \hline
$A+((B \times C)+D)$ & $A \times B+C+D$ & $A+((B/C)+D)$ & $A/B+C+D$  \\ \bottomrule
\end{tabular}
\end{table}

\clearpage
\begin{table}[!t]
\caption{Expressions and their Canonical Forms (Patterns) (Page 2 of 4)}
\label{table:patterns_list_2}
\centering
\begin{tabular}{|l|l|l|l|}
\toprule
Expression & Pattern & Expression & Pattern \\ \hline
$A+((B+C)-D)$ & $A+B+C-D$ & $A+((B-C)-D)$ & $A+B-C-D$  \\ \hline
$A+((B \times C)-D)$ & $A \times B+C-D$ & $A+((B/C)-D)$ & $A/B+C-D$  \\ \hline
$A+((B+C) \times D)$ & $(A+B) \times C+D$ & $A+((B-C) \times D)$ & $(A-B) \times C+D$  \\ \hline
$A+((B \times C) \times D)$ & $A \times B \times C+D$ & $A+((B/C) \times D)$ & $A \times B/C+D$  \\ \hline
$A+((B+C)/D)$ & $(A+B)/C+D$ & $A+((B-C)/D)$ & $(A-B)/C+D$  \\ \hline
$A+((B \times C)/D)$ & $A \times B/C+D$ & $A+((B/C)/D)$ & $A/B/C+D$  \\ \hline
$A-((B+C)+D)$ & $A-B-C-D$ & $A-((B-C)+D)$ & $A+B-C-D$  \\ \hline
$A-((B \times C)+D)$ & $A-B \times C-D$ & $A-((B/C)+D)$ & $A-B/C-D$  \\ \hline
$A-((B+C)-D)$ & $A+B-C-D$ & $A-((B-C)-D)$ & $A+B+C-D$  \\ \hline
$A-((B \times C)-D)$ & $A+B-C \times D$ & $A-((B/C)-D)$ & $A+B-C/D$  \\ \hline
$A-((B+C) \times D)$ & $A-(B+C) \times D$ & $A-((B-C) \times D)$ & $A-(B-C) \times D$  \\ \hline
$A-((B \times C) \times D)$ & $A-B \times C \times D$ & $A-((B/C) \times D)$ & $A-B \times C/D$  \\ \hline
$A-((B+C)/D)$ & $A-(B+C)/D$ & $A-((B-C)/D)$ & $A-(B-C)/D$  \\ \hline
$A-((B \times C)/D)$ & $A-B \times C/D$ & $A-((B/C)/D)$ & $A-B/C/D$  \\ \hline
$A \times ((B+C)+D)$ & $(A+B+C) \times D$ & $A \times ((B-C)+D)$ & $(A+B-C) \times D$  \\ \hline
$A \times ((B \times C)+D)$ & $(A \times B+C) \times D$ & $A \times ((B/C)+D)$ & $(A/B+C) \times D$  \\ \hline
$A \times ((B+C)-D)$ & $(A+B-C) \times D$ & $A \times ((B-C)-D)$ & $(A-B-C) \times D$  \\ \hline
$A \times ((B \times C)-D)$ & $(A \times B-C) \times D$ & $A \times ((B/C)-D)$ & $(A/B-C) \times D$  \\ \hline
$A \times ((B+C) \times D)$ & $(A+B) \times C \times D$ & $A \times ((B-C) \times D)$ & $(A-B) \times C \times D$  \\ \hline
$A \times ((B \times C) \times D)$ & $A \times B \times C \times D$ & $A \times ((B/C) \times D)$ & $A \times B \times C/D$  \\ \hline
$A \times ((B+C)/D)$ & $(A+B) \times C/D$ & $A \times ((B-C)/D)$ & $(A-B) \times C/D$  \\ \hline
$A \times ((B \times C)/D)$ & $A \times B \times C/D$ & $A \times ((B/C)/D)$ & $A \times B/C/D$  \\ \hline
$A/((B+C)+D)$ & $A/(B+C+D)$ & $A/((B-C)+D)$ & $A/(B+C-D)$  \\ \hline
$A/((B \times C)+D)$ & $A/(B \times C+D)$ & $A/((B/C)+D)$ & $A/(B/C+D)$  \\ \hline
$A/((B+C)-D)$ & $A/(B+C-D)$ & $A/((B-C)-D)$ & $A/(B-C-D)$  \\ \hline
$A/((B \times C)-D)$ & $A/(B \times C-D)$ & $A/((B/C)-D)$ & $A/(B/C-D)$  \\ \hline
$A/((B+C) \times D)$ & $A/(B+C)/D$ & $A/((B-C) \times D)$ & $A/(B-C)/D$  \\ \hline
$A/((B \times C) \times D)$ & $A/B/C/D$ & $A/((B/C) \times D)$ & $A \times B/C/D$  \\ \hline
$A/((B+C)/D)$ & $A \times B/(C+D)$ & $A/((B-C)/D)$ & $A \times B/(C-D)$  \\ \hline
$A/((B \times C)/D)$ & $A \times B/C/D$ & $A/((B/C)/D)$ & $A \times B \times C/D$  \\ \hline
$(A+B)+(C+D)$ & $A+B+C+D$ & $(A+B)+(C-D)$ & $A+B+C-D$  \\ \hline
$(A+B)+(C \times D)$ & $A \times B+C+D$ & $(A+B)+(C/D)$ & $A/B+C+D$  \\ \hline
$(A-B)+(C+D)$ & $A+B+C-D$ & $(A-B)+(C-D)$ & $A+B-C-D$  \\ \hline
$(A-B)+(C \times D)$ & $A \times B+C-D$ & $(A-B)+(C/D)$ & $A/B+C-D$  \\ \hline
$(A \times B)+(C+D)$ & $A \times B+C+D$ & $(A \times B)+(C-D)$ & $A \times B+C-D$  \\ \hline
$(A \times B)+(C \times D)$ & $A \times B+C \times D$ & $(A \times B)+(C/D)$ & $A \times B+C/D$  \\ \hline
$(A/B)+(C+D)$ & $A/B+C+D$ & $(A/B)+(C-D)$ & $A/B+C-D$  \\ \hline
$(A/B)+(C \times D)$ & $A \times B+C/D$ & $(A/B)+(C/D)$ & $A/B+C/D$  \\ \hline
$(A+B)-(C+D)$ & $A+B-C-D$ & $(A+B)-(C-D)$ & $A+B+C-D$  \\ \hline
$(A+B)-(C \times D)$ & $A+B-C \times D$ & $(A+B)-(C/D)$ & $A+B-C/D$  \\ \hline
$(A-B)-(C+D)$ & $A-B-C-D$ & $(A-B)-(C-D)$ & $A+B-C-D$  \\ \hline
$(A-B)-(C \times D)$ & $A-B \times C-D$ & $(A-B)-(C/D)$ & $A-B/C-D$  \\ \hline
$(A \times B)-(C+D)$ & $A \times B-C-D$ & $(A \times B)-(C-D)$ & $A \times B+C-D$  \\ \hline
$(A \times B)-(C \times D)$ & $A \times B-C \times D$ & $(A \times B)-(C/D)$ & $A \times B-C/D$  \\ \hline
$(A/B)-(C+D)$ & $A/B-C-D$ & $(A/B)-(C-D)$ & $A/B+C-D$  \\ \hline
$(A/B)-(C \times D)$ & $A/B-C \times D$ & $(A/B)-(C/D)$ & $A/B-C/D$  \\ \hline
$(A+B) \times (C+D)$ & $(A+B) \times (C+D)$ & $(A+B) \times (C-D)$ & $(A+B) \times (C-D)$  \\ \hline
$(A+B) \times (C \times D)$ & $(A+B) \times C \times D$ & $(A+B) \times (C/D)$ & $(A+B) \times C/D$  \\ \hline
$(A-B) \times (C+D)$ & $(A+B) \times (C-D)$ & $(A-B) \times (C-D)$ & $(A-B) \times (C-D)$  \\ \hline
$(A-B) \times (C \times D)$ & $(A-B) \times C \times D$ & $(A-B) \times (C/D)$ & $(A-B) \times C/D$  \\ \bottomrule
\end{tabular}
\end{table}

\clearpage
\begin{table}[!t]
\caption{Expressions and their Canonical Forms (Patterns) (Page 3 of 4)}
\label{table:patterns_list_3}
\centering
\begin{tabular}{|l|l|l|l|}
\toprule
Expression & Pattern & Expression & Pattern \\ \hline
$(A \times B) \times (C+D)$ & $(A+B) \times C \times D$ & $(A \times B) \times (C-D)$ & $(A-B) \times C \times D$  \\ \hline
$(A \times B) \times (C \times D)$ & $A \times B \times C \times D$ & $(A \times B) \times (C/D)$ & $A \times B \times C/D$  \\ \hline
$(A/B) \times (C+D)$ & $(A+B) \times C/D$ & $(A/B) \times (C-D)$ & $(A-B) \times C/D$  \\ \hline
$(A/B) \times (C \times D)$ & $A \times B \times C/D$ & $(A/B) \times (C/D)$ & $A \times B/C/D$  \\ \hline
$(A+B)/(C+D)$ & $(A+B)/(C+D)$ & $(A+B)/(C-D)$ & $(A+B)/(C-D)$  \\ \hline
$(A+B)/(C \times D)$ & $(A+B)/C/D$ & $(A+B)/(C/D)$ & $(A+B) \times C/D$  \\ \hline
$(A-B)/(C+D)$ & $(A-B)/(C+D)$ & $(A-B)/(C-D)$ & $(A-B)/(C-D)$  \\ \hline
$(A-B)/(C \times D)$ & $(A-B)/C/D$ & $(A-B)/(C/D)$ & $(A-B) \times C/D$  \\ \hline
$(A \times B)/(C+D)$ & $A \times B/(C+D)$ & $(A \times B)/(C-D)$ & $A \times B/(C-D)$  \\ \hline
$(A \times B)/(C \times D)$ & $A \times B/C/D$ & $(A \times B)/(C/D)$ & $A \times B \times C/D$  \\ \hline
$(A/B)/(C+D)$ & $A/(B+C)/D$ & $(A/B)/(C-D)$ & $A/(B-C)/D$  \\ \hline
$(A/B)/(C \times D)$ & $A/B/C/D$ & $(A/B)/(C/D)$ & $A \times B/C/D$  \\ \hline
$(A+(B+C))+D$ & $A+B+C+D$ & $(A+(B-C))+D$ & $A+B+C-D$  \\ \hline
$(A+(B \times C))+D$ & $A \times B+C+D$ & $(A+(B/C))+D$ & $A/B+C+D$  \\ \hline
$(A-(B+C))+D$ & $A+B-C-D$ & $(A-(B-C))+D$ & $A+B+C-D$  \\ \hline
$(A-(B \times C))+D$ & $A+B-C \times D$ & $(A-(B/C))+D$ & $A+B-C/D$  \\ \hline
$(A \times (B+C))+D$ & $(A+B) \times C+D$ & $(A \times (B-C))+D$ & $(A-B) \times C+D$  \\ \hline
$(A \times (B \times C))+D$ & $A \times B \times C+D$ & $(A \times (B/C))+D$ & $A \times B/C+D$  \\ \hline
$(A/(B+C))+D$ & $A/(B+C)+D$ & $(A/(B-C))+D$ & $A/(B-C)+D$  \\ \hline
$(A/(B \times C))+D$ & $A/B/C+D$ & $(A/(B/C))+D$ & $A \times B/C+D$  \\ \hline
$(A+(B+C))-D$ & $A+B+C-D$ & $(A+(B-C))-D$ & $A+B-C-D$  \\ \hline
$(A+(B \times C))-D$ & $A \times B+C-D$ & $(A+(B/C))-D$ & $A/B+C-D$  \\ \hline
$(A-(B+C))-D$ & $A-B-C-D$ & $(A-(B-C))-D$ & $A+B-C-D$  \\ \hline
$(A-(B \times C))-D$ & $A-B \times C-D$ & $(A-(B/C))-D$ & $A-B/C-D$  \\ \hline
$(A \times (B+C))-D$ & $(A+B) \times C-D$ & $(A \times (B-C))-D$ & $(A-B) \times C-D$  \\ \hline
$(A \times (B \times C))-D$ & $A \times B \times C-D$ & $(A \times (B/C))-D$ & $A \times B/C-D$  \\ \hline
$(A/(B+C))-D$ & $A/(B+C)-D$ & $(A/(B-C))-D$ & $A/(B-C)-D$  \\ \hline
$(A/(B \times C))-D$ & $A/B/C-D$ & $(A/(B/C))-D$ & $A \times B/C-D$  \\ \hline
$(A+(B+C)) \times D$ & $(A+B+C) \times D$ & $(A+(B-C)) \times D$ & $(A+B-C) \times D$  \\ \hline
$(A+(B \times C)) \times D$ & $(A \times B+C) \times D$ & $(A+(B/C)) \times D$ & $(A/B+C) \times D$  \\ \hline
$(A-(B+C)) \times D$ & $(A-B-C) \times D$ & $(A-(B-C)) \times D$ & $(A+B-C) \times D$  \\ \hline
$(A-(B \times C)) \times D$ & $(A-B \times C) \times D$ & $(A-(B/C)) \times D$ & $(A-B/C) \times D$  \\ \hline
$(A \times (B+C)) \times D$ & $(A+B) \times C \times D$ & $(A \times (B-C)) \times D$ & $(A-B) \times C \times D$  \\ \hline
$(A \times (B \times C)) \times D$ & $A \times B \times C \times D$ & $(A \times (B/C)) \times D$ & $A \times B \times C/D$  \\ \hline
$(A/(B+C)) \times D$ & $A \times B/(C+D)$ & $(A/(B-C)) \times D$ & $A \times B/(C-D)$  \\ \hline
$(A/(B \times C)) \times D$ & $A \times B/C/D$ & $(A/(B/C)) \times D$ & $A \times B \times C/D$  \\ \hline
$(A+(B+C))/D$ & $(A+B+C)/D$ & $(A+(B-C))/D$ & $(A+B-C)/D$  \\ \hline
$(A+(B \times C))/D$ & $(A \times B+C)/D$ & $(A+(B/C))/D$ & $(A/B+C)/D$  \\ \hline
$(A-(B+C))/D$ & $(A-B-C)/D$ & $(A-(B-C))/D$ & $(A+B-C)/D$  \\ \hline
$(A-(B \times C))/D$ & $(A-B \times C)/D$ & $(A-(B/C))/D$ & $(A-B/C)/D$  \\ \hline
$(A \times (B+C))/D$ & $(A+B) \times C/D$ & $(A \times (B-C))/D$ & $(A-B) \times C/D$  \\ \hline
$(A \times (B \times C))/D$ & $A \times B \times C/D$ & $(A \times (B/C))/D$ & $A \times B/C/D$  \\ \hline
$(A/(B+C))/D$ & $A/(B+C)/D$ & $(A/(B-C))/D$ & $A/(B-C)/D$  \\ \hline
$(A/(B \times C))/D$ & $A/B/C/D$ & $(A/(B/C))/D$ & $A \times B/C/D$  \\ \hline
$((A+B)+C)+D$ & $A+B+C+D$ & $((A-B)+C)+D$ & $A+B+C-D$  \\ \hline
$((A \times B)+C)+D$ & $A \times B+C+D$ & $((A/B)+C)+D$ & $A/B+C+D$  \\ \hline
$((A+B)-C)+D$ & $A+B+C-D$ & $((A-B)-C)+D$ & $A+B-C-D$  \\ \hline
$((A \times B)-C)+D$ & $A \times B+C-D$ & $((A/B)-C)+D$ & $A/B+C-D$  \\ \hline
$((A+B) \times C)+D$ & $(A+B) \times C+D$ & $((A-B) \times C)+D$ & $(A-B) \times C+D$  \\ \hline
$((A \times B) \times C)+D$ & $A \times B \times C+D$ & $((A/B) \times C)+D$ & $A \times B/C+D$  \\ \bottomrule
\end{tabular}
\end{table}

\clearpage
\begin{table}[!t]
\caption{Expressions and their Canonical Forms (Patterns) (Page 4 of 4)}
\label{table:patterns_list_4}
\centering
\begin{tabular}{|l|l|l|l|}
\toprule
Expression & Pattern & Expression & Pattern \\ \hline
$((A+B)/C)+D$ & $(A+B)/C+D$ & $((A-B)/C)+D$ & $(A-B)/C+D$  \\ \hline
$((A \times B)/C)+D$ & $A \times B/C+D$ & $((A/B)/C)+D$ & $A/B/C+D$  \\ \hline
$((A+B)+C)-D$ & $A+B+C-D$ & $((A-B)+C)-D$ & $A+B-C-D$  \\ \hline
$((A \times B)+C)-D$ & $A \times B+C-D$ & $((A/B)+C)-D$ & $A/B+C-D$  \\ \hline
$((A+B)-C)-D$ & $A+B-C-D$ & $((A-B)-C)-D$ & $A-B-C-D$  \\ \hline
$((A \times B)-C)-D$ & $A \times B-C-D$ & $((A/B)-C)-D$ & $A/B-C-D$  \\ \hline
$((A+B) \times C)-D$ & $(A+B) \times C-D$ & $((A-B) \times C)-D$ & $(A-B) \times C-D$  \\ \hline
$((A \times B) \times C)-D$ & $A \times B \times C-D$ & $((A/B) \times C)-D$ & $A \times B/C-D$  \\ \hline
$((A+B)/C)-D$ & $(A+B)/C-D$ & $((A-B)/C)-D$ & $(A-B)/C-D$  \\ \hline
$((A \times B)/C)-D$ & $A \times B/C-D$ & $((A/B)/C)-D$ & $A/B/C-D$  \\ \hline
$((A+B)+C) \times D$ & $(A+B+C) \times D$ & $((A-B)+C) \times D$ & $(A+B-C) \times D$  \\ \hline
$((A \times B)+C) \times D$ & $(A \times B+C) \times D$ & $((A/B)+C) \times D$ & $(A/B+C) \times D$  \\ \hline
$((A+B)-C) \times D$ & $(A+B-C) \times D$ & $((A-B)-C) \times D$ & $(A-B-C) \times D$  \\ \hline
$((A \times B)-C) \times D$ & $(A \times B-C) \times D$ & $((A/B)-C) \times D$ & $(A/B-C) \times D$  \\ \hline
$((A+B) \times C) \times D$ & $(A+B) \times C \times D$ & $((A-B) \times C) \times D$ & $(A-B) \times C \times D$  \\ \hline
$((A \times B) \times C) \times D$ & $A \times B \times C \times D$ & $((A/B) \times C) \times D$ & $A \times B \times C/D$  \\ \hline
$((A+B)/C) \times D$ & $(A+B) \times C/D$ & $((A-B)/C) \times D$ & $(A-B) \times C/D$  \\ \hline
$((A \times B)/C) \times D$ & $A \times B \times C/D$ & $((A/B)/C) \times D$ & $A \times B/C/D$  \\ \hline
$((A+B)+C)/D$ & $(A+B+C)/D$ & $((A-B)+C)/D$ & $(A+B-C)/D$  \\ \hline
$((A \times B)+C)/D$ & $(A \times B+C)/D$ & $((A/B)+C)/D$ & $(A/B+C)/D$  \\ \hline
$((A+B)-C)/D$ & $(A+B-C)/D$ & $((A-B)-C)/D$ & $(A-B-C)/D$  \\ \hline
$((A \times B)-C)/D$ & $(A \times B-C)/D$ & $((A/B)-C)/D$ & $(A/B-C)/D$  \\ \hline
$((A+B) \times C)/D$ & $(A+B) \times C/D$ & $((A-B) \times C)/D$ & $(A-B) \times C/D$  \\ \hline
$((A \times B) \times C)/D$ & $A \times B \times C/D$ & $((A/B) \times C)/D$ & $A \times B/C/D$  \\ \hline
$((A+B)/C)/D$ & $(A+B)/C/D$ & $((A-B)/C)/D$ & $(A-B)/C/D$  \\ \hline
$((A \times B)/C)/D$ & $A \times B/C/D$ & $((A/B)/C)/D$ & $A/B/C/D$  \\ \bottomrule
\end{tabular}
\end{table}

\clearpage \clearpage
\section{Acknowledging LLM Usage}
\label{appendix:llm}
At various stages of the project, we received the aid of LLMs. Here, we report all such use cases.

\begin{itemize}
    \item We received aid on writing the code to automate the process of extracting and canonicalizing an expression. The list of $n=4$ patterns were initially manually mapped and were used as test cases for the LLM generated code. All code was manually verified.

    \item We received aid on writing the code to draw plots from existing data that were generated from human-written scripts. All code was manually verified.

    \item We used an LLM for copy-editing tasks, such as rephrasing sentences to improve clarity, conciseness, and flow, as well as for correcting grammatical errors.
\end{itemize} \clearpage
\section{Licenses of Models and Codebase Used In This Paper}
\label{appendix:license}

\begin{enumerate}
    \item \texttt{Qwen-2.5} models \citep{qwen2025qwen25technicalreport} have Qwen Research license.
    \item \texttt{Llama-3.1} models have Llama3.1 license.
    \item \texttt{Llama-3.2} models have Llama3.2 license.
    \item \texttt{Qwen-3} models \citep{yang2025qwen3technicalreport} have Apache-2.0 license.
    \item \texttt{Goedel-Prover-V2-32B} model and its SFT dataset \texttt{Goedel-LM} / \texttt{SFT\_dataset\_v2} \citep{goedelv2} have Apache-2.0 license.
    \item The LeanDojo 4 Benchmark \citep{yang2023leandojo} has CC BY 2.0 license.
    \item Mathlib4 \citep{mathlib2020} has Apache-2.0 license.
    \item The codebase \citet{tinyzero}, from which we derived our data generation pipeline, has Apache-2.0 license.
\end{enumerate}
 \clearpage


\end{document}